\documentclass[preprint,12pt,authoryear]{elsarticle}
\usepackage[ruled,vlined,linesnumbered]{algorithm2e}
\usepackage{algpseudocode}
\usepackage{amsmath, amssymb, booktabs}
\usepackage{geometry}
\usepackage{graphicx} 
\usepackage{caption}
\usepackage{multirow}
\usepackage{subcaption}
 \usepackage{float}
 \usepackage{siunitx}
\usepackage{tikz}
\usetikzlibrary{quotes}
\usetikzlibrary{automata}
\usetikzlibrary{calc,positioning}
\usetikzlibrary{arrows.meta, positioning, shapes.geometric}

\tikzset{
  block/.style    = {rectangle, draw, rounded corners, align=center, minimum width=3.2cm, minimum height=1cm},
  IO/.style    = {circle, draw, rounded corners, align=center, minimum width=3.2cm, minimum height=1cm},
  decision/.style = {diamond, draw, aspect=2, align=center, inner xsep=1.2em, inner ysep=0.8em},
  line/.style     = {->, thick},
  font=\small
}

\tikzstyle{mainnode} = [rectangle, rounded corners, minimum width=3cm, minimum height=1cm, text centered, draw=black, fill=blue!30]
\tikzstyle{subnode} = [rectangle, rounded corners, minimum width=2.5cm, minimum height=0.8cm, text centered, draw=black, fill=cyan!30]
\tikzstyle{startnode} = [circle, rounded corners, minimum width=2.5cm, minimum height=0.8cm, text centered, draw=black, fill=red!30]
\tikzstyle{endnode} = [rectangle, rounded corners, minimum width=2.5cm, minimum height=0.8cm, text centered, draw=black, fill=orange!30]
\tikzstyle{arrow} = [thick,->,>=stealth]
\usepackage[export]{adjustbox} 
\usepackage{amsthm}

\journal{EJOR}

\begin{document}

\begin{frontmatter}

\title{Kempe Swap $K$-Means: A Scalable Near-Optimal Solution for Semi-Supervised Clustering\tnoteref{label1}}
\author{Yuxuan Ren\corref{cor1}\fnref{label2}}
\ead{yren79@gatech.edu}
\author{Shijie Deng\corref{cor2}\fnref{label3}}
\ead{sd111@gatech.edu}
\affiliation{organization={School of Industrial and Systems Engineering \\
Georgia Institute of Technology},
            country={USA}
            }


\begin{abstract}

This paper presents a novel centroid-based heuristic algorithm, termed Kempe Swap $K$-Means, for constrained clustering under rigid must-link (ML) and cannot-link (CL) constraints. The algorithm employs a dual-phase iterative process: an assignment step that utilizes Kempe chain swaps to refine current clustering in the constrained solution space and a centroid update step that computes optimal cluster centroids. To enhance global search capabilities and avoid local optima, the framework incorporates controlled perturbations during the update phase. Empirical evaluations demonstrate that the proposed method achieves near-optimal partitions while maintaining high computational efficiency and scalability. The results indicate that Kempe Swap $K$-Means consistently outperforms state-of-the-art benchmarks in both clustering accuracy and algorithmic efficiency for large-scale datasets.
\end{abstract}




\end{frontmatter}

\section{Introduction}
\label{sec1}
Clustering is a fundamental tool in modern applications such as pattern recognition, data mining, computer vision, machine learning, and knowledge discovery. However, in the absence of supervision, clustering algorithms may generate partitions that do not align with expert understanding or domain structure. To enhance cluster quality and obtain more meaningful solutions, researchers have increasingly explored ways to incorporate prior knowledge so that the clustering process can be guided and constrained in a principled manner. 

Semi-supervised clustering, often referred to as constrained clustering, has become a prominent paradigm for incorporating prior knowledge into otherwise unsupervised learning tasks. Unlike conventional clustering methods that rely exclusively on geometric similarity in the feature space, constrained clustering enables domain knowledge, expert annotations, or user-defined preferences to be explicitly encoded as constraints on the clustering structure. In particular, must-link constraints enforce that pairs of instances are assigned to the same cluster, whereas cannot-link constraints require them to belong to different clusters. This additional supervision can substantially improve both the interpretability and practical relevance of the resulting partitions. 

These advantages have been demonstrated across a range of application domains. In bioinformatics, for instance, gene expression data are typically high-dimensional and characterized by complex dependencies among genes and experimental conditions, which are frequently overlooked by standard clustering algorithms. By extending co-clustering frameworks to accommodate explicit constraints, \cite{pensa2008constrained} demonstrated that biologically motivated prior knowledge—such as known functional relationships among genes—can be systematically integrated into the clustering process, producing results that are more scientifically meaningful and better suited for downstream biological analysis. In information retrieval and document management, classical text clustering approaches treat documents as unlabeled data points, despite the fact that users often have partial knowledge about document relationships. \cite{huang2006text} showed that encoding such knowledge as must-link and cannot-link constraints can guide the clustering process toward structures that better reflect users’ semantic intent and organizational goals, thereby improving the interpretability and practical utility of document clusters. In electric power markets, accurate forecasting of locational marginal prices (LMPs) in a power grid is critical for electricity trading and risk management. LMPs are determined by the System Operator through the solution of a security-constrained economic dispatch (SCED) problem that clears electricity generations and loads subject to transmission network and operational constraints. From a multiparametric programming perspective, the SCED solution induces a partition of the feasible load space into a collection of critical regions, within each of which LMPs and shadow prices of transmission lines depend affinely on system loads. Because the number of critical regions can grow exponentially with the number of nodes in the power grid, clustering-based methods are commonly adopted to construct probabilistic forecasting models that trade off bias and variance. By incorporating domain knowledge through pairwise constraints, such as must-link constraints for operating points with identical binding transmission constraints and cannot-link constraints for points that differ substantially in load or meteorological conditions, constrained clustering can more effectively identify data associated with similar critical regions, thereby improving the accuracy and robustness of LMP forecasts.

Constrained clustering extends classical clustering frameworks by explicitly incorporating domain knowledge, but this added expressiveness introduces significant challenges related to feasibility, optimality, and computational tractability. In the presence of hard cannot-link constraints, the clustering problem becomes closely related to the NP-hard graph coloring problem \citep{brooks1941colouring}. As a result, clustering algorithms must contend with complex interactions among constraints, which may render some problem instances infeasible or give rise to a large number of locally optimal solutions.

Exact solutions to constrained clustering can be obtained using methods such as constraint programming \citep{dao2013constraint}, column generation \citep{babaki2014constrained}, and positive semi-definite programming \citep{piccialli2022exact}. However, the applicability of these approaches is typically limited to relatively small datasets, often on the order of a few hundred instances, due to the inherent combinatorial nature of the problem, the reliance on Big-M constraints in linear formulations, or the rank constraints required in semi-definite formulations. Consequently, for large-scale datasets, research has increasingly focused on suboptimal but scalable heuristic algorithms. In particular, centroid-based iterative methods inspired by the $K$-means algorithm, which alternate between constrained assignment steps and centroid updates, have been shown to produce high-quality, near locally optimal solutions at a fraction of the computational cost of exact methods.


The classical COP-$K$-Means algorithm, introduced by \cite{wagstaff2001constrained}, represents one of the earliest heuristic approaches to constrained clustering. Its assignment step, where data points are greedily assigned to the nearest centroid subject to must-link and cannot-link constraints, is the key distinction from the unconstrained $K$-Means algorithm. Despite its simplicity, COP-$K$-Means may generate infeasible intermediate solutions in which certain data points cannot be assigned to any cluster without violating constraints. Consequently, the algorithm is very sensitive to the order in which data points are processed. This limitation has motivated several subsequent improvements. For example, ICOP-$K$-Means \citep{tan2010improved} addresses constraint violations by ordering data points during the assignment step according to confidence scores, while CLC-$K$-Means \citep{yang2013improved} alleviates infeasibility by sequencing cannot-link instances using a breadth-first search strategy. Nevertheless, due to the presence of hard constraints, neither COP-$K$-Means nor its extensions can generally guarantee monotonic improvement of the clustering objective across iterations.

To enhance clustering quality, several studies have incorporated hard constraints into binary or integer programming formulations \citep{le2018binary, baumann2020, baumann2025algorithm}, in which each assignment step is posed as an optimization problem that enforces global satisfaction of all constraints. These approaches retain the iterative centroid and assignment structure of the $K$-Means algorithm while guaranteeing feasibility at every iteration and producing a monotonic decrease in the objective of within-cluster sum of squares (WCSS).

A recent strand of research combines integer programming based iterative schemes with concepts from genetic algorithms in order to improve exploration of the solution space. In this setting, locally optimal solutions are intentionally perturbed to escape poor local minima. For example, \cite{baumann2025algorithm} encourage exploration by relocating one centroid to the position of another, while \cite{mansueto2025efficiently} integrate memetic differential evolution into constrained clustering by maintaining multiple solution populations and generating offspring through crossover operations on population centroids. By explicitly incorporating mutation mechanisms, both approaches yield solutions that are closer to global optima.

Centroid-based iterative methods also serve as effective sources of high-quality upper-bound incumbents for exact branch-and-bound or branch-and-price algorithms \citep{babaki2014constrained, piccialli2022exact}. Consequently, improvements in centroid-based iterative heuristics directly facilitate progress in exact solution methods. However, existing iterative constrained clustering approaches rely on either greedy sequential assignment or integer programming based assignment, and they do not fully balance assignment quality with computational efficiency. While integer programming based methods ensure feasibility, they incur substantial computational cost due to the need of solving a fixed-centroid graph coloring problem at each iteration. \citet{baumann2025algorithm} partially alleviate this burden by restricting assignments to the $q$ nearest centroids rather than all $k$ centroids, thereby reducing the number of variables and constraints. Their PCCC algorithm is therefore able to handle datasets containing up to 70,000 points under a moderate number of cannot-link constraints. However, in order to guarantee feasibility, the parameter $q$ must be at least as large as the degree of each vertex in the cannot-link constraint graph. For dense constraint graphs, this reduction in complexity is thus limited.



Existing constrained clustering methods face a fundamental trade-off between solution quality, feasibility, and computational efficiency. Greedy assignment based heuristics offer scalability but often suffer from infeasibility and sensitivity to processing order, whereas integer programming based assignment methods guarantee feasibility and monotonic objective improvement at the expense of substantial computational overhead. In particular, the need to repeatedly solve fixed-centroid graph coloring problems limits the applicability of such approaches in settings with dense constraint graphs or large datasets. This gap highlights the need for alternative assignment mechanisms that preserve feasibility and enable effective local improvement while remaining computationally tractable.

Kempe chain exchange provides a natural foundation for addressing this challenge. As a well-established technique in graph coloring, Kempe chain exchange enables exploration of large, connected neighborhoods while maintaining feasibility under a fixed number of colors. Originally developed in the context of the four-color theorem, a Kempe chain is defined as a maximal connected subgraph induced by two color classes, within which adjacent vertices alternate between those colors. Exchanging the colors along such a chain yields a new legal coloring without increasing the chromatic number or introducing conflicts. This property has made Kempe chain exchanges particularly effective in local search based optimization for graph coloring. In practical scheduling applications, including university examination and course timetabling, Kempe swaps are widely used to perform substantial structural changes while respecting hard feasibility constraints \citep{chams1987some, chiarandini2018stochastic}.

Building on the above insights, this paper develops a constrained clustering framework that leverages Kempe chain exchanges to bridge the gap between greedy heuristics and integer programming based assignment methods. Feasible cluster assignments are improved through Kempe swaps applied in a steepest descent manner, enabling systematic objective reduction without sacrificing feasibility. The resulting assignment step admits a formulation as a maximum-weight independent set problem, which is significantly simpler than the fixed-centroid graph coloring formulations required by existing integer programming based approaches. Specifically, the main contributions are summarized as follows.
\begin{itemize}
\item A novel constrained cluster assignment mechanism based on Kempe chain exchange is developed, enabling systematic exploration of large, connected neighborhoods while preserving feasibility under hard must-link and cannot-link constraints. 
\item The constrained assignment step is reformulated as a maximum-weight independent set problem, which is substantially simpler than the fixed-centroid graph coloring formulations commonly used in existing integer programming based approaches, thereby achieving a more favorable balance between solution quality and computational complexity.
\item A centroid perturbation mechanism is incorporated to enhance exploration beyond local optima and enable broader neighborhood search without significantly degrading the clustering objective.
\item Extensive numerical experiments demonstrate that Kempe Swap $K$-Means and its extensions consistently outperform state-of-the-art constrained clustering algorithms in terms of solution quality, robustness, and computational efficiency, particularly for large-scale problems with dense constraint graphs.
\end{itemize}

The remainder of the paper is organized as follows. Section 2 presents the model formulation and describes the proposed algorithm. Section 3 reports numerical experiments that demonstrate the computational efficiency and solution quality of the approach. Section 4 concludes and discusses directions for future research.

\newpage
\section{Formulations and Algorithms}
\label{sec2}
\subsection{Background}
We introduce the following notations:
\begin{itemize}
 \item $\mathcal{ML}$: must-link constraints;
 \item $\mathcal{CL}$: cannot-link constraints;
 \item $V$: set of super-nodes of must-link data points;
 \item $E$: set of edges between cannot-link super-nodes;
 \item $C$: set of clusters/colors;
 \item $X$: decision variables, with $x_{ij} = 1$ if $v_i\in V$ belong to cluster $C_j$;
 \item $y$: data to be clustered, with $y_i$ being the data associated with super-node $v_i\in V$;
 \item $n$: number of data;
 \item $p$: dimension of data;
 \item $k$: number of centroids;
 \item $\mu$: cluster centroids;
 \item $U$: membership of data;
 \item $\mathbb{H}$: set of Kempe chains between two clusters, with $H = (H_i,H_j)\in\mathbb{H}$;
 \item $d_{H}$: cost of swap on Kempe chain $H$;
 \item $S$: set of Kempe swaps;
 \item $\mathcal{CC}^{swap}$: set of clique constraints for swaps;
 \item $\mathcal{CL}^{swap}$: set of cannot-link constraints for swaps;
 \item $||\cdot||$: $L_2$ norm;
 \item $||\cdot||_F$: Frobenius norm;
 \item $\bar{\cdot}$: mean operator.
\end{itemize}

Clustering with must-link $\mathcal{ML}$ and cannot-link $\mathcal{CL}$ constraints can be formulated as a Mixed Integer Second Order Cone Program: 
\begin{subequations}\label{eq:1}
\begin{align}
 \min_{x, m} \quad & \sum_{i=1}^{n} \sum_{j=1}^{k} x_{ij} \| y_i - \mu_j \|_2^2 \tag{\ref{eq:1}} \\
 \text{s.t.} \quad 
 & \sum_{j=1}^{k} x_{ij} = 1, && \forall i \in [n], \label{eq:1a}\\
 & \sum_{i=1}^{n} x_{ij} \ge 1, && \forall j \in [k], \label{eq:1b}\\
 & x_{ih} = x_{jh}, && \forall h \in [k],\ \forall (i,j) \in \mathcal{ML}, \label{eq:1c}\\
 & x_{ih} + x_{jh} \le 1, && \forall h \in [k],\ \forall (i,j) \in \mathcal{CL}, \label{eq:1d}\\
 & x_{ij} \in \{0,1\}, && \forall i \in [n],\ \forall j \in [k], \label{eq:1e}\\
 & \mu_j \in \mathbb{R}^d, && \forall j \in [k]. \label{eq:1f}
\end{align}
\end{subequations}
where $\{y_i\}_{i \in [n]}$ is the collection of data, $\{\mu_j\}_{j \in [k]}$ is the collection of cluster centroids, and $x_{i,j}$ indicates whether the data $y_i$ belong to the cluster $C_j$. With $\mathcal{CL}$ (\ref{eq:1d}), formulation (\ref{eq:1}) describes a $k$-coloring \citep{brooks1941colouring}. In the following sections, the terms \textit{color} and \textit{cluster} are used interchangeably.

However, exact solutions based on formulation (\ref{eq:1}), Semi-definite programming \citep{piccialli2022exact}, or column generation \citep{babaki2014constrained} can only be scaled to medium size problems. In this section, we introduce Kempe Swap $K$-Means, a heuristic for constrained clustering that approaches global optimality by iteratively performing Kempe chain swaps and centroid updates. The method is designed to scale to large datasets, offering capabilities beyond those of current state-of-the-art approaches.

\subsubsection{Preprocessing Must-link Constraints}
When preprocessing must-link constraints in constrained clustering, the \textbf{transitivity property} plays a key simplifying role. Because must-link relations are transitive --- that is, if \( (i,j)\in \mathcal{ML} \) and \( (j,h)\in \mathcal{ML} \), then \( (i,h)\in \mathcal{ML} \) --- all data points connected through any sequence of ML constraints form an equivalence class. These equivalence classes can be identified by computing the \textbf{transitive closure} of the ML graph, where vertices represent data points and edges represent ML relations. Each connected component in this graph can then be collapsed into a single ``super-node'' that aggregates all points in that component. This preprocessing effectively treats all mutually must-linked data as a unified observation, thus reducing the problem size and ensuring that all ML constraints are automatically satisfied during subsequent optimization or clustering. Such preprocessing is standard in exact constrained clustering solvers (e.g., \cite{piccialli2022exact}), where each super-node is represented by the map of its member data points, and only the remaining cannot-link constraints are retained for the reduced clustering problem. In the following, we embrace the same idea of simplification and encode must-link data points as a super-node of map to its members. Therefore, a semi-supervised clustering problem with must-link and cannot-link constraints can be modeled as a graph whose vertices $V$ are super-nodes, with edges $E$ indicating cannot-link relationships; a valid clustering is then given by a $k$-coloring of this graph $G=(V,E)$. In the following, we assume that the graph $G=(V,E)$ is connected, since coloring a disconnected graph can be decomposed into coloring its independent connected components.
\subsubsection{Kempe Chain Swap}

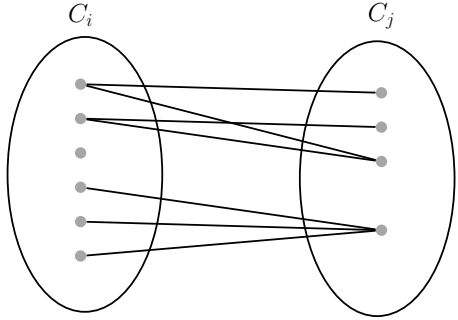
\begin{figure}[ht]
\centering
\resizebox{0.4\textwidth}{!}{%
\begin{tikzpicture}[
 every node/.style={font=\Large\bfseries\color{black}},
 dot/.style={circle,fill=gray!70,draw=none,minimum size=7.5pt,inner sep=0pt},
 edge/.style={black,line width=1.2pt},
 faint/.style={black!60,line width=.6pt},
 dashededge/.style={black,line width=1.2pt,dash pattern=on 5pt off 4pt}
]

\def\xL{0}
\def\xR{7}
\foreach \y/\name in {2.4/L1,1.6/L2,0.8/L3,0.0/L4,-0.8/L5,-1.6/L6}
 \node[dot] (\name) at (\xL,\y) {};
\foreach \y/\name in {2.2/R1,1.4/R2,0.6/R3,-1.0/R4}
 \node[dot] (\name) at (\xR,\y) {};

\draw[edge] (\xL+0.1,0.3) ellipse [x radius=1.8,y radius=3.2];
\draw[edge] (\xR-0.1,0.2) ellipse [x radius=1.8,y radius=3.2];

\node at (\xL,4) {$C_i$};
\node at (\xR,4) {$C_j$};


\draw[edge] (L1) -- (R1);
\draw[edge] (L1) -- (R3);

\draw[edge] (L2) -- (R2);
\draw[edge] (L2) -- (R3);

\draw[edge] (L5) -- (R4);
\draw[edge] (L6) -- (R4);
\draw[edge] (L4) -- (R4);


\end{tikzpicture}
}

\caption{Kempe Chains between clusters $C_i$ and $C_j$.}
\label{fig:KempeChain}
\end{figure}

A \textbf{Kempe chain} is defined with respect to two colors (say, color $C_i$ and $C_j$): it is a maximal connected subgraph comprising vertices (super-nodes) that are colored $C_i$ or $C_j$. Essentially, within the current coloring, consider the induced subgraph on vertices of color $C_i$ or $C_j$; each connected component of this subgraph is a Kempe chain. A Kempe chain move swaps the colors on such a component: all vertices in the chain that were of color $C_i$ are changed to $C_j$, and vise versa. Remarkably, a Kempe chain interchange is a two-color local search move that can recolor a set of vertices in one step without ever breaking the feasibility. No edge can become conflicted by this swap, because any edge inside the chain was originally connecting opposite-colored vertices, and it remains so after swapping, and edges from the chain to the rest of the graph still see the same set of colors on their endpoints. Figure \ref{fig:KempeChain} illustrates three (two connected subgraphs and a single vertex). Kempe chains between color classes $C_i$ and $C_j$ – swapping their colors yields a new coloring without conflicts. Kempe chain moves are a classic technique in graph coloring heuristics and allow exploring the space of colorings more broadly than single-vertex moves.

The configuration of Kempe chains depends on the current coloring of the graph, and the Kempe chain moves represent a large-scale, structured coloring solution neighborhood from the current one. They allow exploration of colorings that differ in potentially many vertices at once, thus greatly enlarging the reachable solution space from any given coloring. This broad exploration can help avoid deep local minima – for instance, a coloring that is locally optimal under one-vertex moves might be improved by a carefully chosen Kempe chain swap that a one-vertex move could never achieve. 

A $k$-coloring of a graph is \textbf{Kempe-equivalent} to another $k$-coloring if they are connected by a series of Kempe swaps. \cite{las1981kempe} showed that theoretically for $k \geq d$, any two $k$-colorings of a $d$-degenerate graph are Kempe-equivalent. 

In this study, we extend the definition of Kempe chain to include empty clusters: the set of Kempe chains $\mathbb{H}$ between a cluster $C_i$ and an empty cluster $C_j$ is the set of nodes in $C_i$.

\subsection{Kempe Swap $K$-Means}

Algorithm (\ref{alg:KCSKM}) $\mathrm{KSKM}$ adopts a $K$-Means–style iterative framework that progressively refines cluster assignments to locally minimize the within-cluster sum of squares. The primary driving mechanism is the Kempe chain swap move, implemented in Algorithm (\ref{alg:KSAssignment}) $\mathrm{KSAssignment}$, which identifies the steepest descent direction in the clustering objective without violating cannot-link constraints.

Beyond this core mechanism, $\mathrm{KSKM}$ integrates additional strategies to escape poor local optima. Hill-climbing procedures adjust the cluster centroids, either within the current neighborhood (Algorithm (\ref{alg:KSPerturb}) $\mathrm{KSPerturb}$) or along a new search direction (Algorithm (\ref{alg:KSShift}) $\mathrm{KSShift}$)—to further improve local optimality. 

The algorithm $\mathrm{ExactKSAssignment}$ includes an optional refinement step that improves a Kempe-swap–optimal solution by solving the fixed-centroid graph coloring problem (GCP) to optimality. Because this step is computationally expensive, it is triggered only when Kempe swaps can no longer improve the solution. When the number of colors $k$ is sufficiently larger than the chromatic number of the super-node graph, this refinement is generally unnecessary.

Moreover, when combined with good centroid initialization and DSATUR-inspired \citep{brelaz1979new} initial cluster assignment (Algorithm (\ref{alg:DSATURAssignment})), $\mathrm{KSKM}$ consistently yields high-quality local optima.

\begin{center}
\resizebox{0.9\linewidth}{!}{%
\begin{algorithm}[H]
\caption{Kempe Swap $K$-Means $\mathrm{KSKM}(\mathcal{ML}, \mathcal{CL}, y, k, L)$}
\label{alg:KCSKM}
\DontPrintSemicolon
\SetKwInOut{Input}{Input}\SetKwInOut{Output}{Output}
\Input{
Must-link constraints $\mathcal{ML}$;\\
Cannot-link constraints $\mathcal{CL}$;\\
Data $y$;\\
Number of clusters $k$;\\
Number of explorations $L$.
}
\tcp{Construct graph of super-nodes and cannot-link edges of super-nodes}
$V, E = \mathrm{preprocessing}(\mathcal{ML}, \mathcal{CL})$\;
\tcp{Initialize centroids}
$\mu = \mathrm{InitCentroids}(y,k)$\;
\BlankLine
\tcp{Initialize clustering with DSATUR Assignment to the nearest centroid}
$U = \mathrm{DSATURAssignment}(V, E, y, \mu, k)$\;
\tcp{Optional: Initialize clustering by solving GCP if DSATUR Assignment is infeasible}
\tcp{$U = \mathrm{ExactAssignment}(V, E, y, \mu)$}
\While{True}{
\For{$l=1,\cdots,L$}{
\While{Not Converged}{
\tcp{Update cluster centroids}
$\mu=\mathrm{Centroids}(U, y)$\;
\tcp{Improve cluster assignment by Kempe swaps}
$U = \mathrm{KSAssignment}(V, E, U, y, \mu)$\;
}
\tcp{Optional: Further improve cluster assignment by solving GCP with incumbent assignment as starting solution}
\tcp{$U = \mathrm{ExactAssignment}(V, E, U, y, \mu)$}
\eIf{$l \ \mathrm{mod} \ 5 = 0$}{
\tcp{Shift to relocated centroids}
$\hat{\mu}=\mathrm{CentroidRelocation}(y,U, \mu)$\;
$U = \mathrm{KSShift}(V, E, U, y, \hat{\mu})$\;
}
{
\tcp{Perturb centroids in neighborhood}
$U = \mathrm{KSPerturb}(V, E, U, y)$\;
}
Keep the best assignments $U^{*}$\;
}
}
\Output{Best assignments $U^{*}$}
\end{algorithm}
}
\end{center}

\subsubsection{Kempe Swap Cluster Assignment}

The procedure begins with the identification of Kempe chains $\mathbb{H}$ between the clusters $C_i$ and $C_j$, by partitioning $C_i \cup C_j \cap E$ into connected subgraphs. Each chain $H \in \mathbb{H}$ is then evaluated through the swap cost
\begin{align*}
 d_{H_i,H_j} &= \bigl(\|y_{H_i} - \vec{1}_{|H_i|}\mu_j^{\top}\|_F^2 - \|y_{H_i} - \vec{1}_{|H_i|}\mu_i^{\top}\|_F^2\bigr) 
+ \bigl(\|y_{H_j} - \vec{1}_{|H_j|}\mu_i^{\top}\|_F^2 - \|y_{H_j} - \vec{1}_{|H_j|}\mu_j^{\top}\|_F^2\bigr)
\\ & = (|H_i| - |H_j|)\bigl(\|\mu_j\|^2 - \|\mu_i\|^2\bigr)
- 2(|H_i|\,\bar y_{H_i}^{\top}- |H_j|\,\bar y_{H_j}^{\top})(\mu_j - \mu_i),
\end{align*}
where $H$ is split into $H_i$ and $H_j$ according to membership in $C_i$ and $C_j$, and $|H_i|$ and $|H_j|$ are the sizes of components that include all must-link data points in the super-nodes. Denote $S = \{(H_i,H_j): d_{H_i,H_j}<0, H \in \mathbb{H}\}$.

While each individual Kempe swap preserves feasibility, performing multiple swaps simultaneously can violate it. Preserving the $k$-coloring requires the swaps to satisfy two classes of constraints:
\begin{itemize}
 \item \textbf{Clique constraints $\mathcal{CC}^{swap}$ (\ref{eq:3a}):} a super-node can only take part in one swap in each assignment. 
 \item \textbf{Swap cannot-link constraints $\mathcal{CL}^{swap}$ (\ref{eq:3b}):} swaps that would result in cannot-link vertices in cluster assignments cannot occur simultaneously. 
\end{itemize}

To ensure rapid convergence, the algorithm greedily selects compatible swaps that improve the clustering objective with the steepest descent. The set of compatible swaps within one iteration can therefore be determined by solving a Maximum Weighted Independent Set (MWIS) problem:
\begin{subequations}\label{eq:3}
\begin{align}
 \max & \sum_{s \in S} -d_su_s \tag{\ref{eq:3}}\\
 \text{s.t. } & \sum_{s \in S : v \in s} u_s \leq 1 && \forall v \in V, \label{eq:3a}\\
 & u_s + u_t \leq 1 && \forall (s,t) \in \mathcal{CL}^{swap}, \label{eq:3b}\\
 & u_s \in \{0,1\} && \forall s \in S. \label{eq:3c}
\end{align}
\end{subequations}

Compared with the approaches proposed in \cite{baumann2020} and \cite{baumann2025algorithm}, the cluster-assignment step in our method requires solving a MWIS problem (\ref{eq:3}) rather than performing a full graph $k$-coloring. This MWIS problem is relatively inexpensive, since the cluster centroids do not shift substantially and thus only a small number of swaps with $d_H<0$ need to be considered.

For the rare case where the number of swaps is huge, fortunately we don't have to solve the MWIS problem (\ref{eq:3}) to optimality. A heuristic solution by \cite{hosseinian2018nonconvex} is introduced in \ref{app1}.

Algorithm (\ref{alg:KSAssignment}) $\mathrm{KSAssignment}$ improves current cluster assignment by performing Kempe swaps that leads to the steepest descent in objective function. 

\begin{center}
\resizebox{0.9\linewidth}{!}{%
\begin{algorithm}[H]
\caption{Kempe Swaps Assignment $\mathrm{KSAssignment}(V, E, U, y, \mu)$}
\label{alg:KSAssignment}
\DontPrintSemicolon
\SetKwInOut{Input}{Input}\SetKwInOut{Output}{Output}
\Input{
Graph of super-nodes and cannot-link edges $G = (V, E)$;\\
Data $\{y_i\}_{v_i \in V}$;\\
Cluster assignments $U = \{U(v_i)\}_{v_i \in V}$;\\
Cluster centroids $\mu$.
}

\BlankLine
\tcp{Find swaps}
$S = \{(H_i,H_j)\in \mathrm{KempeChains}(i,j,V,E):d_{i,j}<0, i,j \in \mathrm{clusters}(U)\}$\;
\tcp{Find clique constraints}
$\mathcal{CC}^{swap} = \{\{s|v\in H_i \cup H_j, s = (H_i,H_j)\in S \}:v\in V\}$\;
\tcp{Find cannot-link swaps}
$\mathcal{CL}^{swap} = \{(s,t), s,t \in S: \exists (v_i,v_j) \in E, U(v_i) = U(v_j), v_i \in s, v_j \in t\}$\;
\tcp{Solve Maximum Weighted Independent Set}
$u_{swap} = \mathrm{MWIS}(S, \mathcal{CC}^{swap}, \mathcal{CL}^{swap})$\;
\tcp{Update cluster assignments}
$U = \mathrm{update}(U,u_{swap})$\;
\Output{Improved cluster assignment $U$}
\end{algorithm}
}
\end{center}

Even if the MWIS subproblem is solved to optimality, cluster assignment based on Kempe swaps cannot guarantee convergence to the global optimum:
\begin{itemize}
 \item Kempe swaps between two clusters are applied greedily. Improvements that require coordinated multi-cluster Kempe swaps may therefore be missed, even if each pairwise swap is locally optimal.
 \item When the number of clusters is close to the graph chromatic number, Kempe-chain exchanges become global, a single chain would span almost all vertices within two clusters, and graph $k$-colorings are often not Kempe-equivalent. In such cases, incorporating an additional exact assignment step for the fixed-centroid graph coloring problem could help the algorithm escape to a solution neighborhood that is not Kempe-equivalent to the current one. It is worth noting that the GCP does not need to be solved to optimality. As long as an improved solution is obtained, the current Kempe-swap local optimum is escaped, allowing subsequent Kempe swaps to further drive the solution toward a better local optimum.
\end{itemize}
Details on multi-cluster Kempe swaps and the GCP can be found in \ref{app2} and \ref{app3}.

\subsubsection{Hill-climbing Centroid Mutation}
A practical way to escape local optima is to permit occasional hill-climbing moves.
In the context of centroid-based methods for constrained clustering,
these moves correspond to perturbations of the cluster centers.
In this paper, we consider two controlled forms of centroid mutation:
\begin{enumerate}
 \item \textbf{Centroid perturbation} with small hill-climbs, and
 \item \textbf{Centroid reposition} with larger hill-climbs.
\end{enumerate}

In centroid-based iterative methods, a local optimum occurs when no further
assignment changes can improve the objective under the current centroids.
Such optima are highly sensitive to the centroids' locations:
small perturbations may not worsen the objective significantly,
but can unlock new improving moves.

For data within a cluster, suppose $y_v\overset{\text{i.i.d.}}{\sim} \mathcal{N}_p(\mu, \Sigma)$, $v \in C$. Denote sample mean and covariance by $\bar{y}$ and $\bar{\Sigma}$, then 
\begin{align*}
|C|(\bar{y} - \mu)^{\top} \bar{\Sigma}^{-1} (\bar{y} - \mu)
\sim \frac{p(|C|-1)}{|C|-p} F_{p,|C|-p}.
\end{align*}

Placing the noninformative Jeffreys prior, we can sample a plausible $\mu$ given the data $\bar{y}$ from multivariate $t$ distribution:
\begin{align}\label{eq:centroid_perturb}
 \mu \;\sim\;
t_{|C|-1}^{(p)}\!\left( \bar{y},\ \frac{1}{|C|} \bar{\Sigma} \right).
\end{align}

Note that neither the normal distribution nor the $t$-distribution is introduced for the purpose of statistical modeling; rather, both are employed within a data-driven procedure to sample a perturbed centroid that does not drastically degrade the objective function. By (\ref{eq:centroid_perturb}), clusters with larger mean-squared error undergo stronger perturbations, encouraging exploration where fitting quality is lower.
The step of perturbation is triggered whenever Algorithm (\ref{alg:KCSKM}) $\mathrm{KSKM}$ reaches a local optimum.

\begin{center}
\resizebox{0.9\linewidth}{!}{%
\begin{algorithm}[H]
\caption{Neighborhood perturbation $\mathrm{KSPerturb}(V, E, U, y)$}
\label{alg:KSPerturb}
\DontPrintSemicolon
\SetKwInOut{Input}{Input}\SetKwInOut{Output}{Output}
\Input{
Graph of super-nodes and cannot-link edges $G = (V, E)$;\\
Data $\{y_i\}_{v_i \in V}$;\\
Cluster assignments $U = \{U(v_i)\}_{v_i \in V}$.
}
\BlankLine
\tcp{Form clusters from membership}
$C = \mathrm{Clusters}(U)$\;
\tcp{Perturb centroids}
$\mu_i \;\sim\;
t_{|C_i|-1}^{(p)}\!\left( \bar{y}_{C_i},\ \frac{1}{|C_i|} \bar{\Sigma}_{C_i} \right), i \in [k]$\;
\tcp{Adjust cluster assignment to the perturbed centroids}
$U = \mathrm{KSAssignment}(V, E, U, y, \mu)$\;
\Output{Cluster assignments $U = \{U(v_i)\}_{v_i \in V}$}
\end{algorithm}
}
\end{center}

While neighborhood perturbation explores locally, centroid reposition allows centroids to jump outside their current basin of attraction.
Shifted centroids \(\{\hat{\mu}_i\}_{i=1}^k\) can be chosen by:
\begin{itemize}
 \item solving an unconstrained clustering solution,
 \item solving the relaxed SDP \citep{piccialli2022exact},
 \item reposition one cluster centroid \citep{baumann2025algorithm},
 \item randomly removing one centroid and substituting by a data point \citep{merz2003iterated,mansueto2025efficiently}.
\end{itemize}
Depending on how target centroids are obtained, we may need to match new centroids the current clusters via a classical Assignment Problem.
\begin{align*}
 \min_{z} & \sum_{i=1}^{k}\sum_{j=1}^{k} z_{i,j}\sum_{v\in C_i}||y_v-\hat{\mu}_j||^2
 \\ \text{s.t. }& \sum_{j = 1}^{k} z_{i,j} = 1 && \forall i \in [k],
 \\ & \sum_{i = 1}^{k} z_{i,j} = 1 && \forall j \in [k],
 \\ & z_{i,j} \in \{0,1\} && \forall i \in [k], \quad j \in [k],
\end{align*}
where the decision variable $z_{i,j}$ indicates whether the current cluster $C_i$ matches the target centroid $\hat{\mu}_j$.

The shifted centroid is then
\begin{align*}
 \mu_i\leftarrow(1-\alpha)\mu_i + \alpha \sum_{j = 1}^{k} z_{i,j} \hat{\mu}_j, \forall i \in [k],
\end{align*}
where \(\alpha \in (0,1]\) controls the step size.
Centroid repositions are applied only if repeated perturbations fail to improve the objective.

\begin{center}
\resizebox{0.9\linewidth}{!}{%
\begin{algorithm}[H]
\caption{Centroid reposition $\mathrm{KSShift}(V, E, U, y, \hat{\mu}, \alpha)$}
\label{alg:KSShift}
\DontPrintSemicolon
\SetKwInOut{Input}{Input}\SetKwInOut{Output}{Output}
\Input{
Graph of super-nodes and cannot-link edges $G = (V, E)$;\\
Data $\{y_i\}_{v_i \in V}$;\\
Cluster assignments $U = \{U(v_i)\}_{v_i \in V}$;\\
Target centroids $\hat{\mu}$;\\
Step size $\alpha$.
}
\BlankLine
\tcp{Match current clusters with target centroids}
$z = \mathrm{Assign}(U,\hat{\mu})$\;
\tcp{Shift centroids}
$\mu_i\leftarrow(1-\alpha)\mu_i + \alpha \sum_{j = 1}^{k} z_{i,j} \hat{\mu}_j, \forall i \in [k]$\;
\While{Not Converged}{
\tcp{Adjust cluster assignment to the shifted centroids}
$U = \mathrm{KSAssignment}(V, E, U, y, \mu)$\;
}
\Output{Cluster assignments $U = \{U(v_i)\}_{v_i \in V}$}
\end{algorithm}
}
\end{center}

Both neighborhood perturbation and centroid reposition are followed by cluster adjustment based on Kempe swaps. This ensures that every centroid update is coupled with a corresponding reassignment of points, maintaining coherence between the two phases of the algorithm. 

\subsubsection{Cluster Initialization based on DSATUR}
The premise behind performing Kempe-chain exchanges is the availability of a feasible $k$-coloring. A natural companion to Kempe swaps is the DSATUR algorithm \cite{brelaz1979new}, one of the most effective heuristics for finding smallest coloring of a graph. If $k$ exceeds the DSATUR chromatic number of $(V, E)$, then, with the extension of Kempe chains to include empty sets, Kempe swaps can populate empty color classes and continue improving the solution until a local optimum is reached.

The key idea of DSATUR is the dynamic sequential ordering of vertices by their degree of saturation, which counts the number of distinct clusters to which a vertex is adjacent. Ties are broken by vertex degree, i.e., the number of incident edges. Each selected vertex is then assigned to the smallest-index permissible color class. In this way, DSATUR tends to reuse color classes with smaller indices, making them increasingly constrained.

We adopt an even greedier assignment rule (Algorithm \ref{alg:DSATURAssignment}): vertices are processed in DSATUR order—that is, in non-increasing order of saturation degree—but each vertex is assigned to the nearest feasible centroid, rather than to the first feasible cluster or the one using the smallest color. 

Prioritizing vertices with high saturation ensures that highly constrained points are settled early, when many feasible clusters remain available. This DSATUR-style ordering significantly reduces the likelihood that a late-stage vertex has no admissible cluster due to previously assigned neighbors or feasibility restrictions.

Compared with classical DSATUR coloring, the proposed variant assigns vertices to their nearest centroids, aiming to produce a starting solution with a lower clustering objective. In addition, it supports random centroid initialization, which is advantageous for Kempe swap operations. Because the solution space of $k$-colorings is not necessarily Kempe-equivalent, introducing randomness helps promote exploration of a larger and more diverse solution space.

\begin{center}
\resizebox{0.9\linewidth}{!}{%
\begin{algorithm}[H]
\caption{DSATUR Assignment $\mathrm{DSATURAssignment}(V, E, y, \mu, k)$}
\label{alg:DSATURAssignment}
\DontPrintSemicolon
\SetKwInOut{Input}{Input}\SetKwInOut{Output}{Output}
\Input{
Graph of super-nodes and cannot-link edges $G = (V, E)$;\\
Data $y$;\\
Centroids $\mu$;\\
Number of clusters $k$.
}
\BlankLine
\tcp{Initiate an empty set of membership}
$U = \{\}$\;
\While{$|U| < |V|$}{
\tcp{Assign the remaining super-node with maximum degree of saturation to the nearest centroid}
$v = \text{argmax}(\mathrm{DegreeSaturation}(V,U))$\;
\For{$i=1\cdots k$}{
$distance = \infty$\;
\If{not $\mathrm{violation}(v,i,U,E)$ and $||y_v-\mu_i||< distance$}{
$distance = ||y_v-\mu_i||$\;
$U(v) = i$\;
}
\If{$distance = \infty$}{
\tcp{DSATUR Assignment to the nearest centroid is infeasible, assign with classic DSATUR instead}
$U = \mathrm{DSATUR}(V,E)$\;
\eIf{$\max(U) \leq k$}{\Output{Cluster assignments $U = \{U(v_i)\}_{v_i \in V}$}}
{\Output{Terminate with infeasibility}}
}
}
}
\tcp{Further shifting assignment to the centroids}
$U = \mathrm{KSShift}(V, E, U, y, \mu)$\;
\Output{Cluster assignments $U = \{U(v_i)\}_{v_i \in V}$}
\end{algorithm}
}
\end{center}

By repeating the step of DSATUR Assignment line 1 to line 8 and the step centroid update, we also proposed another heuristic DSATUR COP-$K$-Means. Details are included in \ref{app4}.

Note that the proposed DSATUR Assignment rule—processing vertices in DSATUR order but assigning each vertex to its nearest feasible centroid—may fail to produce a feasible $k$-coloring even in cases where the classic DSATUR algorithm would successfully color the graph using strictly fewer than $k$ colors.

The reason is that, under the proposed method, clusters are no longer used in increasing index order, and therefore the small-index clusters are not necessarily the most constrained. Assigning vertices to their nearest centroids tends to concentrate similar vertices together, which increases local conflicts and causes saturation degrees to rise more rapidly than in classic DSATUR. Consequently, although DSATUR ordering still ensures that the most constrained vertex is treated first, the overall constraint environment becomes dynamically tighter, potentially eliminating feasible clusters for some later vertex.

When the case occurs, lines 10–12 are triggered, and the algorithm falls back to a classic DSATUR coloring to produce an initial feasible assignment. In practice, the opposite scenario can also occur: the classical DSATUR algorithm may fail to find a feasible $k$-coloring, whereas the proposed method succeeds. If both methods are infeasible under the clustering constraints, the algorithm invokes an GCP-based assignment procedure to initialize the clustering.


\newpage
\section{Numerical Experiments}
\label{sec3}
\subsection{Data Description}
We consider four collections of data sets that are commonly used in related work.

\begin{itemize}
\item \textbf{Data collection 1} \citep{gonzalez2020dils} contains 25 real and synthetic data sets, with the number of data points ranging from 47 to 846, dimensionality from 2 to 90, and ground-truth numbers of clusters from 2 to 15.
\item \textbf{Data collection 2} \citep{baumann2025algorithm} consists of 15 synthetic data sets, each comprising 300 two-dimensional data points with ground-truth clusters between 10 and 50.
\item \textbf{Data collection 3} \citep{baumann2025algorithm} includes 24 synthetic two-dimensional data sets, with the number of data points ranging from 500 to 5000 and ground-truth clusters from 2 to 100.
\item \textbf{Data collection 4} \citep{baumann2025algorithm} comprises 6 real and synthetic data sets, with the number of data points ranging from 5300 to 70{,}000, dimensionality from 2 to 3072, and ground-truth clusters from 2 to 100.
\end{itemize}

Constraints are generated using the method proposed by \citet{wagstaff2001constrained} and implemented in \citet{gonzalez2020dils,baumann2025algorithm}. The constraint level denotes the percentage of Must-link and Cannot-link constraints relative to the $\frac{n(n-1)}{2}$ possible data pairs. Note that the ratio of Must-link to Cannot-link constraints depends on the number of clusters and therefore varies across data sets. The percentage of constraints is not a direct measure of clustering complexity, as must-link constraints reduce the problem difficulty. Datasets with a higher proportion of cannot-link constraints over supernodes are considered \emph{dense} and are generally harder to cluster.

\begin{itemize}
\item \textbf{Data collection 1}: small-scale; 20\% of constraints generated.
\item \textbf{Data collection 2}: small-scale but dense; 50\% of constraints generated.
\item \textbf{Data collection 3}: medium-scale; 20\% of constraints generated.
\item \textbf{Data collection 4}: large-scale; 20\% of constraints generated.
\end{itemize}

All experimental evaluations were carried out on a computer system comprising an AMD Ryzen 7 9800X3D 8-core processor at 4.70 GHz and 32 GB of DDR5 memory running at 6000 MT/s.

For a fair comparison, we include in the comparison methods that will not violate the must-link and cannot-link constraints, excluding methods with soft assignment. All methods were executed in a Python 3.13 environment using GurobiPy 12.0.

\subsection{Overall Comparison}
\label{OverallComparison}
We begin by describing the state-of-the-art methods included in the numerical experiments:

\begin{itemize}
\item \textbf{KSKM}: The proposed Kempe Swap $K$-Means method (Algorithm \ref{alg:KCSKM}) without the exact assignment step involving GCP. The maximum number of mutations is set to 200. Throughout the numerical experiments, mutations comprise 80\% centroid perturbations (Algorithm~\ref{alg:KSPerturb}) and 20\% repositions (Algorithm~\ref{alg:KSShift}).
\item \textbf{KSKM\_E}: The proposed Kempe Swap $K$-Means Extension method (Algorithm \ref{alg:KCSKM}), where the assignment step based on GCP is enabled. Rather than seeking the optimal solution of the GCP, the solver is terminated once the first improving solution is found, starting from a Kempe-swap–optimal solution. Then Kempe swaps will keep improving solution until a new local optima is reached. In this setting, the GCP is not used as an exact optimization tool; instead, it serves as a mechanism to escape local optima induced by Kempe swaps. As a result, it often enables the current solution to transition to a new solution that is not Kempe-equivalent. The maximum number of mutations is set to 50.

\item \textbf{PCCC}: A state-of-the-art iterative integer programming approach \citep{baumann2025algorithm} that considers only hard constraints. Cluster repositioning is enabled, and the model-size reduction technique is applied with $q=10$.

\item \textbf{BLPKMCC}: A state-of-the-art iterative integer programming method \citep{baumann2020} using default settings.

\item \textbf{COPKM}: A classical COP-$K$-means method \citep{wagstaff2001constrained}, where the assignment of points follows the order of the data.

\item \textbf{COPKM\_P}: The improved COP-$K$-Means method that employs DSATUR-based assignment (Algorithm \ref{alg:DSATURKM}).

\item \textbf{S\_MDEClust}: A state-of-the-art method based on memetic differential evolution combined with iterative integer programming \citep{mansueto2025efficiently}. The number of populations is set to 10. We report the results obtained with this method only in Appendix~\ref{app:smde}, as it focuses on improving clustering inertia through extensive solution exploration, which significantly increases computational time and therefore restricts its practical use to small-scale datasets.
\end{itemize}

\begin{table}[H]
\caption{State-of-the-art Methods for Semi-Supervised Clustering}
\label{tab:methods}
\centering
\resizebox{0.9\textwidth}{!}{%
\begin{tabular}{c|cc|cc|ccc}
\toprule
            & \multicolumn{2}{c|}{Initialization} & \multicolumn{2}{c|}{Assignment}                & \multicolumn{3}{c}{Solution Mutation and Crossover}              \\
            & Greedy Method         & GCP        & Greedy Method                           & GCP & Centroid Perturbation & Centroid Reposition & Centroid Crossover \\ \midrule
KSKM        & DSATUR-based          & No         & Kempe Swaps                             & No  & Yes                   & Yes                 & No                 \\
KSKM\_E     & DSATUR-based          & Yes        & Kempe Swaps                             & Yes & Yes                   & Yes                 & No                 \\
PCCC        & No                    & Yes        & No                                      & Yes & No                    & Yes                 & No                 \\
BLPKMCC     & No                    & Yes        & No                                      & Yes & No                    & No                  & No                 \\
S\_MDEClust & Yes                   & Yes        & By order of data                        & Yes & No                    & Optional            & Yes                \\
COPKM       & Yes                   & No         & By order of data                        & No  & No                    & No                  & No                 \\
COPKM\_P    & DSATUR-based          & No         & By order of DSATUR & No  & No                    & No                  & No                 \\ \bottomrule
\end{tabular}
}
\end{table}
Each method is executed 10 times with a different random initialization in each run. For fairness of comparison, the same randomly initialized centroids are used across all methods. We report the average clustering objective (inertia) in Table~\ref{tab:obj} and the average CPU runtime in Table~\ref{tab:time}, both normalized relative to the baseline method (KSKM) in all datasets; lower values indicate better performance. In addition, the average success rate is reported in Table~\ref{tab:rate_success}.

We also report detailed results for dataset collection 4 in Tables~\ref{tab:rate_success_COL4}, \ref{tab:obj_COL4}, and \ref{tab:time_COL4}, as the behavior of the algorithms becomes more complex on larger datasets. Other detailed results, including average clustering inertia, CPU runtime, Adjusted Rand Index (ARI), and success rate for each individual dataset, are provided in the supplementary document.

We begin by comparing the rate of success runs. Each candidate algorithm is allowed to run for 3600 seconds. If no incumbent solution is returned at the end of the runs, we record failure of the runs. Most candidate algorithms have no problem of success runs except for the classic COP-$K$-Means algorithm, which struggles to find a feasible solution due to cannot-link constraints. Such a difficulty is alleviated by assigning data points in DSATUR sequence to the closet centroids. The proposed COPKM\_P has no problem of success runs except for dataset collection 2 with 20\% constraints, where it's rate of success is 0.93. For such a case, KSKM is instead initialized by classic DSATUR coloring to find a feasible starting solution. The GCP-based methods have no problem in finding feasible solutions, except for large cases in dataset collection 3 and 4 when the percentage of constraints is large and the algorithm either cannot find a feasible solution at the end of the runs or run out of memory. There are very rare cases where PCCC experiences an error in 'Weights sum to zero' in clustering reposition in dataset collection 2, which may result from removal of soft constraints in our setting. 
\begin{table}[H]
\caption{Algorithm successful rate throughout test.}
\label{tab:rate_success}
\centering
\resizebox{0.9\textwidth}{!}{%
\begin{tabular}{cc|cccccc}
\toprule
 & Method & KSKM & KSKM\_E & PCCC & BLPKMCC & COPKM\_P & COPKM \\
Dataset & Constraint Level &  &  &  &  &  &  \\
\midrule
\multirow[t]{4}{*}{COL1} & 0.05 & 1.00 & 1.00 & 1.00 & 1.00 & 1.00 & 0.47 \\
 & 0.1 & 1.00 & 1.00 & 1.00 & 1.00 & 1.00 & 0.40 \\
 & 0.15 & 1.00 & 1.00 & 1.00 & 1.00 & 1.00 & 0.63 \\
 & 0.2 & 1.00 & 1.00 & 1.00 & 1.00 & 1.00 & 0.64 \\
\cline{1-8}
\multirow[t]{10}{*}{COL2} & 0.05 & 1.00 & 1.00 & 1.00 & 1.00 & 1.00 & 1.00 \\
 & 0.1 & 1.00 & 1.00 & 1.00 & 1.00 & 1.00 & 1.00 \\
 & 0.15 & 1.00 & 1.00 & 1.00 & 1.00 & 1.00 & 1.00 \\
 & 0.2 & 1.00 & 1.00 & 1.00 & 1.00 & 0.93 & 0.81 \\
 & 0.25 & 1.00 & 1.00 & 1.00 & 1.00 & 1.00 & 0.91 \\
 & 0.3 & 1.00 & 1.00 & 1.00 & 1.00 & 1.00 & 0.69 \\
 & 0.35 & 1.00 & 1.00 & 0.99 & 1.00 & 1.00 & 0.75 \\
 & 0.4 & 1.00 & 1.00 & 0.99 & 1.00 & 1.00 & 0.99 \\
 & 0.45 & 1.00 & 1.00 & 1.00 & 1.00 & 1.00 & 1.00 \\
 & 0.5 & 1.00 & 1.00 & 0.99 & 1.00 & 1.00 & 0.99 \\
\cline{1-8}
\multirow[t]{4}{*}{COL3} & 0.05 & 1.00 & 1.00 & 1.00 & 1.00 & 1.00 & 0.88 \\
 & 0.1 & 1.00 & 1.00 & 1.00 & 0.97 & 1.00 & 0.83 \\
 & 0.15 & 1.00 & 1.00 & 0.98 & 0.99 & 1.00 & 0.79 \\
 & 0.2 & 1.00 & 1.00 & 1.00 & 1.00 & 1.00 & 0.86 \\
\cline{1-8}
\multirow[t]{3}{*}{COL4} & 0.005 & 1.00 & 1.00 & 1.00 & 1.00 & 1.00 & 0.83 \\
 & 0.01 & 1.00 & 1.00 & 0.92 & 0.78 & 1.00 & 0.50 \\
 & 0.05 & 1.00 & 1.00 & 0.83 & 0.83 & 1.00 & 0.67 \\
\cline{1-8}
\bottomrule
\end{tabular}
}
\end{table}

Table~\ref{tab:rate_success_COL4} reports the success rates of the runs on the largest dataset, collection 4. For the $\mathrm{CIFAR100}$ dataset, the GCP-based methods fail due to memory exhaustion at constraint levels of 1\% and 5\%. However, KSKM\_E is still able to return a valid solution, as it alternates between improving the solution through Kempe swaps and GCP.
\begin{table}[H]
\caption{Algorithm success rate among dataset collection 4}
\label{tab:rate_success_COL4}
\centering
\resizebox{0.9\textwidth}{!}{%
\begin{tabular}{cc|cccccc}
\toprule
 & Method & KSKM & KSKM\_E & PCCC & BLPKMCC & COPKM\_P & COPKM \\
Dataset & Constraints Level &  &  &  &  &  &  \\
\midrule
\multirow[t]{3}{*}{banana} & 0.005 & 1.00 & 1.00 & 1.00 & 1.00 & 1.00 & 0.00 \\
 & 0.01 & 1.00 & 1.00 & 1.00 & 1.00 & 1.00 & 0.00 \\
 & 0.05 & 1.00 & 1.00 & 1.00 & 1.00 & 1.00 & 1.00 \\
\cline{1-8}
\multirow[t]{3}{*}{cifar10} & 0.005 & 1.00 & 1.00 & 1.00 & 1.00 & 1.00 & 1.00 \\
 & 0.01 & 1.00 & 1.00 & 1.00 & 1.00 & 1.00 & 0.00 \\
 & 0.05 & 1.00 & 1.00 & 1.00 & 1.00 & 1.00 & 1.00 \\
\cline{1-8}
\multirow[t]{3}{*}{cifar100} & 0.005 & 1.00 & 1.00 & 1.00 & 1.00 & 1.00 & 1.00 \\
 & 0.01 & 1.00 & 1.00 & 0.50 & 0.50 & 1.00 & 1.00 \\
 & 0.05 & 1.00 & 1.00 & 0.00 & 0.00 & 1.00 & 0.00 \\
\cline{1-8}
\multirow[t]{3}{*}{letter} & 0.005 & 1.00 & 1.00 & 1.00 & 1.00 & 1.00 & 1.00 \\
 & 0.01 & 1.00 & 1.00 & 1.00 & 1.00 & 1.00 & 1.00 \\
 & 0.05 & 1.00 & 1.00 & 1.00 & 1.00 & 1.00 & 0.00 \\
\cline{1-8}
\multirow[t]{3}{*}{mnist} & 0.005 & 1.00 & 1.00 & 1.00 & 1.00 & 1.00 & 1.00 \\
 & 0.01 & 1.00 & 1.00 & 1.00 & 0.20 & 1.00 & 0.00 \\
 & 0.05 & 1.00 & 1.00 & 1.00 & 1.00 & 1.00 & 1.00 \\
\cline{1-8}
\multirow[t]{3}{*}{shuttle} & 0.005 & 1.00 & 1.00 & 1.00 & 1.00 & 1.00 & 1.00 \\
 & 0.01 & 1.00 & 1.00 & 1.00 & 1.00 & 1.00 & 1.00 \\
 & 0.05 & 1.00 & 1.00 & 1.00 & 1.00 & 1.00 & 1.00 \\
\cline{1-8}
\bottomrule
\end{tabular}
}
\end{table}

Clustering inertia, defined as the within-cluster sum of squares, is one of the most important criteria in our experimental setting. Unlike the Adjusted Rand Index (ARI), inertia directly measures an algorithm’s ability to find a global optimal solution, independent of ground-truth labels.

COP-$K$-Means methods and BLPKMCC tend to yield higher inertia values, as these methods lack mechanisms for exploring the solution neighborhood, such as cluster repositioning or centroid perturbation. Notably, KSKM consistently demonstrates the ability to find near–globally optimal solutions at a level comparable to KSKM\_E and PCCC across all datasets, approaching the global optimality reported in \cite{baumann2025algorithm}. 
\begin{table}[H]
\caption{Comparison of clustering inertia, reported as multiples of the baseline method (KSKM) across datasets among success runs. Lower inertia multiples indicate better performance.}
\label{tab:obj}
\centering
\resizebox{0.9\textwidth}{!}{%
\begin{tabular}{cc|cccccc}
\toprule
 & Method & KSKM & KSKM\_E & PCCC & BLPKMCC & COPKM\_P & COPKM \\
Dataset & Constraint Level &  &  &  &  &  &  \\
\midrule
\multirow[t]{4}{*}{COL1} & 0.05 & 1.00 & 1.00 & 1.01 & 1.04 & 1.03 & 1.05 \\
 & 0.1 & 1.00 & 1.00 & 1.00 & 1.03 & 1.02 & 1.04 \\
 & 0.15 & 1.00 & 1.00 & 1.00 & 1.02 & 1.01 & 1.02 \\
 & 0.2 & 1.00 & 1.00 & 1.00 & 1.01 & 1.00 & 1.00 \\
\cline{1-8}
\multirow[t]{10}{*}{COL2} & 0.05 & 1.00 & 1.00 & 1.04 & 1.31 & 1.10 & 1.11 \\
 & 0.1 & 1.00 & 1.00 & 1.05 & 1.37 & 1.14 & 1.17 \\
 & 0.15 & 1.00 & 1.00 & 1.02 & 1.33 & 1.17 & 1.17 \\
 & 0.2 & 1.00 & 1.00 & 1.02 & 1.20 & 1.15 & 1.15 \\
 & 0.25 & 1.00 & 0.99 & 1.04 & 1.22 & 1.17 & 1.23 \\
 & 0.3 & 1.00 & 0.98 & 1.00 & 1.11 & 1.07 & 1.09 \\
 & 0.35 & 1.00 & 1.01 & 1.02 & 1.06 & 1.05 & 1.08 \\
 & 0.4 & 1.00 & 1.00 & 1.02 & 1.08 & 1.06 & 1.11 \\
 & 0.45 & 1.00 & 1.00 & 1.02 & 1.07 & 1.05 & 1.08 \\
 & 0.5 & 1.00 & 1.00 & 1.03 & 1.07 & 1.07 & 1.10 \\
\cline{1-8}
\multirow[t]{4}{*}{COL3} & 0.05 & 1.00 & 1.00 & 1.02 & 1.10 & 1.12 & 1.11 \\
 & 0.1 & 1.00 & 0.98 & 1.00 & 1.10 & 1.05 & 1.10 \\
 & 0.15 & 1.00 & 0.99 & 1.00 & 1.08 & 1.04 & 1.05 \\
 & 0.2 & 1.00 & 1.00 & 1.01 & 1.03 & 1.03 & 1.06 \\
\cline{1-8}
\multirow[t]{3}{*}{COL4} & 0.005 & 1.00 & 1.00 & 1.00 & 1.01 & 1.01 & 1.01 \\
 & 0.01 & 1.00 & 0.99 & 1.02 & 1.03 & 1.00 & 1.01 \\
 & 0.05 & 1.00 & 1.00 & 1.00 & 1.00 & 1.00 & 1.00 \\
\cline{1-8}
\bottomrule
\end{tabular}
}
\end{table}

For the largest datasets, the GCP-based methods achieve higher clustering inertia than KSKM and COP-$K$-Means, as shown in Table~\ref{tab:obj_COL4}. Together with the results in Table~\ref{tab:time_COL4}, this indicates that the GCP-based methods are unable to complete the sequence of exact cluster-assignment problems~(\ref{eq:7}) within the 3600-second time limit.
\begin{table}[H]
\caption{Algorithm clustering inertia among dataset collection 4}
\label{tab:obj_COL4}
\centering
\resizebox{0.9\textwidth}{!}{%
\begin{tabular}{cc|cccccc}
\toprule
 & Method & KSKM & KSKM\_E & PCCC & BLPKMCC & COPKM\_P & COPKM \\
Dataset & Constraints Level &  &  &  &  &  &  \\
\midrule
\multirow[t]{3}{*}{banana} & 0.005 & 6482.97 & 6482.97 & 6486.67 & 6482.98 & 6519.10 & - \\
 & 0.01 & 7566.29 & 7566.29 & 7567.65 & 7566.29 & 7707.22 & - \\
 & 0.05 & 10556.11 & 10556.11 & 10556.11 & 10556.11 & 10556.11 & 10556.11 \\
\cline{1-8}
\multirow[t]{3}{*}{cifar10} & 0.005 & 126770854.91 & 126492578.88 & 126612825.71 & 126519249.09 & 127530232.16 & 127579490.59 \\
 & 0.01 & 147887573.71 & 147457869.95 & 157005767.08 & 160655672.87 & 147220022.93 & - \\
 & 0.05 & 172608323.88 & 172608323.88 & 172608323.88 & 172608323.88 & 172608323.88 & 172608323.88 \\
\cline{1-8}
\multirow[t]{3}{*}{cifar100} & 0.005 & 90713888.95 & 90701179.10 & 91012502.35 & 91217697.77 & 90808595.10 & 90817932.44 \\
 & 0.01 & 92748064.89 & 92752048.21 & 93640143.35 & 94537010.08 & 93006014.31 & 92934054.98 \\
 & 0.05 & 147067151.28 & 147074949.64 & - & - & 147181909.52 & - \\
\cline{1-8}
\multirow[t]{3}{*}{letter} & 0.005 & 123150.01 & 123194.49 & 123658.15 & 124061.70 & 124107.98 & 124320.11 \\
 & 0.01 & 129305.07 & 128946.60 & 129235.82 & 129625.72 & 130686.28 & 130623.89 \\
 & 0.05 & 208141.24 & 208124.29 & 208232.38 & 208124.32 & 208187.97 & - \\
\cline{1-8}
\multirow[t]{3}{*}{mnist} & 0.005 & 43212761.15 & 43179496.08 & 43263288.23 & 43184194.94 & 43436554.98 & 43437504.01 \\
 & 0.01 & 46121516.68 & 44644860.85 & 48542487.99 & 48776943.77 & 44631657.08 & - \\
 & 0.05 & 44514740.31 & 44514740.31 & 44514740.31 & 44514740.31 & 44514740.31 & 44514740.31 \\
\cline{1-8}
\multirow[t]{3}{*}{shuttle} & 0.005 & 293608.03 & 293144.76 & 299481.92 & 305160.20 & 296049.18 & 296829.94 \\
 & 0.01 & 328046.93 & 327845.18 & 331753.88 & 338035.86 & 330841.59 & 332760.92 \\
 & 0.05 & 366680.49 & 366680.49 & 367244.67 & 366760.67 & 367447.69 & 367716.94 \\
\cline{1-8}
\bottomrule
\end{tabular}
}
\end{table}

In terms of runtime, the COP-$K$-Means methods—both the classical version and the enhanced variant with DSATUR assignment—are the fastest algorithms. Their efficiency stems from the polynomial-time cluster assignment step (when feasible) and the absence of solution mutation. BLPKMCC also considers only a single solution neighborhood and does not perform mutation; however, its runtime is among the largest due to the exact GCP-based assignment~(\ref{eq:7}). Although KSKM\_E and PCCC also explore multiple neighborhoods, their runtimes are comparable to that of BLPKMCC. This is explained by their algorithmic design: PCCC leverages a reduced-size formulation by restricting assignments to the nearest $q$ clusters, while KSKM\_E accelerates neighborhood exploration by applying Kempe swaps, thereby reducing the cost of solving the GCP. Finally, KSKM, which relies on Kempe swaps instead of exact GCP-based assignment and cluster mutation, is significantly faster than BLPKMCC, KSKM\_E, and PCCC, while still achieving near-optimal solutions. The reduction in runtime scales with the size of the data (for the larger datasets in collections 3 and 4, the GCP-based methods reach the time limit and terminate early).
\begin{table}[H]
\caption{Comparison of CPU time, reported as multiples of the baseline method (KSKM) across datasets among success runs. Smaller time multiples indicate faster performance.}
\label{tab:time}
\centering
\resizebox{0.9\textwidth}{!}{%
\begin{tabular}{cc|cccccc}
\toprule
 & Method & KSKM & KSKM\_E & PCCC & BLPKMCC & COPKM\_P & COPKM \\
Dataset & Constraint Level &  &  &  &  &  &  \\
\midrule
\multirow[t]{4}{*}{COL1} & 0.05 & 1.00 & 2.69 & 10.49 & 4.35 & 0.41 & 2.45 \\
 & 0.1 & 1.00 & 3.31 & 10.23 & 10.28 & 0.73 & 0.44 \\
 & 0.15 & 1.00 & 3.51 & 10.17 & 24.46 & 0.62 & 0.69 \\
 & 0.2 & 1.00 & 3.52 & 11.79 & 42.85 & 0.69 & 0.87 \\
\cline{1-8}
\multirow[t]{10}{*}{COL2} & 0.05 & 1.00 & 8.24 & 11.72 & 5.65 & 0.46 & 0.79 \\
 & 0.1 & 1.00 & 16.14 & 23.23 & 10.42 & 0.09 & 0.13 \\
 & 0.15 & 1.00 & 24.79 & 45.89 & 20.03 & 0.10 & 0.14 \\
 & 0.2 & 1.00 & 33.55 & 58.36 & 33.13 & 0.11 & 0.13 \\
 & 0.25 & 1.00 & 49.66 & 77.43 & 52.00 & 0.11 & 0.14 \\
 & 0.3 & 1.00 & 50.69 & 81.96 & 73.33 & 0.15 & 0.20 \\
 & 0.35 & 1.00 & 52.42 & 73.53 & 100.90 & 0.19 & 0.22 \\
 & 0.4 & 1.00 & 70.14 & 80.79 & 119.36 & 0.21 & 0.21 \\
 & 0.45 & 1.00 & 71.25 & 82.06 & 152.90 & 0.23 & 0.24 \\
 & 0.5 & 1.00 & 33.76 & 58.74 & 163.37 & 0.25 & 0.27 \\
\cline{1-8}
\multirow[t]{4}{*}{COL3} & 0.05 & 1.00 & 29.20 & 59.25 & 47.01 & 0.34 & 0.40 \\
 & 0.1 & 1.00 & 68.28 & 84.75 & 129.45 & 0.40 & 0.44 \\
 & 0.15 & 1.00 & 87.68 & 94.75 & 197.38 & 0.42 & 0.51 \\
 & 0.2 & 1.00 & 120.86 & 113.58 & 314.56 & 0.48 & 0.52 \\
\cline{1-8}
\multirow[t]{3}{*}{COL4} & 0.005 & 1.00 & 4.68 & 31.17 & 24.41 & 1.22 & 0.29 \\
 & 0.01 & 1.00 & 17.69 & 46.46 & 95.88 & 1.30 & 0.63 \\
 & 0.05 & 1.00 & 62.26 & 93.37 & 1249.09 & 1.18 & 1.60 \\
\cline{1-8}
\bottomrule
\end{tabular}
}
\end{table}

\begin{table}[H]
\caption{Algorithm clustering CPU time among dataset collection 4}
\label{tab:time_COL4}
\centering
\resizebox{0.9\textwidth}{!}{%
\begin{tabular}{cc|cccccc}
\toprule
 & Method & KSKM & KSKM\_E & PCCC & BLPKMCC & COPKM\_P & COPKM \\
Dataset & Constraints Level &  &  &  &  &  &  \\
\midrule
\multirow[t]{3}{*}{banana} & 0.005 & 0.10 & 0.11 & 4.73 & 1.94 & 0.37 & - \\
 & 0.01 & 0.20 & 0.31 & 5.22 & 2.50 & 0.18 & - \\
 & 0.05 & 0.01 & 0.01 & 0.08 & 1.55 & 0.03 & 0.03 \\
\cline{1-8}
\multirow[t]{3}{*}{cifar10} & 0.005 & 322.69 & 645.73 & 1224.33 & 1881.88 & 55.18 & 16.07 \\
 & 0.01 & 118.05 & 3602.63 & 3633.96 & 3624.14 & 26.65 & - \\
 & 0.05 & 2.95 & 3.30 & 8.61 & 1135.66 & 2.63 & 2.68 \\
\cline{1-8}
\multirow[t]{3}{*}{cifar100} & 0.005 & 1359.85 & 3128.66 & 3638.03 & 3619.73 & 620.46 & 166.84 \\
 & 0.01 & 1266.71 & 3705.35 & 3851.99 & 3949.69 & 754.80 & 154.89 \\
 & 0.05 & 329.19 & 3696.78 & - & - & 28.41 & - \\
\cline{1-8}
\multirow[t]{3}{*}{letter} & 0.005 & 30.17 & 116.65 & 749.37 & 571.06 & 2.73 & 6.77 \\
 & 0.01 & 17.44 & 262.23 & 695.34 & 1002.48 & 12.79 & 5.47 \\
 & 0.05 & 2.64 & 944.50 & 1154.58 & 3376.45 & 1.67 & - \\
\cline{1-8}
\multirow[t]{3}{*}{mnist} & 0.005 & 136.71 & 434.18 & 804.60 & 1805.80 & 109.58 & 12.76 \\
 & 0.01 & 60.67 & 1357.07 & 3790.51 & 3627.71 & 60.77 & - \\
 & 0.05 & 1.36 & 1.40 & 9.13 & 1572.45 & 2.02 & 2.07 \\
\cline{1-8}
\multirow[t]{3}{*}{shuttle} & 0.005 & 2.29 & 35.71 & 236.72 & 198.64 & 4.96 & 2.22 \\
 & 0.01 & 0.68 & 22.89 & 78.96 & 278.99 & 2.95 & 0.99 \\
 & 0.05 & 0.84 & 0.92 & 10.37 & 2746.80 & 1.33 & 1.37 \\
\cline{1-8}
\bottomrule
\end{tabular}
}
\end{table}


Overall, KSKM and KSKM\_E achieve the best performance in terms of feasibility, efficiency, scalability and solution optimality. KSKM significantly improves clustering inertia compared to COP-$K$-Means methods with an acceptable increase in runtime, while attaining a level of optimality comparable to the state-of-the-art method PCCC at a lower computational cost. KSKM\_E further enhances solution quality by improving upon this optimality.

\subsection{Speed Improvement on Cluster Assignment by Kempe Swaps}
In this experiment, we compare how Kempe Swaps helps accelerating cluster assignment compared with GCP-based assignment~(\ref{eq:7}). The methods include:

\begin{itemize}
\item \textbf{KSKM}: The proposed Kempe Swap $K$-Means method (Algorithm \ref{alg:KCSKM}) without solving the fixed-centroid graph coloring problem. No step of mutations is allowed. KSKM is included as a benchmark for comparison.
\item \textbf{KSKM\_E}: The proposed Kempe Swap $K$-Means Extension method (Algorithm \ref{alg:KCSKM}), where the GCP-assignment is enabled. Still, GCP-assignment only serves as finding an improved solution for further Kempe swaps. No step of mutations is allowed.
\item \textbf{PCCC}: A state-of-the-art iterative integer programming approach \citep{baumann2025algorithm} that considers only hard constraints. Cluster repositioning is disabled, and the model-size reduction technique is applied with $q=10$.
\end{itemize}

Each method is run 10 times with random initializations, and we report the mean clustering objective (inertia) and mean CPU runtime across these runs.

\begin{table}[H]
\caption{Comparison of clustering inertia (left) and CPU time (right) among KSKM, KSKM\_E, and PCCC with no step of centroid mutation, reported as multiples of the baseline method (KSKM) across datasets. Lower inertia and smaller time multiples indicate better performance.}
\label{tab:obj_time_no_mutation}
\centering
\resizebox{0.9\textwidth}{!}{%
\begin{tabular}{cc|ccc|ccc}
\toprule
 & Stats & \multicolumn{3}{c}{Inertia} & \multicolumn{3}{c}{CPU Time} \\
 & Method & KSKM & KSKM\_E & PCCC & KSKM & KSKM\_E & PCCC \\
Dataset & Constraint Level &  &  &  &  &  &  \\
\midrule
\multirow[t]{4}{*}{COL1} & 0.05 & 1.00 & 1.00 & 1.00 & 1.00 & 4.09 & 25.84 \\
 & 0.1 & 1.00 & 1.00 & 1.00 & 1.00 & 5.84 & 27.47 \\
 & 0.15 & 1.00 & 1.00 & 0.99 & 1.00 & 6.74 & 25.80 \\
 & 0.2 & 1.00 & 1.00 & 1.00 & 1.00 & 7.96 & 37.80 \\
\cline{1-8}
\multirow[t]{10}{*}{COL2} & 0.05 & 1.00 & 1.00 & 0.99 & 1.00 & 7.57 & 81.68 \\
 & 0.1 & 1.00 & 0.98 & 0.98 & 1.00 & 37.19 & 180.17 \\
 & 0.15 & 1.00 & 0.96 & 0.96 & 1.00 & 59.67 & 286.06 \\
 & 0.2 & 1.00 & 0.96 & 0.97 & 1.00 & 83.40 & 338.96 \\
 & 0.25 & 1.00 & 0.95 & 0.97 & 1.00 & 105.16 & 412.30 \\
 & 0.3 & 1.00 & 0.94 & 0.97 & 1.00 & 108.52 & 450.33 \\
 & 0.35 & 1.00 & 0.99 & 0.99 & 1.00 & 101.87 & 372.77 \\
 & 0.4 & 1.00 & 0.99 & 0.98 & 1.00 & 111.36 & 368.39 \\
 & 0.45 & 1.00 & 0.98 & 0.97 & 1.00 & 103.70 & 438.68 \\
 & 0.5 & 1.00 & 0.99 & 1.00 & 1.00 & 50.52 & 205.19 \\
\cline{1-8}
\multirow[t]{4}{*}{COL3} & 0.05 & 1.00 & 0.98 & 0.98 & 1.00 & 48.23 & 412.70 \\
 & 0.1 & 1.00 & 0.96 & 0.96 & 1.00 & 329.98 & 1162.69 \\
 & 0.15 & 1.00 & 0.96 & 0.96 & 1.00 & 485.59 & 2245.15 \\
 & 0.2 & 1.00 & 0.98 & 0.98 & 1.00 & 425.59 & 1085.59 \\
\cline{1-8}
\bottomrule
\end{tabular}
}
\end{table}


Without the cluster-mutation step, KSKM explores a smaller solution space than KSKM\_E and PCCC, which results in slightly higher clustering inertia. However, its runtime is up to three orders of magnitude faster. This drastic reduction in runtime is particularly advantageous when the cluster-mutation step is enabled, since KSKM can then explore a much larger solution space within a significantly shorter runtime, as discussed in the previous subsection.

KSKM\_E achieves an average clustering inertia comparable to that of PCCC, while requiring substantially less time to converge to a local optimum. The runtime reduction arises primarily from performing Kempe swaps in place of repeated exact assignments~(\ref{eq:7}), and secondarily from avoiding the repeated reformulation of the GCP induced by the model-size reduction technique used in PCCC. Notably, in a centroid-based iterative method, the cluster-assignment step does not need to correspond to the steepest descent direction until the algorithm approaches local optimality.

\subsection{Accuracy Improvement by Centroid Perturbation}
In this experiment, we compare how the centroid perturbation step~(\ref{alg:KSPerturb}) improves cluster assignment, including additional methods:
\begin{itemize}
\item \textbf{KSKM\_repo}: The proposed Kempe Swap $K$-Means method. The step of centroid perturbation has been replaced by centroid repositioning.
\item \textbf{KSKM\_E\_repo}: The proposed Kempe Swap $K$-Means Extension method. The step of centroid perturbation has been replaced by centroid repositioning.
\end{itemize}
\begin{table}[H]
\caption{Comparison of clustering inertia for KSKM variants with and without centroid perturbation, reported as multiples of the baseline KSKM across datasets for successful runs.}
\label{tab:obj_repo}
\centering
\resizebox{0.9\textwidth}{!}{%
\begin{tabular}{cc|cccc}
\toprule
 & Method & KSKM & KSKM\_repo & KSKM\_E & KSKM\_E\_repo \\
Dataset & Constraint Level &  &  &  &  \\
\midrule
\multirow[t]{4}{*}{COL1} & 0.05 & 1.000 & 1.001 & 1.000 & 1.002 \\
 & 0.1 & 1.000 & 1.003 & 0.998 & 1.001 \\
 & 0.15 & 1.000 & 1.005 & 0.999 & 1.001 \\
 & 0.2 & 1.000 & 1.002 & 0.998 & 1.000 \\
\cline{1-6}
\multirow[t]{10}{*}{COL2} & 0.05 & 1.000 & 1.019 & 0.999 & 1.017 \\
 & 0.1 & 1.000 & 1.023 & 1.001 & 1.018 \\
 & 0.15 & 1.000 & 1.026 & 0.998 & 1.018 \\
 & 0.2 & 1.000 & 1.027 & 1.003 & 1.014 \\
 & 0.25 & 1.000 & 1.030 & 0.990 & 1.006 \\
 & 0.3 & 1.000 & 1.035 & 0.977 & 0.991 \\
 & 0.35 & 1.000 & 1.018 & 1.005 & 1.009 \\
 & 0.4 & 1.000 & 1.021 & 1.001 & 1.013 \\
 & 0.45 & 1.000 & 1.011 & 1.002 & 1.006 \\
 & 0.5 & 1.000 & 1.009 & 1.002 & 1.002 \\
\cline{1-6}
\bottomrule
\end{tabular}
}
\end{table}

Without centroid perturbation, the clustering objective deteriorates. This behavior is expected, as cluster assignments are highly sensitive to centroid positions. Centroid perturbation enables exploration of the local neighborhood around the current solution, allowing for refined improvements that cannot be achieved through centroid repositioning alone, which primarily encourages exploration of more distant regions of the solution space.

The benefit of centroid perturbation is more pronounced for the base Kempe Swap 
$K$-Means method without GCP-based assignment. Because the pure Kempe Swap assignment is greedy and highly sensitive to centroid placement, a Kempe-Swap-optimal solution can often be further improved through either a GCP-assignment step or centroid perturbation.

\subsection{Disadvantages of Kempe Swaps}

We showed in the previous section that methods based on Kempe swaps are efficient and powerful for most numerical examples: regardless of dataset size, KSKM is able to find near-optimal solutions much faster than state-of-the-art methods. However, Kempe swaps are not a universally optimal choice. Note that even at a 50\% constraint level, when the ratio of must-link constraints is large or the number of clusters is high, the graph-coloring problem is relatively unconstrained and it is not difficult to find and improve a feasible solution. In contrast, for a truly constrained and dense graph, the effectiveness of KSKM degrades: Kempe chains become large and global, such that a single chain may span almost an entire pair of clusters $(C_i, C_j)$. This situation typically arises when both of the following conditions are satisfied:
\begin{enumerate}
    \item the graph $(V,E)$ is dense, and super-nodes are densely connected by cannot-link constraints;
    \item the number of clusters $k$ is close to the chromatic number of $(V,E)$.
\end{enumerate}

When a dense graph is combined with a small number of clusters, the subgraph induced by any pair of clusters $(C_i, C_j)$ becomes dense and usually forms one or a few giant Kempe chains. In such cases, a Kempe swap produces a large, global perturbation rather than a localized adjustment. Consequently, Kempe-swap-based methods lose fine-grained control over clustering inertia by exploring a tiny part of the solution space and may become trapped on plateaus corresponding to large attraction basins, where further improvements are difficult to obtain.

In this subsection we further compare the state of the art methods in \ref{OverallComparison} in datasets collection 1 but remove the must-link constraints, $\mathcal{ML} = \emptyset$. Again, each methods is implemented 10 times and the average results of clustering inertia, CPU time, rate of success is reported.

\begin{table}[H]
\caption{Clustering inertia and success rate among dataset collection 1 without ML constraints}
\label{tab:COL1_no_ml_obj_rate_success}
\centering
\resizebox{0.85\textwidth}{!}{%
\begin{tabular}{ccc|cccccc}
\toprule
 &  & Method & KSKM & KSKM\_E & PCCC & BLPKMCC & COPKM\_P & COPKM \\
Dataset & Constraints Level &  &  &  &  &  &  &  \\
\midrule
\multirow[t]{3}{*}{hayes-roth} & \multirow[t]{3}{*}{0.2} & Inertia & 601.79 & 535.76 & 536.08 & 546.98 & 608.25 & - \\
 &  & rate\_success & 0.50 & 1.00 & 1.00 & 1.00 & 0.50 & 0.00 \\
 &  & time & 0.01 & 0.24 & 0.73 & 0.28 & 0.00 & - \\
\cline{1-9} \cline{2-9}
\multirow[t]{3}{*}{tae} & \multirow[t]{3}{*}{0.2} & Inertia & 719.54 & 681.81 & 678.54 & 683.70 & 731.36 & - \\
 &  & rate\_success & 0.10 & 1.00 & 1.00 & 1.00 & 0.10 & 0.00 \\
 &  & time & 0.00 & 0.39 & 3.12 & 1.00 & 0.00 & - \\
\cline{1-9} \cline{2-9}
\multirow[t]{6}{*}{vehicle} & \multirow[t]{3}{*}{0.15} & Inertia & - & 13308.06 & 13326.75 & 13321.72 & - & - \\
 &  & rate\_success & 0.00 & 0.40 & 0.70 & 0.40 & 0.00 & 0.00 \\
 &  & time & 0.01 & 3600.10 & 3600.25 & 3600.22 & - & - \\
\cline{2-9}
 & \multirow[t]{3}{*}{0.2} & Inertia & - & 13334.17 & 13334.17 & 13334.17 & - & - \\
 &  & rate\_success & 0.00 & 1.00 & 1.00 & 1.00 & 0.00 & 0.00 \\
 &  & time & 0.01 & 53.79 & 3600.43 & 3600.46 & - & - \\
\cline{1-9} \cline{2-9}
\bottomrule
\end{tabular}
}
\end{table}

A concurrent phenomenon associated with large, dense Kempe chains is the infeasibility of solutions produced by the DSATUR algorithm; specifically, the target number of clusters $k$ is smaller than the number of colors found by DSATUR. For example, on the $\mathrm{hayes-roth}$ dataset with constraint level 0.2, the true number of clusters is 3, whereas the classic DSATUR algorithm cannot find a coloring with fewer than 4 clusters. In 5 out of 10 runs the DSATUR-inspired assignment yields a feasible solution, and KSKM tends to converge to suboptimal local minima. The best run among the 5 trials achieves a near-optimal solution with inertia 535.76; however, the average inertia is 601.79, because in most runs KSKM is initialized with a solution that is not Kempe-equivalent to the best one.

In contrast, KSKM\_E allows communication between non–Kempe-equivalent solutions by improving the solution through solving a GCP, thereby mitigating this limitation. A similar phenomenon is observed on the $\mathrm{tae}$ dataset with constraint level 0.2, where the classic DSATUR algorithm cannot find a feasible clustering with $k=3$. In this case, the DSATUR-inspired nearest-centroid assignment produces a feasible solution in only 1 out of 10 runs, and KSKM using only Kempe swaps fails to escape local optima and reach the global optimum.

An even more challenging scenario arises for the $\mathrm{vehicle}$ dataset with constraint levels 0.15 and 0.2. The corresponding cannot-link constraint graphs are highly complex, and both DSATUR-based methods fail to find a feasible solution. As a result, KSKM cannot be initialized and therefore fails to produce any feasible clustering. Notably, GCP-based methods such as BLPKMCC and PCCC either fail to identify a feasible solution or exceed the time limit on these complex graphs.

As a rule of thumb, when the DSATUR algorithm cannot find a feasible $k$-clustering, the performance of KSKM is inherently limited. In such cases, we recommend using KSKM\_E, which combines Kempe swaps with GCP-based refinement.

\newpage
\section{Conclusion}
\label{sec4}
The introduction of Kempe Swap $K$-Means (KSKM) bridges a key gap in centroid-based iterative methods for clustering problems with hard must-link and cannot-link constraints.

In the cluster assignment step, given fixed centroids, KSKM provides a middle ground between two major paradigms:
\begin{itemize}
    \item Sequential assignment of data to the nearest feasible centroid, e.g., COP-$K$-Means \citep{wagstaff2001constrained}, ICOP-$K$-Means \citep{tan2010improved}, CLC-$K$-Means \citep{yang2013improved}; and
    \item Simultaneous assignment by solving a graph coloring or optimization problem, e.g., BCKM \citep{le2018binary}, BLPKMCC \citep{baumann2020}, PCCC \citep{baumann2025algorithm}, IPC-$K$-Means \citep{piccialli2022exact}.
\end{itemize}

The sequential approaches are computationally efficient but struggle with violations of cannot-link constraints and cannot guarantee improvement of the objective function at each iteration. In contrast, the simultaneous approaches achieve deeper objective improvements but are computationally expensive and do not scale well to large datasets.

KSKM addresses this trade-off by introducing Kempe Swaps, which iteratively improve the current clustering solution without breaking constraints. Each swap involves solving a less challanging maximum weighted independent set (MWIS) problem, which can be efficiently approximated in polynomial time, thus maintaining scalability to large problems.

In the centroid update step, KSKM introduces hill-climbing perturbations to escape local minima by adjusting cluster centroids. Local optima in constrained clustering—particularly with cannot-link constraints—are highly sensitive to centroid positions. Small perturbations around a centroid typically incur little cost increase but often enable beneficial Kempe Swaps that further reduce the objective. Larger shifts, meanwhile, allow exploration beyond local neighborhoods.


KSKM pairs effectively with DSATUR-inspired assignment to generate an initial solution. In terms of scalability to large graphs, KSKM scales more effectively to large datasets than state-of-the-art methods. As long as DSATUR algorithms can find a solution, KSKM can further improve the solution by performing swaps. On the other hand, KSKM integrates naturally with existing constrained clustering methods. For example, its Kempe Swap assignment can provide a high-quality heuristic starting point for fixed-centroid graph coloring formulations such as \cite{baumann2020} and \cite{mansueto2025efficiently} or upper bounds for exact solution such as \cite{babaki2014constrained}, \cite{piccialli2022exact}. By incorporating Kempe swap assignment, existing methods can be dramatically accelerated.

In summary, KSKM improves iterative constrained clustering by jointly optimizing efficiency and solution quality. Kempe swaps facilitate efficient identification of local optima, while centroid perturbation enables the algorithm to escape unfavorable local minima, resulting in more accurate clusterings.


\newpage

\appendix
\section{Algorithmic and Numeric Details}
\label{app}

\subsection{Heuristic Solution for MWIS}
\label{app1}
Since the clique constraints can be equivalently represented as cannot-link constraints, the MWIS problem (\ref{eq:3}) can be reformulated as a binary quadratic program (BQP) (\ref{eq:4}).  

Let $Q$ be a square matrix of size $|S|$, defined as
\[
Q_{s,t} =
\begin{cases}
d_s, & s = t, \\[6pt]
-D, & (s,t) \in \mathcal{CL}^{swap} \ \text{or} \ \exists v \in V \ \text{with } v \in s \cap t,\; s \neq t, \\[6pt]
0, & \text{otherwise},
\end{cases}
\]
where $D$ is a penalty parameter chosen such that $D > 0.5 \max_{s \in S} d_s$. The optimization problem becomes

\begin{equation}\label{eq:4}
\min_{u \in \{0,1\}^{S}} \; u^\top Q u.
\end{equation}
\cite{hosseinian2018nonconvex} showed that the binary constraint in (\ref{eq:4}) can be relaxed to $u \in [0,1]^S$, and that local minima of the relaxed quadratic program coincide with those of MWIS (\ref{eq:3}). The good news is that we don't have to solve (\ref{eq:3}) or (\ref{eq:4}) to optimality to achieve steep descent. Following the heuristic by \cite{hosseinian2018nonconvex}, we solve an approximate problem on the unit sphere (\ref{eq:5}) and extract independent sets from stationary points of eigenvector:

\begin{equation}\label{eq:5}
\min_{\bar{u}} \; \bar{u}^\top Q \bar{u}
\quad \text{s.t.} \quad \bar{u}^\top \bar{u} = 1.
\end{equation}
\newpage
\subsection{Multi-Kempe Chain Swap}
\label{app2}

Kempe Chain Swap $K$-Means leads to many local optimal solutions. On one hand, there are many combinations of cluster centroids that are local optimal. On the other hand, the optimality proposed is two-cluster-swap optimal. For instance, a cyclic rotation such as $C_1 \to C_2, C_2 \to C_3, C_3 \to C_1$ may yield an improved objective even when every myopic two-cluster swap among $C_1, C_2, C_3$ is sub-optimal.

Multi‑Kempe $s$-chains generalise Kempe chains and introduce perturbations inaccessible to $s-1$ swaps, enabling transitions that cannot be decomposed into feasibility-preserving $(s-1)$-color flips and addressing the problem of $(s-1)$-color-swap-optimal.

However, evaluating a multi-Kempe $s$-chain swap is computationally challenging due to the combinatorial number of possible chain combinations and the fact that assessing the swap cost requires solving $(s-1)$-chain improvements in a nested manner. In this work, we therefore restrict our consideration to $3$-chain swaps once two-cluster-swap optimality has been achieved.

For any $3$-chain tuple $(H_i, H_j, H_l)$ with corresponding cluster memberships $(C_i, C_j, C_l)$ in a two-cluster-swap-optimal assignment, we apply two-cluster Kempe swaps on the rotated subgraphs induced by $H_i \cup H_j \cup H_l$, $(H_i, H_j, H_l) \leftarrow (C_l, C_i, C_j)$ and $(H_i, H_j, H_l) \leftarrow (C_j, C_l, C_i)$, iteratively until no further improvement can be obtained. We then maintain a set $S$ of swaps that yield an improvement in the clustering objective and resolve compatible improving swaps using the MWIS formulation in (\ref{eq:3}).

\begin{center}
\resizebox{0.9\linewidth}{!}{%
\begin{algorithm}[H]
\caption{MultiKempe Swaps Assignment $\mathrm{MultiKSAssignment}(V,E, U, y, \mu)$}
\label{alg:MultiKSAssignment}
\DontPrintSemicolon
\SetKwInOut{Input}{Input}\SetKwInOut{Output}{Output}
\Input{
Graph of super-nodes and cannot-link edges $G = (V, E)$;\\
Data $\{y_i\}_{v_i \in V}$;\\
Two-swap optimal Cluster assignments $U = \{u(v_i)\}_{v_i \in V}$;\\
Cluster centroids $\mu$.
}

\BlankLine
\tcp{Keep improving Kempe swaps}
$S = dict()$\;
\tcp{Keep clique constraints}
$\mathcal{CC} = dict()$\;
\For{$i,j,l \in \text{cluster tuples}(U)$}{
    \tcp{Find maximal connected subgraphs induced on $i,j,l$}
    $\mathbb{H} = \mathrm{KempeChains}(i,j,l,\mathcal{CL})$\;
    \For{$H \in \mathbb{H}$}{
        \tcp{Split Kempe Chain}
         $H_i,H_j,H_l = \mathrm{split}(H)$\;
        \For{$U(H_i,H_j,H_l) \in \{(C_l,C_i,C_j), (C_j,C_l,C_i)\}$}{
        \tcp{Improve assignment after membership rotation}
        \While{not converged}{
        $\mu(i,j,l) = \mathrm{Centroid}(U,U(H_i,H_j,H_l))$\;
        $U(H_i,H_j,H_l) = \mathrm{KSAssignment}(H,E, U(H_i,H_j,H_l), y, \mu(i,j,l))$\;
        }
        }
        \If{objective is improved}{
        \tcp{Record improving swaps}
        $S(H_i,H_j,H_l)=(U(H_i,H_j,H_l),-d_{H_i,H_j,H_l})$\;
        \tcp{Record swap id $s = \mathrm{length}(S)$ in the Clique constraints}
        \For{$p \in H_i \cup H_j \cup H_l$}{$\mathrm{append}(\mathcal{CC}(p), length(S))$\;}
        }
    }
}
\tcp{Find cannot-link swaps}
$\mathcal{CL}^{swap} = \{(s,t), s,t \in S: \exists (v_i,v_j) \in \mathcal{CL}, U(v_i) = U(v_j), v_i \in s, v_j \in t\}$\;
\tcp{Solve Maximum Weighted Independent Sets with a heuristic}
$u_{swap} = \mathrm{MWIS}(S, \mathcal{CC}, \mathcal{CL}^{swap})$\;
\tcp{Update cluster assignments}
$U = \mathrm{update}(U,u_{swap})$\;
\Output{Improved cluster assignment $U$}
\end{algorithm}
}
\end{center}
\newpage
\subsection{Exact Assignment: Fixed-Centroid Graph Coloring Problem}
\label{app3}

\begin{subequations}\label{eq:7}
\begin{align}
    \min_{x, m} \quad & \sum_{i=1}^{n} \sum_{j=1}^{k} D_{ij} x_{ij}  \tag{\ref{eq:7}} \\
    \text{s.t.} \quad 
    & \sum_{j=1}^{k} x_{ij} = 1, && \forall i \in [n], \label{eq:7a}\\
    & \sum_{i=1}^{n} x_{ij} \ge 1, && \forall j \in [k], \label{eq:7b}\\
    & x_{ih} = x_{jh}, && \forall h \in [k],\ \forall (i,j) \in \mathcal{ML}, \label{eq:7c}\\
    & x_{ih} + x_{jh} \le 1, && \forall h \in [k],\ \forall (i,j) \in \mathcal{CL}, \label{eq:7d}\\
    & x_{ij} \in \{0,1\}, && \forall i \in [n],\ \forall j \in [k], \label{eq:7e}
\end{align}
\end{subequations}
where $D_{i,j} =  \| y_i - \mu_j \|_2^2$. With fixed centroids, the mixed integer second order cone program (\ref{eq:1}) reduced to a integer program (\ref{eq:7}), still essentially a hard graph coloring problem.
\newpage
\subsection{DSATUR COP-$K$-Means}
\label{app4}
\begin{center}
\resizebox{0.9\linewidth}{!}{%
\begin{algorithm}[H]
\caption{DSATUR $K$-Means $\mathrm{DSATURKM}(V, E, y, k, L)$}
\label{alg:DSATURKM}
\DontPrintSemicolon
\SetKwInOut{Input}{Input}\SetKwInOut{Output}{Output}
\Input{
Graph of super-nodes and cannot-link edges $G = (V, E)$;\\
Data $y$;\\
Number of clusters $k$;\\
Number of iterations $L$.
}
\BlankLine
\tcp{Initialize centroids}
$\mu = \mathrm{InitCentroids}(y,k)$\;
\For{$l=1\cdots L$}{
\tcp{Initiate an empty set of membership}
$U = \{\}$\;
\While{$|U| < |V|$}{
\tcp{Assign the remaining super-node with maximum degree of saturation to the nearest centroid}
$v = \text{argmax}(\mathrm{DegreeSaturation}(V,U))$\;
\For{$i=1\cdots k$}{
$distance = \infty$\;
\If{not $\mathrm{violation}(v,i,U,E)$ and $||y_v-\mu_i||< distance$}{
$distance = ||y_v-\mu_i||$\;
$U(v) = i$\;
}
\eIf{$distance = \infty$}{\Output{Terminate with infeasibility}
}{
\tcp{Update cluster centroids}
$\mu=\mathrm{Centroids}(U, y)$\;
Keep the best assignments $U^{*}$\;
\If{$\mu$ is not updated}{\Output{Best assignments $U^{*}$}}
}
}
}
\Output{Best assignments $U^{*}$}}
\end{algorithm}
}
\end{center}

\newpage
\subsection{Comparison with S\_MDEClust}
\label{app:smde}
\textbf{S\_MDEClust} is a state-of-the-art method based on memetic differential evolution combined with iterative integer programming \citep{mansueto2025efficiently}. It aims to improve clustering inertia through extensive exploration of the solution space starting from multiple solution populations, which significantly increases computational cost. To ensure a fair comparison between S\_MDEClust and the other methods, we initialize S\_MDEClust with 10 solution populations and compare its performance against the best results from 10 independent runs of KSKM, KSKM\_E, PCCC, and BLPKMCC. We report the minimum clustering inertia, total runtime, and maximum ARI across the 10 runs. We only compare the methods on dataset collection 1 and 2 as the cost of running S\_MDEClust is substential for larger datasets.
\begin{table}[H]
\caption{Comparison of minimum clustering inertia over 10, reported as multiples of the baseline method (KSKM) across datasets among success runs. Lower inertia multiples indicate better performance.}
\label{tab:obj_S_MDEClust}
\centering
\resizebox{0.9\textwidth}{!}{%
\begin{tabular}{cc|cccccccc}
\toprule
 & Method & KSKM & KSKM\_E & PCCC & BLPKMCC & COPKM\_P & COPKM & S\_MDEClust \\
Dataset & Constraint Level &  &  &  &  &  &  &  \\
\midrule
\multirow[t]{4}{*}{COL1} & 0.05 & 1.00 & 1.00 & 1.00 & 1.01 & 1.01 & 1.02 & 1.00 \\
 & 0.1 & 1.00 & 1.00 & 1.00 & 1.00 & 1.01 & 1.01 & 1.00 \\
 & 0.15 & 1.00 & 1.00 & 1.00 & 1.00 & 1.00 & 1.01 & 1.00 \\
 & 0.2 & 1.00 & 1.00 & 1.00 & 1.00 & 1.00 & 1.00 & 1.00 \\
\cline{1-9}
\multirow[t]{10}{*}{COL2} & 0.05 & 1.00 & 1.00 & 1.02 & 1.08 & 1.04 & 1.03 & 1.01 \\
 & 0.1 & 1.00 & 1.00 & 1.02 & 1.08 & 1.04 & 1.04 & 1.02 \\
 & 0.15 & 1.00 & 1.00 & 1.01 & 1.09 & 1.06 & 1.06 & 1.00 \\
 & 0.2 & 1.00 & 1.00 & 1.01 & 1.11 & 1.06 & 1.07 & 1.01 \\
 & 0.25 & 1.00 & 1.00 & 1.01 & 1.04 & 1.04 & 1.08 & 1.01 \\
 & 0.3 & 1.00 & 1.00 & 1.01 & 1.04 & 1.04 & 1.05 & 1.00 \\
 & 0.35 & 1.00 & 1.00 & 1.01 & 1.03 & 1.03 & 1.04 & 1.00 \\
 & 0.4 & 1.00 & 1.00 & 1.00 & 1.04 & 1.04 & 1.04 & 1.01 \\
 & 0.45 & 1.00 & 1.00 & 1.01 & 1.03 & 1.02 & 1.03 & 1.00 \\
 & 0.5 & 1.00 & 1.00 & 1.00 & 1.02 & 1.01 & 1.01 & 1.01 \\
\cline{1-9}
\bottomrule
\end{tabular}
}
\end{table}

S\_MDEClust improves the clustering objective compared to BLPKMCC by incorporating a centroid crossover step that enables solution mutation. However, it does not yield significant improvements over KSKM, KSKM\_E, and PCCC, which also allow solution mutation through centroid perturbation and centroid repositioning.
\begin{table}[H]
\caption{Comparison of total CPU time over 10, reported as multiples of the baseline method (KSKM) across datasets among success runs. Smaller time multiples indicate faster performance.}
\label{tab:time_S_MDEClust}
\centering
\resizebox{0.9\textwidth}{!}{%
\begin{tabular}{cc|cccccccc}
\toprule
 & Method & KSKM & KSKM\_E & PCCC & BLPKMCC & COPKM\_P & COPKM & S\_MDEClust \\
Dataset & Constraint Level &  &  &  &  &  &  &  \\
\midrule
\multirow[t]{4}{*}{COL1} & 0.05 & 1.00 & 2.62 & 10.43 & 4.19 & 0.40 & 0.85 & 13.82 \\
 & 0.1 & 1.00 & 3.35 & 10.15 & 10.03 & 0.68 & 0.14 & 23.79 \\
 & 0.15 & 1.00 & 3.49 & 9.98 & 23.87 & 0.60 & 0.45 & 36.18 \\
 & 0.2 & 1.00 & 3.51 & 11.69 & 41.31 & 0.66 & 0.56 & 58.43 \\
\cline{1-9}
\multirow[t]{10}{*}{COL2} & 0.05 & 1.00 & 10.87 & 16.10 & 7.70 & 0.17 & 0.39 & 93.98 \\
 & 0.1 & 1.00 & 22.02 & 30.01 & 13.42 & 0.11 & 0.17 & 130.46 \\
 & 0.15 & 1.00 & 31.86 & 59.10 & 25.81 & 0.13 & 0.18 & 222.16 \\
 & 0.2 & 1.00 & 44.05 & 75.83 & 43.07 & 0.13 & 0.14 & 274.98 \\
 & 0.25 & 1.00 & 61.73 & 94.77 & 63.62 & 0.14 & 0.16 & 405.00 \\
 & 0.3 & 1.00 & 63.07 & 105.19 & 100.33 & 0.22 & 0.21 & 446.07 \\
 & 0.35 & 1.00 & 67.82 & 86.11 & 146.76 & 0.32 & 0.31 & 564.55 \\
 & 0.4 & 1.00 & 88.14 & 105.77 & 177.43 & 0.34 & 0.35 & 666.39 \\
 & 0.45 & 1.00 & 102.36 & 119.02 & 237.11 & 0.41 & 0.40 & 751.57 \\
 & 0.5 & 1.00 & 44.95 & 76.50 & 250.65 & 0.42 & 0.43 & 670.85 \\
\cline{1-9}
\bottomrule
\end{tabular}
}
\end{table}

However, the computational cost of this extensive exploration is substantial, as shown in Table~(\ref{tab:time_S_MDEClust}).


\newpage
 \bibliographystyle{elsarticle-harv} 
 \bibliography{ref}

\end{document}


\appendix
\section{Numerical Details}
\subsection{Data Description}

\begin{table}[H]
\centering
\caption{Characteristics of data sets from collection COL1}
\label{tab:col1}
\resizebox{0.99\textwidth}{!}{%
\begin{tabular}{cccccc}
\toprule
Data set & Objects & Features & Classes & Type & Source \\
\midrule
Appendicitis     & 106 & 7  & 2  & real      & keel \\
Breast Cancer    & 569 & 30 & 2  & real      & Sklearn \\
Bupa             & 345 & 6  & 2  & real      & Keel \\
Circles          & 300 & 2  & 2  & synthetic & GitHub \\
Ecoli            & 336 & 7  & 8  & real      & Keel \\
Glass            & 214 & 9  & 6  & real      & Keel \\
Haberman         & 306 & 3  & 2  & real      & Keel \\
Hayesroth        & 160 & 4  & 3  & real      & Keel \\
Heart            & 270 & 13 & 2  & real      & Keel \\
Ionosphere       & 351 & 33 & 2  & real      & Keel \\
Iris             & 150 & 4  & 3  & real      & Keel \\
Led7Digit        & 500 & 7  & 10 & real      & Keel \\
Monk2            & 432 & 6  & 2  & real      & Keel \\
Moons            & 300 & 2  & 2  & synthetic & GitHub \\
Movement Libras  & 360 & 90 & 15 & real      & Keel \\
Newthyroid       & 215 & 5  & 3  & real      & Keel \\
Saheart          & 462 & 9  & 2  & real      & Keel \\
Sonar            & 208 & 60 & 2  & real      & Keel \\
Spectfheart      & 267 & 44 & 2  & real      & Keel \\
Spiral           & 300 & 2  & 2  & synthetic & GitHub \\
Soybean          & 47  & 35 & 4  & real      & Keel \\
Tae              & 151 & 5  & 3  & real      & Keel \\
Vehicle          & 846 & 18 & 4  & real      & Keel \\
Wine             & 178 & 13 & 3  & real      & Keel \\
Zoo              & 101 & 16 & 7  & real      & Keel \\
\bottomrule
\end{tabular}
}
\end{table}

\begin{table}[H]
\centering
\caption{Characteristics of data sets from collection COL3}
\label{tab:col3}
\resizebox{0.99\textwidth}{!}{%
\begin{tabular}{cccccc}
\toprule
Data set & Objects & Features & Classes & Type & Source \\
\midrule
n500-k2    & 500  & 2 & 2   & synthetic & Sklearn \\
n500-k5    & 500  & 2 & 5   & synthetic & Sklearn \\
n500-k10   & 500  & 2 & 10  & synthetic & Sklearn \\
n500-k20   & 500  & 2 & 20  & synthetic & Sklearn \\
n500-k50   & 500  & 2 & 50  & synthetic & Sklearn \\
n500-k100  & 500  & 2 & 100 & synthetic & Sklearn \\
n1000-k2   & 1,000 & 2 & 2   & synthetic & Sklearn \\
n1000-k5   & 1,000 & 2 & 5   & synthetic & Sklearn \\
n1000-k10  & 1,000 & 2 & 10  & synthetic & Sklearn \\
n1000-k20  & 1,000 & 2 & 20  & synthetic & Sklearn \\
n1000-k50  & 1,000 & 2 & 50  & synthetic & Sklearn \\
n1000-k100 & 1,000 & 2 & 100 & synthetic & Sklearn \\
n2000-k2   & 2,000 & 2 & 2   & synthetic & Sklearn \\
n2000-k5   & 2,000 & 2 & 5   & synthetic & Sklearn \\
n2000-k10  & 2,000 & 2 & 10  & synthetic & Sklearn \\
n2000-k20  & 2,000 & 2 & 20  & synthetic & Sklearn \\
n2000-k50  & 2,000 & 2 & 50  & synthetic & Sklearn \\
n2000-k100 & 2,000 & 2 & 100 & synthetic & Sklearn \\
n5000-k2   & 5,000 & 2 & 2   & synthetic & Sklearn \\
n5000-k5   & 5,000 & 2 & 5   & synthetic & Sklearn \\
n5000-k10  & 5,000 & 2 & 10  & synthetic & Sklearn \\
n5000-k20  & 5,000 & 2 & 20  & synthetic & Sklearn \\
n5000-k50  & 5,000 & 2 & 50  & synthetic & Sklearn \\
n5000-k100 & 5,000 & 2 & 100 & synthetic & Sklearn \\
\bottomrule
\end{tabular}
}
\end{table}

\begin{table}[H]
\centering
\caption{Characteristics of data sets from collection COL4}
\label{tab:col4}
\resizebox{0.99\textwidth}{!}{%
\begin{tabular}{cccccc}
\toprule
Data set & Objects & Features & Classes & Type & Source \\
\midrule
Banana     & 5,300  & 2    & 2   & synthetic & Keel \\
Letter     & 20,000 & 16   & 26  & real      & Keel \\
Shuttle    & 57,999 & 9    & 7   & real      & Keel \\
Cifar 10   & 60,000 & 3,072 & 10 & real      & Cifar website \\
Cifar 100  & 60,000 & 3,072 & 100 & real     & Cifar website \\
Mnist      & 70,000 & 784  & 10  & real      & Mnist database \\
\bottomrule
\end{tabular}
}
\end{table}

\begin{table}[H]
\centering
\caption{Number of must-link (ML) and cannot-link (CL) constraints in the different constraint sets of collection COL1}
\label{tab:col1-constraints}
\resizebox{0.99\textwidth}{!}{%
\begin{tabular}{c cc cc cc cc}
\toprule
 & \multicolumn{2}{c}{5\% CS} & \multicolumn{2}{c}{10\% CS} &
   \multicolumn{2}{c}{15\% CS} & \multicolumn{2}{c}{20\% CS} \\
Data set & ML & CL & ML & CL & ML & CL & ML & CL \\
\midrule
Appendicitis   & 13 & 2   & 39  & 16  & 71   & 49   & 164  & 67   \\
Breast Cancer  & 216 & 190 & 876 & 720 & 1,965 & 1,690 & 3,487 & 2,954 \\
Bupa           & 79 & 74  & 323 & 272 & 699  & 627  & 1,201 & 1,145 \\
Circles        & 50 & 55  & 208 & 227 & 502  & 488  & 853  & 917  \\
Ecoli          & 30 & 106 & 163 & 398 & 357  & 918  & 609  & 1,669 \\
Glass          & 11 & 44  & 52  & 179 & 139  & 389  & 259  & 644  \\
Haberman       & 76 & 44  & 304 & 161 & 634  & 401  & 1,135 & 756  \\
Hayesroth      & 12 & 16  & 39  & 81  & 102  & 174  & 177  & 319  \\
Heart          & 41 & 50  & 178 & 173 & 396  & 424  & 744  & 687  \\
Ionosphere     & 92 & 61  & 330 & 300 & 732  & 646  & 1,299 & 1,186 \\
Iris           & 6  & 22  & 25  & 80  & 57   & 196  & 94   & 341  \\
Led7Digit      & 25 & 275 & 126 & 1,099 & 267 & 2,508 & 460 & 4,490 \\
Monk2          & 101 & 130 & 473 & 473 & 979 & 1,101 & 1,917 & 1,824 \\
Moons          & 55 & 50  & 200 & 235 & 494 & 496 & 900 & 870 \\
Movement Libras & 6 & 147 & 27 & 603 & 112 & 1,319 & 158 & 2,398 \\
Newthyroid     & 25 & 30  & 108 & 123 & 270 & 258 & 449 & 454 \\
Saheart        & 152 & 124 & 595 & 486 & 1,292 & 1,123 & 2,330 & 1,948 \\
Sonar          & 29 & 26  & 100 & 110 & 245 & 251 & 436 & 425 \\
Spectfheart    & 56 & 35  & 233 & 118 & 543 & 277 & 965 & 466 \\
Spiral         & 52 & 53  & 224 & 211 & 487 & 503 & 918 & 852 \\
Soybean        & 0  & 3   & 4   & 6   & 22  & 12  & 33  & 33  \\
Tae            & 8  & 20  & 40  & 80  & 82  & 171 & 151 & 314 \\
Vehicle        & 221 & 682 & 874 & 2,696 & 1,955 & 6,046 & 3,589 & 10,776 \\
Wine           & 8  & 28  & 31  & 122 & 70  & 281 & 143 & 487 \\
Zoo            & 7  & 8   & 21  & 34  & 29  & 91  & 41  & 169 \\
\bottomrule
\end{tabular}
}
\end{table}

\begin{table}[H]
\centering
\caption{Number of must-link (ML) and cannot-link (CL) constraints in the different constraint sets of collection COL2}
\label{tab:col2-constraints}
\resizebox{0.99\textwidth}{!}{%
\begin{tabular}{c cc cc cc cc cc cc cc cc cc cc}
\toprule
 & \multicolumn{2}{c}{5\% CS} & \multicolumn{2}{c}{10\% CS} &
   \multicolumn{2}{c}{15\% CS} & \multicolumn{2}{c}{20\% CS} &
   \multicolumn{2}{c}{25\% CS} & \multicolumn{2}{c}{30\% CS} &
   \multicolumn{2}{c}{35\% CS} & \multicolumn{2}{c}{40\% CS} &
   \multicolumn{2}{c}{45\% CS} & \multicolumn{2}{c}{50\% CS} \\
Data set
 & ML & CL & ML & CL & ML & CL & ML & CL
 & ML & CL & ML & CL & ML & CL & ML & CL
 & ML & CL & ML & CL \\
\midrule
n300-k10-s10 & 11 & 94 & 46 & 389 & 89 & 946 & 160 & 1,610 & 287 & 2,488 & 403 & 3,602 & 523 & 5,042 & 703 & 6,437 & 903 & 8,142 & 1,088 & 10,087 \\
n300-k10-s20 & 11 & 94 & 46 & 389 & 89 & 946 & 160 & 1,610 & 287 & 2,488 & 403 & 3,602 & 523 & 5,042 & 703 & 6,437 & 903 & 8,142 & 1,088 & 10,087 \\
n300-k10-s30 & 11 & 94 & 46 & 389 & 89 & 946 & 160 & 1,610 & 287 & 2,488 & 403 & 3,602 & 523 & 5,042 & 703 & 6,437 & 903 & 8,142 & 1,088 & 10,087 \\
n300-k10-s40 & 11 & 94 & 46 & 389 & 89 & 946 & 160 & 1,610 & 287 & 2,488 & 403 & 3,602 & 523 & 5,042 & 703 & 6,437 & 903 & 8,142 & 1,088 & 10,087 \\
n300-k10-s50 & 11 & 94 & 46 & 389 & 89 & 946 & 160 & 1,610 & 287 & 2,488 & 403 & 3,602 & 523 & 5,042 & 703 & 6,437 & 903 & 8,142 & 1,088 & 10,087 \\
n300-k20-s10 & 4 & 101 & 22 & 413 & 40 & 995 & 70 & 1,700 & 137 & 2,638 & 185 & 3,820 & 273 & 5,292 & 341 & 6,799 & 424 & 8,621 & 525 & 10,650 \\
n300-k20-s20 & 4 & 101 & 22 & 413 & 40 & 995 & 70 & 1,700 & 137 & 2,638 & 185 & 3,820 & 273 & 5,292 & 341 & 6,799 & 424 & 8,621 & 525 & 10,650 \\
n300-k20-s30 & 4 & 101 & 22 & 413 & 40 & 995 & 70 & 1,700 & 137 & 2,638 & 185 & 3,820 & 273 & 5,292 & 341 & 6,799 & 424 & 8,621 & 525 & 10,650 \\
n300-k20-s40 & 4 & 101 & 22 & 413 & 40 & 995 & 70 & 1,700 & 137 & 2,638 & 185 & 3,820 & 273 & 5,292 & 341 & 6,799 & 424 & 8,621 & 525 & 10,650 \\
n300-k20-s50 & 4 & 101 & 22 & 413 & 40 & 995 & 70 & 1,700 & 137 & 2,638 & 185 & 3,820 & 273 & 5,292 & 341 & 6,799 & 424 & 8,621 & 525 & 10,650 \\
n300-k50-s10 & 1 & 104 & 3 & 432 & 19 & 1,016 & 34 & 1,736 & 55 & 2,720 & 48 & 3,957 & 106 & 5,459 & 134 & 7,006 & 158 & 8,887 & 209 & 10,966 \\
n300-k50-s20 & 1 & 104 & 3 & 432 & 19 & 1,016 & 34 & 1,736 & 55 & 2,720 & 48 & 3,957 & 106 & 5,459 & 134 & 7,006 & 158 & 8,887 & 209 & 10,966 \\
n300-k50-s30 & 1 & 104 & 3 & 432 & 19 & 1,016 & 34 & 1,736 & 55 & 2,720 & 48 & 3,957 & 106 & 5,459 & 134 & 7,006 & 158 & 8,887 & 209 & 10,966 \\
n300-k50-s40 & 1 & 104 & 3 & 432 & 19 & 1,016 & 34 & 1,736 & 55 & 2,720 & 48 & 3,957 & 106 & 5,459 & 134 & 7,006 & 158 & 8,887 & 209 & 10,966 \\
n300-k50-s50 & 1 & 104 & 3 & 432 & 19 & 1,016 & 34 & 1,736 & 55 & 2,720 & 48 & 3,957 & 106 & 5,459 & 134 & 7,006 & 158 & 8,887 & 209 & 10,966 \\
\bottomrule
\end{tabular}
}
\end{table}

\begin{table}[H]
\centering
\caption{Number of must-link (ML) and cannot-link (CL) constraints for data sets in collection COL3}
\label{tab:col3-constraints}
\resizebox{0.99\textwidth}{!}{%
\begin{tabular}{c cc cc cc cc}
\toprule
 & \multicolumn{2}{c}{5\% CS} & \multicolumn{2}{c}{10\% CS} &
   \multicolumn{2}{c}{15\% CS} & \multicolumn{2}{c}{20\% CS} \\
Data set & ML & CL & ML & CL & ML & CL & ML & CL \\
\midrule
n500-k2    & 147 & 153 & 607 & 618 & 1,380 & 1,395 & 2,465 & 2,485 \\
n500-k5    & 56  & 244 & 233 & 992 & 594  & 2,181 & 999  & 3,951 \\
n500-k10   & 26  & 274 & 134 & 1,091 & 280 & 2,495 & 482 & 4,460 \\
n500-k20   & 16  & 284 & 58  & 1,167 & 129 & 2,646 & 270 & 4,680 \\
n500-k50   & 6   & 294 & 28  & 1,197 & 45  & 2,730 & 105 & 4,845 \\
n500-k100  & 4   & 296 & 7   & 1,218 & 23  & 2,752 & 46  & 4,904 \\
n1000-k2   & 607 & 618 & 2,494 & 2,456 & 5,559 & 5,616 & 10,007 & 9,893 \\
n1000-k5   & 245 & 980 & 989 & 3,961 & 2,249 & 8,926 & 3,966 & 15,934 \\
n1000-k10  & 127 & 1,098 & 449 & 4,501 & 1,119 & 10,056 & 1,996 & 17,904 \\
n1000-k20  & 50  & 1,175 & 229 & 4,721 & 522 & 10,653 & 929 & 18,971 \\
n1000-k50  & 26  & 1,199 & 91  & 4,859 & 203 & 10,972 & 404 & 19,496 \\
n1000-k100 & 1   & 1,224 & 55 & 4,895 & 104 & 11,071 & 189 & 19,711 \\
n2000-k2   & 2,472 & 2,478 & 9,984 & 9,916 & 22,405 & 22,445 & 39,612 & 40,188 \\
n2000-k5   & 972  & 3,978  & 3,860 & 16,040 & 8,943 & 35,907 & 15,928 & 63,878 \\
n2000-k10  & 472  & 4,478  & 1,978 & 17,922 & 4,405 & 40,445 & 7,950 & 71,850 \\
n2000-k20  & 229  & 4,721  & 950 & 18,950 & 2,175 & 42,675 & 4,009 & 75,791 \\
n2000-k50  & 90   & 4,860  & 363 & 19,537 & 866 & 43,990 & 1,594 & 78,206 \\
n2000-k100 & 58   & 4,992  & 201 & 19,699 & 411 & 44,439 & 807 & 78,993 \\
n5000-k2   & 15,632 & 15,493 & 62,161 & 62,589 & 140,016 & 140,859 & 249,044 & 250,460 \\
n5000-k5   & 6,252  & 24,873 & 24,892 & 99,858 & 56,001 & 224,874 & 99,497 & 400,003 \\
n5000-k10  & 3,106  & 28,019 & 12,458 & 112,292 & 27,954 & 252,921 & 49,307 & 450,193 \\
n5000-k20  & 1,546  & 29,579 & 6,075 & 118,675 & 14,097 & 266,778 & 25,003 & 474,497 \\
n5000-k50  & 607    & 30,518 & 2,440 & 122,310 & 5,536 & 275,339 & 9,762 & 489,738 \\
n5000-k100 & 292    & 30,833 & 1,229 & 123,521 & 2,694 & 278,181 & 4,854 & 494,646 \\
\bottomrule
\end{tabular}
}
\end{table}

\begin{table}[H]
\centering
\caption{Number of must-link (ML) and cannot-link (CL) constraints in the different constraint sets of collection COL4}
\label{tab:col4-constraints}
\resizebox{0.99\textwidth}{!}{%
\begin{tabular}{c cc cc cc}
\toprule
 & \multicolumn{2}{c}{0.5\% CS} & \multicolumn{2}{c}{1\% CS} &
   \multicolumn{2}{c}{5\% CS} \\
Data set & ML & CL & ML & CL & ML & CL \\
\midrule
Banana    & 160   & 191     & 707     & 671       & 17,797    & 17,183 \\
Letter    & 190   & 4,760   & 743     & 19,157    & 19,165    & 480,335 \\
Shuttle   & 27,099 & 14,806  & 108,254 & 59,656    & 2,707,873 & 1,495,677 \\
Cifar 10  & 4,511  & 40,339  & 17,840  & 161,860   & 450,153   & 4,048,347 \\
Cifar 100 & 467    & 44,383  & 1,760   & 177,940   & 44,792    & 4,453,708 \\
Mnist     & 6,074  & 55,001  & 24,592  & 220,058   & 612,630   & 5,510,620 \\
\bottomrule
\end{tabular}
}
\end{table}

\subsection{Detailed Numerical Results}
\label{app:numerical}
\subsubsection{General Results}
\begin{table}[H]
\caption{Clustering inertia on dataset collection 1 with 5.0\% constraints}
\label{tab:obj_COL1_5}
\centering
\resizebox{0.9\textwidth}{!}{%
\begin{tabular}{c|cccccccc}
\toprule
Constraints Level & \multicolumn{8}{c}{0.05} \\
Method & COPKM\_P & KSKM & KSKM\_E & KSKM\_repo & KSKM\_E\_repo & BLPKMCC & PCCC & COPKM \\
Dataset &  &  &  &  &  &  &  &  \\
\midrule
appendicitis & 492.08 & 491.86 & 491.83 & 492.08 & 492.08 & 492.07 & 493.26 & 503.87 \\
breast\_cancer & 12473.22 & 12084.17 & 12084.17 & 12084.17 & 12084.17 & 12084.17 & 12084.17 & - \\
bupa & 1876.10 & 1849.95 & 1849.95 & 1850.31 & 1850.65 & 1851.36 & 1852.94 & - \\
circles & 501.02 & 477.16 & 474.66 & 477.27 & 477.22 & 476.00 & 476.21 & - \\
ecoli & 819.00 & 796.38 & 799.74 & 802.22 & 801.50 & 1023.05 & 824.42 & 830.42 \\
\bottomrule
\end{tabular}
}
\end{table}
\begin{table}[H]
\caption{CPU time on dataset collection 1 with 5.0\% constraints}
\label{tab:time_COL1_5}
\centering
\resizebox{0.9\textwidth}{!}{%
\begin{tabular}{c|cccccccc}
\toprule
Constraints Level & \multicolumn{8}{c}{0.05} \\
Method & COPKM\_P & KSKM & KSKM\_E & KSKM\_repo & KSKM\_E\_repo & BLPKMCC & PCCC & COPKM \\
Dataset &  &  &  &  &  &  &  &  \\
\midrule
appendicitis & 0.00 & 0.00 & 0.01 & 0.01 & 0.00 & 0.02 & 0.05 & 0.09 \\
breast\_cancer & 0.01 & 0.01 & 0.01 & 0.01 & 0.02 & 0.05 & 0.10 & - \\
bupa & 0.01 & 0.01 & 0.03 & 0.02 & 0.03 & 0.06 & 0.10 & - \\
circles & 0.01 & 0.01 & 0.04 & 0.01 & 0.02 & 0.04 & 0.08 & - \\
ecoli & 0.01 & 0.12 & 0.34 & 0.08 & 0.22 & 0.43 & 0.83 & 0.01 \\
glass & 0.00 & 0.05 & 0.08 & 0.03 & 0.06 & 0.16 & 0.27 & 0.01 \\
haberman & 0.01 & 0.01 & 0.05 & 0.01 & 0.02 & 0.07 & 0.12 & - \\
hayes-roth & 0.00 & 0.01 & 0.03 & 0.01 & 0.02 & 0.05 & 0.11 & 0.00 \\
heart & 0.01 & 0.01 & 0.03 & 0.01 & 0.02 & 0.03 & 0.11 & - \\
ionosphere & 0.01 & 0.01 & 0.02 & 0.01 & 0.02 & 0.03 & 0.12 & - \\
led7digit & 0.01 & 0.15 & 0.98 & 0.15 & 0.44 & 0.35 & 0.72 & 0.01 \\
monk-2 & 0.01 & 0.02 & 0.03 & 0.03 & 0.05 & 0.09 & 0.15 & - \\
moons & 0.01 & 0.01 & 0.01 & 0.01 & 0.01 & 0.03 & 0.11 & 0.01 \\
movement\_libras & 0.01 & 0.10 & 0.60 & 0.12 & 0.42 & 0.63 & 1.57 & 0.02 \\
newthyroid & 0.00 & 0.02 & 0.02 & 0.03 & 0.04 & 0.06 & 0.16 & 0.00 \\
saheart & 0.01 & 0.01 & 0.03 & 0.01 & 0.03 & 0.06 & 0.12 & - \\
sonar & 0.00 & 0.01 & 0.02 & 0.01 & 0.01 & 0.02 & 0.09 & - \\
soybean & 0.00 & 0.00 & 0.00 & 0.00 & 0.00 & 0.01 & 0.03 & 0.00 \\
spectfheart & 0.00 & 0.01 & 0.03 & 0.01 & 0.01 & 0.04 & 0.08 & - \\
spiral & 0.01 & 0.01 & 0.02 & 0.02 & 0.02 & 0.04 & 0.12 & - \\
tae & 0.00 & 0.02 & 0.03 & 0.02 & 0.02 & 0.04 & 0.12 & 0.00 \\
vehicle & 0.02 & 0.13 & 0.56 & 0.08 & 0.27 & 0.59 & 1.67 & - \\
zoo & 0.00 & 0.01 & 0.01 & 0.00 & 0.01 & 0.03 & 0.09 & 0.00 \\
\bottomrule
\end{tabular}
}
\end{table}
\begin{table}[H]
\caption{ARI on dataset collection 1 with 5.0\% constraints}
\label{tab:ARI_COL1_5}
\centering
\resizebox{0.9\textwidth}{!}{%
\begin{tabular}{c|cccccccc}
\toprule
Constraints Level & \multicolumn{8}{c}{0.05} \\
Method & COPKM\_P & KSKM & KSKM\_E & KSKM\_repo & KSKM\_E\_repo & BLPKMCC & PCCC & COPKM \\
Dataset &  &  &  &  &  &  &  &  \\
\midrule
appendicitis & 0.40 & 0.50 & 0.54 & 0.40 & 0.40 & 0.40 & 0.47 & 0.30 \\
breast\_cancer & 0.86 & 0.93 & 0.93 & 0.93 & 0.93 & 0.93 & 0.93 & - \\
bupa & 0.02 & -0.00 & -0.00 & -0.00 & -0.00 & -0.00 & 0.00 & - \\
circles & -0.00 & -0.00 & -0.00 & -0.00 & -0.00 & -0.00 & -0.00 & - \\
ecoli & 0.46 & 0.37 & 0.41 & 0.38 & 0.41 & 0.36 & 0.35 & 0.43 \\
glass & 0.24 & 0.25 & 0.26 & 0.26 & 0.25 & 0.23 & 0.21 & 0.22 \\
haberman & 0.09 & 0.19 & 0.19 & 0.18 & 0.16 & 0.15 & 0.11 & - \\
hayes-roth & 0.08 & 0.11 & 0.11 & 0.12 & 0.13 & 0.09 & 0.09 & 0.08 \\
heart & 0.59 & 0.69 & 0.69 & 0.67 & 0.67 & 0.67 & 0.61 & - \\
ionosphere & 0.15 & 0.35 & 0.35 & 0.35 & 0.35 & 0.35 & 0.37 & - \\
led7digit & 0.44 & 0.49 & 0.49 & 0.48 & 0.47 & 0.41 & 0.48 & 0.44 \\
monk-2 & 0.58 & 0.73 & 0.73 & 0.73 & 0.73 & 0.71 & 0.72 & - \\
moons & 0.68 & 0.69 & 0.69 & 0.70 & 0.70 & 0.70 & 0.70 & 0.62 \\
movement\_libras & 0.30 & 0.31 & 0.31 & 0.32 & 0.31 & 0.28 & 0.31 & 0.30 \\
newthyroid & 0.75 & 0.81 & 0.81 & 0.81 & 0.81 & 0.79 & 0.81 & 0.73 \\
saheart & 0.37 & 0.44 & 0.44 & 0.43 & 0.43 & 0.43 & 0.44 & - \\
sonar & 0.06 & 0.03 & 0.04 & 0.04 & 0.05 & 0.04 & 0.06 & - \\
soybean & 0.87 & 1.00 & 1.00 & 1.00 & 1.00 & 0.72 & 0.91 & 0.87 \\
spectfheart & 0.01 & 0.04 & 0.04 & 0.04 & 0.04 & 0.03 & 0.05 & - \\
spiral & 0.07 & 0.13 & 0.13 & 0.13 & 0.13 & 0.06 & 0.09 & - \\
tae & 0.04 & 0.06 & 0.05 & 0.06 & 0.06 & 0.05 & 0.06 & 0.03 \\
vehicle & 0.09 & 0.09 & 0.12 & 0.10 & 0.11 & 0.12 & 0.13 & - \\
zoo & 0.73 & 0.71 & 0.74 & 0.76 & 0.77 & 0.62 & 0.75 & 0.73 \\
\bottomrule
\end{tabular}
}
\end{table}
\begin{table}[H]
\caption{Algorithm success rate on dataset collection 1 with 5.0\% constraints}
\label{tab:rate_success_COL1_5}
\centering
\resizebox{0.9\textwidth}{!}{%
\begin{tabular}{c|cccccccc}
\toprule
Constraints Level & \multicolumn{8}{c}{0.05} \\
Method & COPKM\_P & KSKM & KSKM\_E & KSKM\_repo & KSKM\_E\_repo & BLPKMCC & PCCC & COPKM \\
Dataset &  &  &  &  &  &  &  &  \\
\midrule
appendicitis & 1.00 & 1.00 & 1.00 & 1.00 & 1.00 & 1.00 & 1.00 & 1.00 \\
breast\_cancer & 1.00 & 1.00 & 1.00 & 1.00 & 1.00 & 1.00 & 1.00 & 0.00 \\
bupa & 1.00 & 1.00 & 1.00 & 1.00 & 1.00 & 1.00 & 1.00 & 0.00 \\
circles & 1.00 & 1.00 & 1.00 & 1.00 & 1.00 & 1.00 & 1.00 & 0.00 \\
ecoli & 1.00 & 1.00 & 1.00 & 1.00 & 1.00 & 1.00 & 1.00 & 1.00 \\
glass & 1.00 & 1.00 & 1.00 & 1.00 & 1.00 & 1.00 & 1.00 & 1.00 \\
haberman & 1.00 & 1.00 & 1.00 & 1.00 & 1.00 & 1.00 & 1.00 & 0.00 \\
hayes-roth & 1.00 & 1.00 & 1.00 & 1.00 & 1.00 & 1.00 & 1.00 & 1.00 \\
heart & 1.00 & 1.00 & 1.00 & 1.00 & 1.00 & 1.00 & 1.00 & 0.00 \\
ionosphere & 1.00 & 1.00 & 1.00 & 1.00 & 1.00 & 1.00 & 1.00 & 0.00 \\
led7digit & 1.00 & 1.00 & 1.00 & 1.00 & 1.00 & 1.00 & 1.00 & 1.00 \\
monk-2 & 1.00 & 1.00 & 1.00 & 1.00 & 1.00 & 1.00 & 1.00 & 0.00 \\
moons & 1.00 & 1.00 & 1.00 & 1.00 & 1.00 & 1.00 & 1.00 & 0.70 \\
movement\_libras & 1.00 & 1.00 & 1.00 & 1.00 & 1.00 & 1.00 & 1.00 & 1.00 \\
newthyroid & 1.00 & 1.00 & 1.00 & 1.00 & 1.00 & 1.00 & 1.00 & 1.00 \\
saheart & 1.00 & 1.00 & 1.00 & 1.00 & 1.00 & 1.00 & 1.00 & 0.00 \\
sonar & 1.00 & 1.00 & 1.00 & 1.00 & 1.00 & 1.00 & 1.00 & 0.00 \\
soybean & 1.00 & 1.00 & 1.00 & 1.00 & 1.00 & 1.00 & 1.00 & 1.00 \\
spectfheart & 1.00 & 1.00 & 1.00 & 1.00 & 1.00 & 1.00 & 1.00 & 0.00 \\
spiral & 1.00 & 1.00 & 1.00 & 1.00 & 1.00 & 1.00 & 1.00 & 0.00 \\
tae & 1.00 & 1.00 & 1.00 & 1.00 & 1.00 & 1.00 & 1.00 & 1.00 \\
vehicle & 1.00 & 1.00 & 1.00 & 1.00 & 1.00 & 1.00 & 1.00 & 0.00 \\
zoo & 1.00 & 1.00 & 1.00 & 1.00 & 1.00 & 1.00 & 1.00 & 1.00 \\
\bottomrule
\end{tabular}
}
\end{table}
\begin{table}[H]
\caption{Clustering inertia on dataset collection 1 with 10.0\% constraints}
\label{tab:obj_COL1_10}
\centering
\resizebox{0.9\textwidth}{!}{%
\begin{tabular}{c|cccccccc}
\toprule
Constraints Level & \multicolumn{8}{c}{0.1} \\
Method & COPKM\_P & KSKM & KSKM\_E & KSKM\_repo & KSKM\_E\_repo & BLPKMCC & PCCC & COPKM \\
Dataset &  &  &  &  &  &  &  &  \\
\midrule
appendicitis & 544.23 & 544.18 & 544.18 & 544.23 & 544.23 & 544.24 & 544.98 & - \\
breast\_cancer & 12185.03 & 12185.03 & 12185.03 & 12185.03 & 12185.03 & 12185.03 & 12185.03 & 12185.03 \\
bupa & 2041.65 & 2041.64 & 2041.64 & 2041.65 & 2041.65 & 2041.69 & 2041.66 & - \\
circles & 598.83 & 598.80 & 598.80 & 598.84 & 598.83 & 598.83 & 598.98 & - \\
ecoli & 967.28 & 954.31 & 954.12 & 961.39 & 958.11 & 1180.93 & 957.88 & 1140.72 \\
\bottomrule
\end{tabular}
}
\end{table}
\begin{table}[H]
\caption{CPU time on dataset collection 1 with 10.0\% constraints}
\label{tab:time_COL1_10}
\centering
\resizebox{0.9\textwidth}{!}{%
\begin{tabular}{c|cccccccc}
\toprule
Constraints Level & \multicolumn{8}{c}{0.1} \\
Method & COPKM\_P & KSKM & KSKM\_E & KSKM\_repo & KSKM\_E\_repo & BLPKMCC & PCCC & COPKM \\
Dataset &  &  &  &  &  &  &  &  \\
\midrule
appendicitis & 0.02 & 0.00 & 0.01 & 0.01 & 0.01 & 0.01 & 0.03 & - \\
breast\_cancer & 0.00 & 0.00 & 0.00 & 0.00 & 0.00 & 0.07 & 0.01 & 0.00 \\
bupa & 0.00 & 0.00 & 0.01 & 0.00 & 0.01 & 0.04 & 0.03 & - \\
circles & 0.00 & 0.01 & 0.01 & 0.00 & 0.01 & 0.04 & 0.03 & - \\
ecoli & 0.01 & 0.08 & 0.60 & 0.04 & 0.32 & 0.51 & 0.77 & 0.00 \\
glass & 0.00 & 0.09 & 0.35 & 0.06 & 0.22 & 0.17 & 0.39 & 0.00 \\
haberman & 0.00 & 0.00 & 0.01 & 0.00 & 0.01 & 0.04 & 0.02 & - \\
hayes-roth & 0.00 & 0.05 & 0.11 & 0.03 & 0.08 & 0.06 & 0.18 & - \\
heart & 0.00 & 0.00 & 0.01 & 0.01 & 0.01 & 0.03 & 0.04 & 0.00 \\
ionosphere & 0.00 & 0.00 & 0.01 & 0.00 & 0.01 & 0.04 & 0.04 & 0.00 \\
led7digit & 0.01 & 0.13 & 1.73 & 0.06 & 1.07 & 1.34 & 2.13 & 0.01 \\
monk-2 & 0.00 & 0.00 & 0.01 & 0.00 & 0.01 & 0.06 & 0.03 & 0.00 \\
moons & 0.00 & 0.01 & 0.01 & 0.01 & 0.01 & 0.04 & 0.04 & - \\
movement\_libras & 0.01 & 0.07 & 1.11 & 0.06 & 0.64 & 1.28 & 3.18 & 0.02 \\
newthyroid & 0.00 & 0.01 & 0.01 & 0.02 & 0.03 & 0.04 & 0.08 & 0.00 \\
saheart & 0.00 & 0.00 & 0.01 & 0.00 & 0.01 & 0.06 & 0.02 & 0.00 \\
sonar & 0.00 & 0.00 & 0.01 & 0.01 & 0.01 & 0.03 & 0.04 & - \\
soybean & 0.00 & 0.00 & 0.00 & 0.00 & 0.00 & 0.01 & 0.03 & 0.00 \\
spectfheart & 0.00 & 0.00 & 0.00 & 0.00 & 0.01 & 0.03 & 0.03 & - \\
spiral & 0.00 & 0.00 & 0.01 & 0.00 & 0.01 & 0.03 & 0.03 & - \\
tae & 0.00 & 0.04 & 0.09 & 0.03 & 0.09 & 0.05 & 0.19 & - \\
vehicle & 0.01 & 0.01 & 0.02 & 0.01 & 0.02 & 0.37 & 0.17 & - \\
zoo & 0.00 & 0.01 & 0.04 & 0.01 & 0.03 & 0.03 & 0.14 & 0.00 \\
\bottomrule
\end{tabular}
}
\end{table}
\begin{table}[H]
\caption{ARI on dataset collection 1 with 10.0\% constraints}
\label{tab:ARI_COL1_10}
\centering
\resizebox{0.9\textwidth}{!}{%
\begin{tabular}{c|cccccccc}
\toprule
Constraints Level & \multicolumn{8}{c}{0.1} \\
Method & COPKM\_P & KSKM & KSKM\_E & KSKM\_repo & KSKM\_E\_repo & BLPKMCC & PCCC & COPKM \\
Dataset &  &  &  &  &  &  &  &  \\
\midrule
appendicitis & 0.61 & 0.62 & 0.62 & 0.61 & 0.61 & 0.60 & 0.61 & - \\
breast\_cancer & 0.99 & 0.99 & 0.99 & 0.99 & 0.99 & 0.99 & 0.99 & 0.99 \\
bupa & 0.87 & 0.87 & 0.87 & 0.87 & 0.87 & 0.88 & 0.87 & - \\
circles & 0.85 & 0.85 & 0.85 & 0.85 & 0.85 & 0.85 & 0.83 & - \\
ecoli & 0.69 & 0.71 & 0.72 & 0.71 & 0.70 & 0.69 & 0.73 & 0.60 \\
glass & 0.19 & 0.20 & 0.25 & 0.21 & 0.22 & 0.24 & 0.27 & 0.28 \\
haberman & 0.92 & 0.93 & 0.93 & 0.92 & 0.92 & 0.92 & 0.90 & - \\
hayes-roth & 0.08 & 0.17 & 0.17 & 0.18 & 0.17 & 0.16 & 0.16 & - \\
heart & 0.91 & 0.91 & 0.91 & 0.91 & 0.91 & 0.91 & 0.88 & 0.91 \\
ionosphere & 0.95 & 0.95 & 0.95 & 0.95 & 0.95 & 0.95 & 0.95 & 0.95 \\
led7digit & 0.52 & 0.61 & 0.63 & 0.59 & 0.63 & 0.58 & 0.67 & 0.52 \\
monk-2 & 0.97 & 0.97 & 0.97 & 0.97 & 0.97 & 0.97 & 0.97 & 0.97 \\
moons & 0.99 & 0.99 & 0.99 & 0.99 & 0.99 & 0.99 & 0.99 & - \\
movement\_libras & 0.32 & 0.33 & 0.34 & 0.33 & 0.34 & 0.32 & 0.33 & 0.31 \\
newthyroid & 0.93 & 0.92 & 0.92 & 0.92 & 0.92 & 0.93 & 0.93 & 0.93 \\
saheart & 0.98 & 0.97 & 0.97 & 0.97 & 0.98 & 0.98 & 0.98 & 0.97 \\
sonar & 0.74 & 0.75 & 0.75 & 0.75 & 0.75 & 0.74 & 0.74 & - \\
soybean & 0.96 & 1.00 & 1.00 & 1.00 & 1.00 & 0.71 & 1.00 & 0.96 \\
spectfheart & 0.86 & 0.86 & 0.86 & 0.86 & 0.86 & 0.86 & 0.86 & - \\
spiral & 0.88 & 0.88 & 0.88 & 0.88 & 0.88 & 0.88 & 0.88 & - \\
tae & 0.08 & 0.15 & 0.20 & 0.15 & 0.19 & 0.12 & 0.16 & - \\
vehicle & 0.96 & 0.96 & 0.96 & 0.96 & 0.96 & 0.96 & 0.96 & - \\
zoo & 0.76 & 0.77 & 0.77 & 0.76 & 0.74 & 0.67 & 0.78 & 0.76 \\
\bottomrule
\end{tabular}
}
\end{table}
\begin{table}[H]
\caption{Algorithm success rate on dataset collection 1 with 10.0\% constraints}
\label{tab:rate_success_COL1_10}
\centering
\resizebox{0.9\textwidth}{!}{%
\begin{tabular}{c|cccccccc}
\toprule
Constraints Level & \multicolumn{8}{c}{0.1} \\
Method & COPKM\_P & KSKM & KSKM\_E & KSKM\_repo & KSKM\_E\_repo & BLPKMCC & PCCC & COPKM \\
Dataset &  &  &  &  &  &  &  &  \\
\midrule
appendicitis & 1.00 & 1.00 & 1.00 & 1.00 & 1.00 & 1.00 & 1.00 & 0.00 \\
breast\_cancer & 1.00 & 1.00 & 1.00 & 1.00 & 1.00 & 1.00 & 1.00 & 0.30 \\
bupa & 1.00 & 1.00 & 1.00 & 1.00 & 1.00 & 1.00 & 1.00 & 0.00 \\
circles & 1.00 & 1.00 & 1.00 & 1.00 & 1.00 & 1.00 & 1.00 & 0.00 \\
ecoli & 1.00 & 1.00 & 1.00 & 1.00 & 1.00 & 1.00 & 1.00 & 1.00 \\
glass & 1.00 & 1.00 & 1.00 & 1.00 & 1.00 & 1.00 & 1.00 & 1.00 \\
haberman & 1.00 & 1.00 & 1.00 & 1.00 & 1.00 & 1.00 & 1.00 & 0.00 \\
hayes-roth & 1.00 & 1.00 & 1.00 & 1.00 & 1.00 & 1.00 & 1.00 & 0.00 \\
heart & 1.00 & 1.00 & 1.00 & 1.00 & 1.00 & 1.00 & 1.00 & 0.20 \\
ionosphere & 1.00 & 1.00 & 1.00 & 1.00 & 1.00 & 1.00 & 1.00 & 0.10 \\
led7digit & 1.00 & 1.00 & 1.00 & 1.00 & 1.00 & 1.00 & 1.00 & 0.90 \\
monk-2 & 1.00 & 1.00 & 1.00 & 1.00 & 1.00 & 1.00 & 1.00 & 1.00 \\
moons & 1.00 & 1.00 & 1.00 & 1.00 & 1.00 & 1.00 & 1.00 & 0.00 \\
movement\_libras & 1.00 & 1.00 & 1.00 & 1.00 & 1.00 & 1.00 & 1.00 & 1.00 \\
newthyroid & 1.00 & 1.00 & 1.00 & 1.00 & 1.00 & 1.00 & 1.00 & 0.70 \\
saheart & 1.00 & 1.00 & 1.00 & 1.00 & 1.00 & 1.00 & 1.00 & 1.00 \\
sonar & 1.00 & 1.00 & 1.00 & 1.00 & 1.00 & 1.00 & 1.00 & 0.00 \\
soybean & 1.00 & 1.00 & 1.00 & 1.00 & 1.00 & 1.00 & 1.00 & 1.00 \\
spectfheart & 1.00 & 1.00 & 1.00 & 1.00 & 1.00 & 1.00 & 1.00 & 0.00 \\
spiral & 1.00 & 1.00 & 1.00 & 1.00 & 1.00 & 1.00 & 1.00 & 0.00 \\
tae & 1.00 & 1.00 & 1.00 & 1.00 & 1.00 & 1.00 & 1.00 & 0.00 \\
vehicle & 1.00 & 1.00 & 1.00 & 1.00 & 1.00 & 1.00 & 1.00 & 0.00 \\
zoo & 1.00 & 1.00 & 1.00 & 1.00 & 1.00 & 1.00 & 1.00 & 1.00 \\
\bottomrule
\end{tabular}
}
\end{table}
\begin{table}[H]
\caption{Clustering inertia on dataset collection 1 with 15.0\% constraints}
\label{tab:obj_COL1_15}
\centering
\resizebox{0.9\textwidth}{!}{%
\begin{tabular}{c|cccccccc}
\toprule
Constraints Level & \multicolumn{8}{c}{0.15} \\
Method & COPKM\_P & KSKM & KSKM\_E & KSKM\_repo & KSKM\_E\_repo & BLPKMCC & PCCC & COPKM \\
Dataset &  &  &  &  &  &  &  &  \\
\midrule
appendicitis & 612.93 & 612.93 & 612.93 & 612.93 & 612.93 & 612.93 & 612.93 & - \\
breast\_cancer & 12214.56 & 12214.56 & 12214.56 & 12214.56 & 12214.56 & 12214.56 & 12214.56 & 12214.56 \\
bupa & 2047.30 & 2047.30 & 2047.30 & 2047.30 & 2047.30 & 2047.30 & 2047.30 & 2047.30 \\
circles & 599.99 & 599.99 & 599.99 & 599.99 & 599.99 & 599.99 & 599.99 & 599.99 \\
ecoli & 1034.85 & 1030.49 & 1027.78 & 1044.68 & 1032.17 & 1149.84 & 1032.63 & - \\
\bottomrule
\end{tabular}
}
\end{table}
\begin{table}[H]
\caption{CPU time on dataset collection 1 with 15.0\% constraints}
\label{tab:time_COL1_15}
\centering
\resizebox{0.9\textwidth}{!}{%
\begin{tabular}{c|cccccccc}
\toprule
Constraints Level & \multicolumn{8}{c}{0.15} \\
Method & COPKM\_P & KSKM & KSKM\_E & KSKM\_repo & KSKM\_E\_repo & BLPKMCC & PCCC & COPKM \\
Dataset &  &  &  &  &  &  &  &  \\
\midrule
appendicitis & 0.00 & 0.00 & 0.00 & 0.00 & 0.00 & 0.01 & 0.02 & - \\
breast\_cancer & 0.00 & 0.00 & 0.00 & 0.00 & 0.00 & 0.10 & 0.01 & 0.00 \\
bupa & 0.00 & 0.00 & 0.00 & 0.00 & 0.00 & 0.04 & 0.01 & 0.00 \\
circles & 0.00 & 0.00 & 0.00 & 0.00 & 0.00 & 0.03 & 0.01 & 0.00 \\
ecoli & 0.00 & 0.04 & 0.19 & 0.05 & 0.23 & 0.62 & 0.43 & - \\
glass & 0.00 & 0.05 & 0.26 & 0.03 & 0.13 & 0.23 & 0.32 & 0.00 \\
haberman & 0.00 & 0.00 & 0.00 & 0.00 & 0.00 & 0.03 & 0.01 & 0.00 \\
hayes-roth & 0.00 & 0.00 & 0.01 & 0.00 & 0.01 & 0.04 & 0.07 & - \\
heart & 0.00 & 0.00 & 0.00 & 0.00 & 0.00 & 0.03 & 0.01 & 0.00 \\
ionosphere & 0.00 & 0.00 & 0.00 & 0.00 & 0.00 & 0.04 & 0.01 & 0.00 \\
led7digit & 0.01 & 0.06 & 1.04 & 0.04 & 0.41 & 1.68 & 2.50 & - \\
monk-2 & 0.00 & 0.00 & 0.00 & 0.00 & 0.00 & 0.07 & 0.01 & 0.00 \\
moons & 0.00 & 0.00 & 0.00 & 0.00 & 0.00 & 0.04 & 0.01 & 0.00 \\
movement\_libras & 0.01 & 0.11 & 2.04 & 0.06 & 0.91 & 2.54 & 4.24 & 0.01 \\
newthyroid & 0.00 & 0.01 & 0.01 & 0.01 & 0.01 & 0.05 & 0.03 & 0.00 \\
saheart & 0.00 & 0.00 & 0.00 & 0.00 & 0.00 & 0.07 & 0.01 & 0.00 \\
sonar & 0.00 & 0.00 & 0.00 & 0.00 & 0.00 & 0.02 & 0.01 & - \\
soybean & 0.00 & 0.01 & 0.01 & 0.01 & 0.02 & 0.02 & 0.04 & 0.00 \\
spectfheart & 0.00 & 0.00 & 0.00 & 0.00 & 0.00 & 0.04 & 0.02 & 0.00 \\
spiral & 0.00 & 0.00 & 0.00 & 0.00 & 0.00 & 0.03 & 0.01 & 0.00 \\
tae & 0.00 & 0.01 & 0.03 & 0.01 & 0.02 & 0.05 & 0.09 & - \\
vehicle & 0.00 & 0.00 & 0.00 & 0.00 & 0.00 & 0.42 & 0.03 & 0.00 \\
zoo & 0.00 & 0.01 & 0.13 & 0.01 & 0.02 & 0.06 & 0.13 & 0.00 \\
\bottomrule
\end{tabular}
}
\end{table}
\begin{table}[H]
\caption{ARI on dataset collection 1 with 15.0\% constraints}
\label{tab:ARI_COL1_15}
\centering
\resizebox{0.9\textwidth}{!}{%
\begin{tabular}{c|cccccccc}
\toprule
Constraints Level & \multicolumn{8}{c}{0.15} \\
Method & COPKM\_P & KSKM & KSKM\_E & KSKM\_repo & KSKM\_E\_repo & BLPKMCC & PCCC & COPKM \\
Dataset &  &  &  &  &  &  &  &  \\
\midrule
appendicitis & 1.00 & 1.00 & 1.00 & 1.00 & 1.00 & 1.00 & 1.00 & - \\
breast\_cancer & 1.00 & 1.00 & 1.00 & 1.00 & 1.00 & 1.00 & 1.00 & 1.00 \\
bupa & 1.00 & 1.00 & 1.00 & 1.00 & 1.00 & 1.00 & 1.00 & 1.00 \\
circles & 1.00 & 1.00 & 1.00 & 1.00 & 1.00 & 1.00 & 1.00 & 1.00 \\
ecoli & 0.92 & 0.92 & 0.92 & 0.92 & 0.92 & 0.93 & 0.93 & - \\
glass & 0.79 & 0.83 & 0.79 & 0.80 & 0.81 & 0.80 & 0.80 & 0.69 \\
haberman & 1.00 & 1.00 & 1.00 & 1.00 & 1.00 & 1.00 & 1.00 & 1.00 \\
hayes-roth & 0.98 & 0.98 & 0.98 & 0.98 & 0.98 & 0.98 & 0.97 & - \\
heart & 1.00 & 1.00 & 1.00 & 1.00 & 1.00 & 1.00 & 1.00 & 1.00 \\
ionosphere & 1.00 & 1.00 & 1.00 & 1.00 & 1.00 & 1.00 & 1.00 & 1.00 \\
led7digit & 0.80 & 0.84 & 0.89 & 0.82 & 0.89 & 0.87 & 0.88 & - \\
monk-2 & 1.00 & 1.00 & 1.00 & 1.00 & 1.00 & 1.00 & 1.00 & 1.00 \\
moons & 1.00 & 1.00 & 1.00 & 1.00 & 1.00 & 1.00 & 1.00 & 1.00 \\
movement\_libras & 0.30 & 0.34 & 0.36 & 0.33 & 0.34 & 0.34 & 0.34 & 0.29 \\
newthyroid & 1.00 & 1.00 & 1.00 & 1.00 & 1.00 & 1.00 & 1.00 & 1.00 \\
saheart & 1.00 & 1.00 & 1.00 & 1.00 & 1.00 & 1.00 & 1.00 & 1.00 \\
sonar & 1.00 & 1.00 & 1.00 & 1.00 & 1.00 & 1.00 & 1.00 & - \\
soybean & 0.91 & 1.00 & 0.96 & 1.00 & 1.00 & 0.83 & 1.00 & 0.96 \\
spectfheart & 1.00 & 1.00 & 1.00 & 1.00 & 1.00 & 1.00 & 1.00 & 1.00 \\
spiral & 1.00 & 1.00 & 1.00 & 1.00 & 1.00 & 1.00 & 1.00 & 1.00 \\
tae & 0.66 & 0.72 & 0.73 & 0.72 & 0.71 & 0.71 & 0.69 & - \\
vehicle & 1.00 & 1.00 & 1.00 & 1.00 & 1.00 & 1.00 & 1.00 & 1.00 \\
zoo & 0.83 & 0.84 & 0.84 & 0.83 & 0.83 & 0.77 & 0.92 & 0.84 \\
\bottomrule
\end{tabular}
}
\end{table}
\begin{table}[H]
\caption{Algorithm success rate on dataset collection 1 with 15.0\% constraints}
\label{tab:rate_success_COL1_15}
\centering
\resizebox{0.9\textwidth}{!}{%
\begin{tabular}{c|cccccccc}
\toprule
Constraints Level & \multicolumn{8}{c}{0.15} \\
Method & COPKM\_P & KSKM & KSKM\_E & KSKM\_repo & KSKM\_E\_repo & BLPKMCC & PCCC & COPKM \\
Dataset &  &  &  &  &  &  &  &  \\
\midrule
appendicitis & 1.00 & 1.00 & 1.00 & 1.00 & 1.00 & 1.00 & 1.00 & 0.00 \\
breast\_cancer & 1.00 & 1.00 & 1.00 & 1.00 & 1.00 & 1.00 & 1.00 & 0.80 \\
bupa & 1.00 & 1.00 & 1.00 & 1.00 & 1.00 & 1.00 & 1.00 & 1.00 \\
circles & 1.00 & 1.00 & 1.00 & 1.00 & 1.00 & 1.00 & 1.00 & 1.00 \\
ecoli & 1.00 & 1.00 & 1.00 & 1.00 & 1.00 & 1.00 & 1.00 & 0.00 \\
glass & 1.00 & 1.00 & 1.00 & 1.00 & 1.00 & 1.00 & 1.00 & 0.50 \\
haberman & 1.00 & 1.00 & 1.00 & 1.00 & 1.00 & 1.00 & 1.00 & 1.00 \\
hayes-roth & 1.00 & 1.00 & 1.00 & 1.00 & 1.00 & 1.00 & 1.00 & 0.00 \\
heart & 1.00 & 1.00 & 1.00 & 1.00 & 1.00 & 1.00 & 1.00 & 1.00 \\
ionosphere & 1.00 & 1.00 & 1.00 & 1.00 & 1.00 & 1.00 & 1.00 & 1.00 \\
led7digit & 1.00 & 1.00 & 1.00 & 1.00 & 1.00 & 1.00 & 1.00 & 0.00 \\
monk-2 & 1.00 & 1.00 & 1.00 & 1.00 & 1.00 & 1.00 & 1.00 & 1.00 \\
moons & 1.00 & 1.00 & 1.00 & 1.00 & 1.00 & 1.00 & 1.00 & 1.00 \\
movement\_libras & 1.00 & 1.00 & 1.00 & 1.00 & 1.00 & 1.00 & 1.00 & 0.30 \\
newthyroid & 1.00 & 1.00 & 1.00 & 1.00 & 1.00 & 1.00 & 1.00 & 0.40 \\
saheart & 1.00 & 1.00 & 1.00 & 1.00 & 1.00 & 1.00 & 1.00 & 1.00 \\
sonar & 1.00 & 1.00 & 1.00 & 1.00 & 1.00 & 1.00 & 1.00 & 0.00 \\
soybean & 1.00 & 1.00 & 1.00 & 1.00 & 1.00 & 1.00 & 1.00 & 1.00 \\
spectfheart & 1.00 & 1.00 & 1.00 & 1.00 & 1.00 & 1.00 & 1.00 & 1.00 \\
spiral & 1.00 & 1.00 & 1.00 & 1.00 & 1.00 & 1.00 & 1.00 & 0.50 \\
tae & 1.00 & 1.00 & 1.00 & 1.00 & 1.00 & 1.00 & 1.00 & 0.00 \\
vehicle & 1.00 & 1.00 & 1.00 & 1.00 & 1.00 & 1.00 & 1.00 & 1.00 \\
zoo & 1.00 & 1.00 & 1.00 & 1.00 & 1.00 & 1.00 & 1.00 & 1.00 \\
\bottomrule
\end{tabular}
}
\end{table}
\begin{table}[H]
\caption{Clustering inertia on dataset collection 1 with 20.0\% constraints}
\label{tab:obj_COL1_20}
\centering
\resizebox{0.9\textwidth}{!}{%
\begin{tabular}{c|cccccccc}
\toprule
Constraints Level & \multicolumn{8}{c}{0.2} \\
Method & COPKM\_P & KSKM & KSKM\_E & KSKM\_repo & KSKM\_E\_repo & BLPKMCC & PCCC & COPKM \\
Dataset &  &  &  &  &  &  &  &  \\
\midrule
appendicitis & 612.93 & 612.93 & 612.93 & 612.93 & 612.93 & 612.93 & 612.93 & - \\
breast\_cancer & 12214.56 & 12214.56 & 12214.56 & 12214.56 & 12214.56 & 12214.56 & 12214.56 & 12214.56 \\
bupa & 2047.30 & 2047.30 & 2047.30 & 2047.30 & 2047.30 & 2047.30 & 2047.30 & 2047.30 \\
circles & 599.99 & 599.99 & 599.99 & 599.99 & 599.99 & 599.99 & 599.99 & 599.99 \\
ecoli & 1142.59 & 1126.72 & 1123.46 & 1144.08 & 1140.67 & 1188.07 & 1126.08 & - \\
\bottomrule
\end{tabular}
}
\end{table}
\begin{table}[H]
\caption{CPU time on dataset collection 1 with 20.0\% constraints}
\label{tab:time_COL1_20}
\centering
\resizebox{0.9\textwidth}{!}{%
\begin{tabular}{c|cccccccc}
\toprule
Constraints Level & \multicolumn{8}{c}{0.2} \\
Method & COPKM\_P & KSKM & KSKM\_E & KSKM\_repo & KSKM\_E\_repo & BLPKMCC & PCCC & COPKM \\
Dataset &  &  &  &  &  &  &  &  \\
\midrule
appendicitis & 0.00 & 0.00 & 0.00 & 0.00 & 0.00 & 0.01 & 0.01 & - \\
breast\_cancer & 0.00 & 0.00 & 0.00 & 0.00 & 0.00 & 0.17 & 0.01 & 0.00 \\
bupa & 0.00 & 0.00 & 0.00 & 0.00 & 0.00 & 0.06 & 0.01 & 0.00 \\
circles & 0.00 & 0.00 & 0.00 & 0.00 & 0.00 & 0.05 & 0.00 & 0.00 \\
ecoli & 0.00 & 0.03 & 0.10 & 0.02 & 0.05 & 0.68 & 0.14 & - \\
glass & 0.00 & 0.03 & 0.11 & 0.02 & 0.06 & 0.21 & 0.17 & - \\
haberman & 0.00 & 0.00 & 0.00 & 0.00 & 0.00 & 0.05 & 0.01 & 0.00 \\
hayes-roth & 0.00 & 0.01 & 0.01 & 0.01 & 0.01 & 0.04 & 0.03 & - \\
heart & 0.00 & 0.00 & 0.00 & 0.00 & 0.00 & 0.04 & 0.01 & 0.00 \\
ionosphere & 0.00 & 0.00 & 0.00 & 0.00 & 0.00 & 0.07 & 0.01 & 0.00 \\
led7digit & 0.00 & 0.02 & 0.16 & 0.03 & 0.14 & 1.48 & 0.93 & - \\
monk-2 & 0.00 & 0.00 & 0.00 & 0.00 & 0.00 & 0.11 & 0.01 & 0.00 \\
moons & 0.00 & 0.00 & 0.00 & 0.00 & 0.00 & 0.05 & 0.01 & 0.00 \\
movement\_libras & 0.01 & 0.09 & 3.22 & 0.05 & 1.91 & 5.40 & 8.52 & - \\
newthyroid & 0.00 & 0.00 & 0.00 & 0.00 & 0.00 & 0.06 & 0.03 & 0.00 \\
saheart & 0.00 & 0.00 & 0.00 & 0.00 & 0.00 & 0.11 & 0.01 & 0.00 \\
sonar & 0.00 & 0.00 & 0.00 & 0.00 & 0.00 & 0.03 & 0.01 & 0.00 \\
soybean & 0.00 & 0.00 & 0.00 & 0.01 & 0.01 & 0.01 & 0.04 & 0.00 \\
spectfheart & 0.00 & 0.00 & 0.00 & 0.00 & 0.00 & 0.04 & 0.01 & 0.00 \\
spiral & 0.00 & 0.00 & 0.00 & 0.00 & 0.00 & 0.05 & 0.01 & 0.00 \\
tae & 0.00 & 0.01 & 0.01 & 0.00 & 0.00 & 0.04 & 0.03 & - \\
vehicle & 0.00 & 0.00 & 0.00 & 0.00 & 0.00 & 0.71 & 0.02 & 0.00 \\
zoo & 0.00 & 0.01 & 0.06 & 0.00 & 0.03 & 0.08 & 0.13 & 0.00 \\
\bottomrule
\end{tabular}
}
\end{table}
\begin{table}[H]
\caption{ARI on dataset collection 1 with 20.0\% constraints}
\label{tab:ARI_COL1_20}
\centering
\resizebox{0.9\textwidth}{!}{%
\begin{tabular}{c|cccccccc}
\toprule
Constraints Level & \multicolumn{8}{c}{0.2} \\
Method & COPKM\_P & KSKM & KSKM\_E & KSKM\_repo & KSKM\_E\_repo & BLPKMCC & PCCC & COPKM \\
Dataset &  &  &  &  &  &  &  &  \\
\midrule
appendicitis & 1.00 & 1.00 & 1.00 & 1.00 & 1.00 & 1.00 & 1.00 & - \\
breast\_cancer & 1.00 & 1.00 & 1.00 & 1.00 & 1.00 & 1.00 & 1.00 & 1.00 \\
bupa & 1.00 & 1.00 & 1.00 & 1.00 & 1.00 & 1.00 & 1.00 & 1.00 \\
circles & 1.00 & 1.00 & 1.00 & 1.00 & 1.00 & 1.00 & 1.00 & 1.00 \\
ecoli & 0.98 & 0.97 & 0.98 & 0.98 & 0.98 & 0.97 & 0.98 & - \\
glass & 0.93 & 0.93 & 0.93 & 0.93 & 0.93 & 0.94 & 0.94 & - \\
haberman & 1.00 & 1.00 & 1.00 & 1.00 & 1.00 & 1.00 & 1.00 & 1.00 \\
hayes-roth & 0.92 & 0.92 & 0.92 & 0.92 & 0.92 & 0.93 & 0.95 & - \\
heart & 1.00 & 1.00 & 1.00 & 1.00 & 1.00 & 1.00 & 1.00 & 1.00 \\
ionosphere & 1.00 & 1.00 & 1.00 & 1.00 & 1.00 & 1.00 & 1.00 & 1.00 \\
led7digit & 0.98 & 1.00 & 1.00 & 0.99 & 0.99 & 0.99 & 0.99 & - \\
monk-2 & 1.00 & 1.00 & 1.00 & 1.00 & 1.00 & 1.00 & 1.00 & 1.00 \\
moons & 1.00 & 1.00 & 1.00 & 1.00 & 1.00 & 1.00 & 1.00 & 1.00 \\
movement\_libras & 0.46 & 0.46 & 0.52 & 0.43 & 0.51 & 0.46 & 0.54 & - \\
newthyroid & 0.98 & 0.98 & 0.98 & 0.98 & 0.98 & 0.98 & 1.00 & 0.98 \\
saheart & 1.00 & 1.00 & 1.00 & 1.00 & 1.00 & 1.00 & 1.00 & 1.00 \\
sonar & 1.00 & 1.00 & 1.00 & 1.00 & 1.00 & 1.00 & 1.00 & 1.00 \\
soybean & 1.00 & 1.00 & 1.00 & 1.00 & 1.00 & 0.86 & 1.00 & 1.00 \\
spectfheart & 1.00 & 1.00 & 1.00 & 1.00 & 1.00 & 1.00 & 1.00 & 1.00 \\
spiral & 1.00 & 1.00 & 1.00 & 1.00 & 1.00 & 1.00 & 1.00 & 1.00 \\
tae & 0.96 & 0.96 & 0.96 & 0.96 & 0.96 & 0.96 & 0.96 & - \\
vehicle & 1.00 & 1.00 & 1.00 & 1.00 & 1.00 & 1.00 & 1.00 & 1.00 \\
zoo & 0.83 & 0.85 & 0.86 & 0.84 & 0.83 & 0.83 & 0.93 & 0.83 \\
\bottomrule
\end{tabular}
}
\end{table}
\begin{table}[H]
\caption{Algorithm success rate on dataset collection 1 with 20.0\% constraints}
\label{tab:rate_success_COL1_20}
\centering
\resizebox{0.9\textwidth}{!}{%
\begin{tabular}{c|cccccccc}
\toprule
Constraints Level & \multicolumn{8}{c}{0.2} \\
Method & COPKM\_P & KSKM & KSKM\_E & KSKM\_repo & KSKM\_E\_repo & BLPKMCC & PCCC & COPKM \\
Dataset &  &  &  &  &  &  &  &  \\
\midrule
appendicitis & 1.00 & 1.00 & 1.00 & 1.00 & 1.00 & 1.00 & 1.00 & 0.00 \\
breast\_cancer & 1.00 & 1.00 & 1.00 & 1.00 & 1.00 & 1.00 & 1.00 & 1.00 \\
bupa & 1.00 & 1.00 & 1.00 & 1.00 & 1.00 & 1.00 & 1.00 & 1.00 \\
circles & 1.00 & 1.00 & 1.00 & 1.00 & 1.00 & 1.00 & 1.00 & 1.00 \\
ecoli & 1.00 & 1.00 & 1.00 & 1.00 & 1.00 & 1.00 & 1.00 & 0.00 \\
glass & 1.00 & 1.00 & 1.00 & 1.00 & 1.00 & 1.00 & 1.00 & 0.00 \\
haberman & 1.00 & 1.00 & 1.00 & 1.00 & 1.00 & 1.00 & 1.00 & 1.00 \\
hayes-roth & 1.00 & 1.00 & 1.00 & 1.00 & 1.00 & 1.00 & 1.00 & 0.00 \\
heart & 1.00 & 1.00 & 1.00 & 1.00 & 1.00 & 1.00 & 1.00 & 1.00 \\
ionosphere & 1.00 & 1.00 & 1.00 & 1.00 & 1.00 & 1.00 & 1.00 & 1.00 \\
led7digit & 1.00 & 1.00 & 1.00 & 1.00 & 1.00 & 1.00 & 1.00 & 0.00 \\
monk-2 & 1.00 & 1.00 & 1.00 & 1.00 & 1.00 & 1.00 & 1.00 & 1.00 \\
moons & 1.00 & 1.00 & 1.00 & 1.00 & 1.00 & 1.00 & 1.00 & 1.00 \\
movement\_libras & 1.00 & 1.00 & 1.00 & 1.00 & 1.00 & 1.00 & 1.00 & 0.00 \\
newthyroid & 1.00 & 1.00 & 1.00 & 1.00 & 1.00 & 1.00 & 1.00 & 0.50 \\
saheart & 1.00 & 1.00 & 1.00 & 1.00 & 1.00 & 1.00 & 1.00 & 1.00 \\
sonar & 1.00 & 1.00 & 1.00 & 1.00 & 1.00 & 1.00 & 1.00 & 1.00 \\
soybean & 1.00 & 1.00 & 1.00 & 1.00 & 1.00 & 1.00 & 1.00 & 0.50 \\
spectfheart & 1.00 & 1.00 & 1.00 & 1.00 & 1.00 & 1.00 & 1.00 & 1.00 \\
spiral & 1.00 & 1.00 & 1.00 & 1.00 & 1.00 & 1.00 & 1.00 & 0.70 \\
tae & 1.00 & 1.00 & 1.00 & 1.00 & 1.00 & 1.00 & 1.00 & 0.00 \\
vehicle & 1.00 & 1.00 & 1.00 & 1.00 & 1.00 & 1.00 & 1.00 & 1.00 \\
zoo & 1.00 & 1.00 & 1.00 & 1.00 & 1.00 & 1.00 & 1.00 & 1.00 \\
\bottomrule
\end{tabular}
}
\end{table}
\begin{table}[H]
\caption{Clustering inertia on dataset collection 2 with 5.0\% constraints}
\label{tab:obj_COL2_5}
\centering
\resizebox{0.9\textwidth}{!}{%
\begin{tabular}{c|cccccccc}
\toprule
Constraints Level & \multicolumn{8}{c}{0.05} \\
Method & COPKM\_P & KSKM & KSKM\_E & KSKM\_repo & KSKM\_E\_repo & BLPKMCC & PCCC & COPKM \\
Dataset &  &  &  &  &  &  &  &  \\
\midrule
synthetic\_n300\_d2\_k10\_s10 & 5.72 & 4.94 & 4.94 & 4.94 & 4.94 & 11.69 & 4.94 & 5.84 \\
synthetic\_n300\_d2\_k10\_s20 & 21.28 & 18.76 & 18.16 & 18.16 & 18.16 & 24.40 & 19.13 & 21.24 \\
synthetic\_n300\_d2\_k10\_s30 & 39.19 & 34.95 & 34.93 & 35.06 & 34.96 & 37.53 & 35.04 & 39.14 \\
synthetic\_n300\_d2\_k10\_s40 & 55.49 & 50.89 & 50.68 & 51.11 & 51.08 & 53.61 & 51.58 & 55.26 \\
synthetic\_n300\_d2\_k10\_s50 & 65.07 & 61.65 & 62.05 & 62.11 & 61.86 & 64.11 & 62.95 & 66.26 \\
synthetic\_n300\_d2\_k20\_s10 & 7.08 & 5.85 & 5.99 & 5.88 & 5.86 & 9.60 & 6.15 & 7.09 \\
synthetic\_n300\_d2\_k20\_s20 & 17.26 & 16.01 & 16.05 & 16.39 & 16.21 & 18.89 & 16.53 & 17.40 \\
synthetic\_n300\_d2\_k20\_s30 & 27.87 & 25.61 & 25.49 & 25.82 & 25.85 & 29.74 & 26.54 & 27.86 \\
synthetic\_n300\_d2\_k20\_s40 & 35.52 & 32.61 & 32.89 & 33.37 & 33.34 & 36.20 & 33.89 & 35.81 \\
synthetic\_n300\_d2\_k20\_s50 & 38.60 & 36.72 & 36.94 & 37.55 & 37.58 & 40.76 & 38.21 & 38.75 \\
synthetic\_n300\_d2\_k50\_s10 & 3.43 & 3.16 & 3.13 & 3.27 & 3.22 & 5.06 & 3.22 & 3.45 \\
synthetic\_n300\_d2\_k50\_s20 & 7.76 & 7.13 & 7.08 & 7.44 & 7.36 & 8.93 & 7.58 & 7.74 \\
synthetic\_n300\_d2\_k50\_s30 & 9.59 & 8.75 & 8.64 & 9.10 & 9.18 & 10.90 & 9.26 & 9.56 \\
synthetic\_n300\_d2\_k50\_s40 & 10.75 & 9.80 & 9.89 & 10.32 & 10.33 & 12.47 & 10.56 & 10.80 \\
synthetic\_n300\_d2\_k50\_s50 & 11.37 & 10.54 & 10.56 & 11.04 & 11.18 & 13.65 & 11.29 & 11.39 \\
\bottomrule
\end{tabular}
}
\end{table}
\begin{table}[H]
\caption{CPU time on dataset collection 2 with 5.0\% constraints}
\label{tab:time_COL2_5}
\centering
\resizebox{0.9\textwidth}{!}{%
\begin{tabular}{c|cccccccc}
\toprule
Constraints Level & \multicolumn{8}{c}{0.05} \\
Method & COPKM\_P & KSKM & KSKM\_E & KSKM\_repo & KSKM\_E\_repo & BLPKMCC & PCCC & COPKM \\
Dataset &  &  &  &  &  &  &  &  \\
\midrule
synthetic\_n300\_d2\_k10\_s10 & 0.14 & 0.02 & 0.05 & 0.06 & 0.11 & 0.16 & 0.24 & 0.23 \\
synthetic\_n300\_d2\_k10\_s20 & 0.01 & 0.04 & 0.16 & 0.07 & 0.13 & 0.21 & 0.45 & 0.01 \\
synthetic\_n300\_d2\_k10\_s30 & 0.01 & 0.07 & 0.31 & 0.07 & 0.30 & 0.35 & 0.76 & 0.01 \\
synthetic\_n300\_d2\_k10\_s40 & 0.01 & 0.08 & 0.43 & 0.09 & 0.26 & 0.29 & 0.80 & 0.01 \\
synthetic\_n300\_d2\_k10\_s50 & 0.01 & 0.15 & 0.58 & 0.10 & 0.44 & 0.31 & 0.96 & 0.01 \\
synthetic\_n300\_d2\_k20\_s10 & 0.01 & 0.06 & 0.45 & 0.04 & 0.26 & 0.53 & 0.80 & 0.01 \\
synthetic\_n300\_d2\_k20\_s20 & 0.01 & 0.09 & 0.84 & 0.06 & 0.43 & 0.50 & 1.27 & 0.02 \\
synthetic\_n300\_d2\_k20\_s30 & 0.01 & 0.12 & 0.94 & 0.11 & 0.60 & 0.52 & 1.78 & 0.02 \\
synthetic\_n300\_d2\_k20\_s40 & 0.01 & 0.18 & 1.07 & 0.08 & 0.47 & 0.70 & 1.30 & 0.01 \\
synthetic\_n300\_d2\_k20\_s50 & 0.01 & 0.19 & 0.81 & 0.07 & 0.34 & 0.62 & 1.46 & 0.02 \\
synthetic\_n300\_d2\_k50\_s10 & 0.01 & 0.14 & 1.85 & 0.04 & 0.78 & 0.94 & 2.10 & 0.02 \\
synthetic\_n300\_d2\_k50\_s20 & 0.01 & 0.14 & 1.89 & 0.05 & 1.14 & 1.04 & 1.95 & 0.02 \\
synthetic\_n300\_d2\_k50\_s30 & 0.01 & 0.13 & 2.19 & 0.05 & 0.85 & 1.07 & 2.03 & 0.02 \\
synthetic\_n300\_d2\_k50\_s40 & 0.01 & 0.15 & 1.79 & 0.04 & 0.78 & 1.04 & 1.83 & 0.02 \\
synthetic\_n300\_d2\_k50\_s50 & 0.01 & 0.15 & 1.95 & 0.04 & 0.63 & 0.99 & 2.02 & 0.02 \\
\bottomrule
\end{tabular}
}
\end{table}
\begin{table}[H]
\caption{ARI on dataset collection 2 with 5.0\% constraints}
\label{tab:ARI_COL2_5}
\centering
\resizebox{0.9\textwidth}{!}{%
\begin{tabular}{c|cccccccc}
\toprule
Constraints Level & \multicolumn{8}{c}{0.05} \\
Method & COPKM\_P & KSKM & KSKM\_E & KSKM\_repo & KSKM\_E\_repo & BLPKMCC & PCCC & COPKM \\
Dataset &  &  &  &  &  &  &  &  \\
\midrule
synthetic\_n300\_d2\_k10\_s10 & 0.98 & 0.99 & 0.99 & 0.99 & 0.99 & 0.86 & 0.99 & 0.98 \\
synthetic\_n300\_d2\_k10\_s20 & 0.82 & 0.86 & 0.88 & 0.88 & 0.88 & 0.78 & 0.87 & 0.83 \\
synthetic\_n300\_d2\_k10\_s30 & 0.64 & 0.72 & 0.72 & 0.71 & 0.71 & 0.67 & 0.72 & 0.66 \\
synthetic\_n300\_d2\_k10\_s40 & 0.49 & 0.52 & 0.51 & 0.51 & 0.52 & 0.49 & 0.51 & 0.50 \\
synthetic\_n300\_d2\_k10\_s50 & 0.37 & 0.36 & 0.37 & 0.37 & 0.36 & 0.38 & 0.37 & 0.37 \\
synthetic\_n300\_d2\_k20\_s10 & 0.82 & 0.90 & 0.89 & 0.90 & 0.90 & 0.72 & 0.88 & 0.82 \\
synthetic\_n300\_d2\_k20\_s20 & 0.52 & 0.54 & 0.54 & 0.54 & 0.54 & 0.53 & 0.54 & 0.51 \\
synthetic\_n300\_d2\_k20\_s30 & 0.36 & 0.36 & 0.37 & 0.36 & 0.35 & 0.36 & 0.36 & 0.36 \\
synthetic\_n300\_d2\_k20\_s40 & 0.23 & 0.25 & 0.24 & 0.23 & 0.24 & 0.24 & 0.24 & 0.23 \\
synthetic\_n300\_d2\_k20\_s50 & 0.18 & 0.17 & 0.17 & 0.17 & 0.17 & 0.17 & 0.17 & 0.18 \\
synthetic\_n300\_d2\_k50\_s10 & 0.73 & 0.77 & 0.77 & 0.74 & 0.75 & 0.65 & 0.74 & 0.72 \\
synthetic\_n300\_d2\_k50\_s20 & 0.37 & 0.37 & 0.38 & 0.38 & 0.37 & 0.37 & 0.37 & 0.37 \\
synthetic\_n300\_d2\_k50\_s30 & 0.21 & 0.22 & 0.22 & 0.22 & 0.22 & 0.22 & 0.22 & 0.21 \\
synthetic\_n300\_d2\_k50\_s40 & 0.13 & 0.13 & 0.13 & 0.13 & 0.13 & 0.14 & 0.13 & 0.13 \\
synthetic\_n300\_d2\_k50\_s50 & 0.09 & 0.08 & 0.09 & 0.09 & 0.09 & 0.10 & 0.09 & 0.09 \\
\bottomrule
\end{tabular}
}
\end{table}
\begin{table}[H]
\caption{Algorithm success rate on dataset collection 2 with 5.0\% constraints}
\label{tab:rate_success_COL2_5}
\centering
\resizebox{0.9\textwidth}{!}{%
\begin{tabular}{c|cccccccc}
\toprule
Constraints Level & \multicolumn{8}{c}{0.05} \\
Method & COPKM\_P & KSKM & KSKM\_E & KSKM\_repo & KSKM\_E\_repo & BLPKMCC & PCCC & COPKM \\
Dataset &  &  &  &  &  &  &  &  \\
\midrule
synthetic\_n300\_d2\_k10\_s10 & 1.00 & 1.00 & 1.00 & 1.00 & 1.00 & 1.00 & 1.00 & 1.00 \\
synthetic\_n300\_d2\_k10\_s20 & 1.00 & 1.00 & 1.00 & 1.00 & 1.00 & 1.00 & 1.00 & 1.00 \\
synthetic\_n300\_d2\_k10\_s30 & 1.00 & 1.00 & 1.00 & 1.00 & 1.00 & 1.00 & 1.00 & 1.00 \\
synthetic\_n300\_d2\_k10\_s40 & 1.00 & 1.00 & 1.00 & 1.00 & 1.00 & 1.00 & 1.00 & 1.00 \\
synthetic\_n300\_d2\_k10\_s50 & 1.00 & 1.00 & 1.00 & 1.00 & 1.00 & 1.00 & 1.00 & 1.00 \\
synthetic\_n300\_d2\_k20\_s10 & 1.00 & 1.00 & 1.00 & 1.00 & 1.00 & 1.00 & 1.00 & 1.00 \\
synthetic\_n300\_d2\_k20\_s20 & 1.00 & 1.00 & 1.00 & 1.00 & 1.00 & 1.00 & 1.00 & 1.00 \\
synthetic\_n300\_d2\_k20\_s30 & 1.00 & 1.00 & 1.00 & 1.00 & 1.00 & 1.00 & 1.00 & 1.00 \\
synthetic\_n300\_d2\_k20\_s40 & 1.00 & 1.00 & 1.00 & 1.00 & 1.00 & 1.00 & 1.00 & 1.00 \\
synthetic\_n300\_d2\_k20\_s50 & 1.00 & 1.00 & 1.00 & 1.00 & 1.00 & 1.00 & 1.00 & 1.00 \\
synthetic\_n300\_d2\_k50\_s10 & 1.00 & 1.00 & 1.00 & 1.00 & 1.00 & 1.00 & 1.00 & 1.00 \\
synthetic\_n300\_d2\_k50\_s20 & 1.00 & 1.00 & 1.00 & 1.00 & 1.00 & 1.00 & 1.00 & 1.00 \\
synthetic\_n300\_d2\_k50\_s30 & 1.00 & 1.00 & 1.00 & 1.00 & 1.00 & 1.00 & 1.00 & 1.00 \\
synthetic\_n300\_d2\_k50\_s40 & 1.00 & 1.00 & 1.00 & 1.00 & 1.00 & 1.00 & 1.00 & 1.00 \\
synthetic\_n300\_d2\_k50\_s50 & 1.00 & 1.00 & 1.00 & 1.00 & 1.00 & 1.00 & 1.00 & 1.00 \\
\bottomrule
\end{tabular}
}
\end{table}
\begin{table}[H]
\caption{Clustering inertia on dataset collection 2 with 10.0\% constraints}
\label{tab:obj_COL2_10}
\centering
\resizebox{0.9\textwidth}{!}{%
\begin{tabular}{c|cccccccc}
\toprule
Constraints Level & \multicolumn{8}{c}{0.1} \\
Method & COPKM\_P & KSKM & KSKM\_E & KSKM\_repo & KSKM\_E\_repo & BLPKMCC & PCCC & COPKM \\
Dataset &  &  &  &  &  &  &  &  \\
\midrule
synthetic\_n300\_d2\_k10\_s10 & 6.48 & 4.94 & 4.94 & 4.94 & 4.94 & 15.22 & 6.19 & 8.34 \\
synthetic\_n300\_d2\_k10\_s20 & 21.79 & 18.14 & 18.14 & 18.17 & 18.16 & 24.33 & 19.34 & 25.21 \\
synthetic\_n300\_d2\_k10\_s30 & 44.71 & 36.22 & 36.22 & 36.23 & 36.23 & 42.10 & 36.26 & 40.07 \\
synthetic\_n300\_d2\_k10\_s40 & 62.85 & 55.33 & 55.46 & 55.74 & 55.63 & 58.59 & 55.66 & 61.81 \\
synthetic\_n300\_d2\_k10\_s50 & 77.31 & 71.45 & 71.17 & 71.78 & 71.41 & 72.96 & 71.64 & 79.20 \\
synthetic\_n300\_d2\_k20\_s10 & 7.49 & 6.02 & 5.99 & 6.04 & 5.99 & 10.76 & 6.01 & 7.42 \\
synthetic\_n300\_d2\_k20\_s20 & 19.15 & 17.52 & 17.53 & 17.88 & 17.57 & 20.60 & 18.06 & 19.57 \\
synthetic\_n300\_d2\_k20\_s30 & 31.36 & 29.24 & 29.31 & 29.86 & 29.56 & 32.92 & 29.92 & 31.59 \\
synthetic\_n300\_d2\_k20\_s40 & 40.81 & 37.34 & 37.15 & 38.41 & 37.57 & 41.33 & 38.41 & 40.75 \\
synthetic\_n300\_d2\_k20\_s50 & 47.24 & 42.38 & 42.45 & 43.28 & 42.84 & 47.41 & 44.48 & 46.49 \\
synthetic\_n300\_d2\_k50\_s10 & 3.47 & 3.23 & 3.30 & 3.31 & 3.30 & 5.06 & 3.31 & 3.49 \\
synthetic\_n300\_d2\_k50\_s20 & 8.06 & 7.45 & 7.40 & 7.66 & 7.73 & 9.32 & 7.85 & 8.16 \\
synthetic\_n300\_d2\_k50\_s30 & 10.40 & 9.42 & 9.44 & 9.89 & 9.96 & 11.67 & 9.89 & 10.40 \\
synthetic\_n300\_d2\_k50\_s40 & 11.82 & 10.43 & 10.46 & 11.08 & 11.14 & 13.57 & 11.18 & 11.73 \\
synthetic\_n300\_d2\_k50\_s50 & 12.98 & 12.01 & 11.99 & 12.79 & 12.62 & 14.97 & 12.73 & 13.03 \\
\bottomrule
\end{tabular}
}
\end{table}
\begin{table}[H]
\caption{CPU time on dataset collection 2 with 10.0\% constraints}
\label{tab:time_COL2_10}
\centering
\resizebox{0.9\textwidth}{!}{%
\begin{tabular}{c|cccccccc}
\toprule
Constraints Level & \multicolumn{8}{c}{0.1} \\
Method & COPKM\_P & KSKM & KSKM\_E & KSKM\_repo & KSKM\_E\_repo & BLPKMCC & PCCC & COPKM \\
Dataset &  &  &  &  &  &  &  &  \\
\midrule
synthetic\_n300\_d2\_k10\_s10 & 0.01 & 0.02 & 0.05 & 0.04 & 0.07 & 0.30 & 0.42 & 0.01 \\
synthetic\_n300\_d2\_k10\_s20 & 0.01 & 0.04 & 0.32 & 0.05 & 0.21 & 0.41 & 0.69 & 0.01 \\
synthetic\_n300\_d2\_k10\_s30 & 0.01 & 0.06 & 0.48 & 0.06 & 0.26 & 0.59 & 1.16 & 0.01 \\
synthetic\_n300\_d2\_k10\_s40 & 0.01 & 0.07 & 0.61 & 0.06 & 0.30 & 0.55 & 1.26 & 0.01 \\
synthetic\_n300\_d2\_k10\_s50 & 0.01 & 0.10 & 0.79 & 0.06 & 0.35 & 0.59 & 1.49 & 0.01 \\
synthetic\_n300\_d2\_k20\_s10 & 0.01 & 0.06 & 0.86 & 0.04 & 0.93 & 0.95 & 2.29 & 0.01 \\
synthetic\_n300\_d2\_k20\_s20 & 0.01 & 0.12 & 1.70 & 0.04 & 0.75 & 0.89 & 1.96 & 0.02 \\
synthetic\_n300\_d2\_k20\_s30 & 0.01 & 0.15 & 1.87 & 0.06 & 1.07 & 1.00 & 2.44 & 0.01 \\
synthetic\_n300\_d2\_k20\_s40 & 0.01 & 0.18 & 1.87 & 0.06 & 1.11 & 1.06 & 2.37 & 0.01 \\
synthetic\_n300\_d2\_k20\_s50 & 0.01 & 0.14 & 1.73 & 0.07 & 0.96 & 1.14 & 2.52 & 0.01 \\
synthetic\_n300\_d2\_k50\_s10 & 0.01 & 0.13 & 4.08 & 0.04 & 2.47 & 2.08 & 4.90 & 0.02 \\
synthetic\_n300\_d2\_k50\_s20 & 0.01 & 0.18 & 5.34 & 0.07 & 3.19 & 2.22 & 6.43 & 0.02 \\
synthetic\_n300\_d2\_k50\_s30 & 0.01 & 0.17 & 5.06 & 0.06 & 2.23 & 2.02 & 6.43 & 0.02 \\
synthetic\_n300\_d2\_k50\_s40 & 0.01 & 0.18 & 5.19 & 0.07 & 2.65 & 2.31 & 4.33 & 0.02 \\
synthetic\_n300\_d2\_k50\_s50 & 0.01 & 0.19 & 4.51 & 0.04 & 1.82 & 2.24 & 4.25 & 0.02 \\
\bottomrule
\end{tabular}
}
\end{table}
\begin{table}[H]
\caption{ARI on dataset collection 2 with 10.0\% constraints}
\label{tab:ARI_COL2_10}
\centering
\resizebox{0.9\textwidth}{!}{%
\begin{tabular}{c|cccccccc}
\toprule
Constraints Level & \multicolumn{8}{c}{0.1} \\
Method & COPKM\_P & KSKM & KSKM\_E & KSKM\_repo & KSKM\_E\_repo & BLPKMCC & PCCC & COPKM \\
Dataset &  &  &  &  &  &  &  &  \\
\midrule
synthetic\_n300\_d2\_k10\_s10 & 0.97 & 0.99 & 0.99 & 0.99 & 0.99 & 0.87 & 0.96 & 0.96 \\
synthetic\_n300\_d2\_k10\_s20 & 0.85 & 0.88 & 0.88 & 0.89 & 0.88 & 0.82 & 0.87 & 0.83 \\
synthetic\_n300\_d2\_k10\_s30 & 0.68 & 0.77 & 0.78 & 0.77 & 0.77 & 0.72 & 0.77 & 0.74 \\
synthetic\_n300\_d2\_k10\_s40 & 0.53 & 0.61 & 0.60 & 0.60 & 0.60 & 0.54 & 0.59 & 0.57 \\
synthetic\_n300\_d2\_k10\_s50 & 0.41 & 0.43 & 0.44 & 0.43 & 0.44 & 0.43 & 0.44 & 0.43 \\
synthetic\_n300\_d2\_k20\_s10 & 0.81 & 0.93 & 0.94 & 0.93 & 0.94 & 0.72 & 0.94 & 0.81 \\
synthetic\_n300\_d2\_k20\_s20 & 0.53 & 0.57 & 0.57 & 0.57 & 0.57 & 0.53 & 0.56 & 0.52 \\
synthetic\_n300\_d2\_k20\_s30 & 0.37 & 0.39 & 0.40 & 0.39 & 0.40 & 0.38 & 0.39 & 0.38 \\
synthetic\_n300\_d2\_k20\_s40 & 0.25 & 0.26 & 0.26 & 0.26 & 0.26 & 0.26 & 0.26 & 0.25 \\
synthetic\_n300\_d2\_k20\_s50 & 0.19 & 0.18 & 0.18 & 0.19 & 0.18 & 0.19 & 0.19 & 0.19 \\
synthetic\_n300\_d2\_k50\_s10 & 0.73 & 0.77 & 0.75 & 0.75 & 0.75 & 0.65 & 0.75 & 0.73 \\
synthetic\_n300\_d2\_k50\_s20 & 0.37 & 0.40 & 0.39 & 0.39 & 0.38 & 0.37 & 0.38 & 0.37 \\
synthetic\_n300\_d2\_k50\_s30 & 0.21 & 0.22 & 0.22 & 0.21 & 0.21 & 0.22 & 0.21 & 0.21 \\
synthetic\_n300\_d2\_k50\_s40 & 0.13 & 0.13 & 0.13 & 0.13 & 0.13 & 0.14 & 0.13 & 0.13 \\
synthetic\_n300\_d2\_k50\_s50 & 0.09 & 0.09 & 0.09 & 0.09 & 0.09 & 0.10 & 0.10 & 0.09 \\
\bottomrule
\end{tabular}
}
\end{table}
\begin{table}[H]
\caption{Algorithm success rate on dataset collection 2 with 10.0\% constraints}
\label{tab:rate_success_COL2_10}
\centering
\resizebox{0.9\textwidth}{!}{%
\begin{tabular}{c|cccccccc}
\toprule
Constraints Level & \multicolumn{8}{c}{0.1} \\
Method & COPKM\_P & KSKM & KSKM\_E & KSKM\_repo & KSKM\_E\_repo & BLPKMCC & PCCC & COPKM \\
Dataset &  &  &  &  &  &  &  &  \\
\midrule
synthetic\_n300\_d2\_k10\_s10 & 1.00 & 1.00 & 1.00 & 1.00 & 1.00 & 1.00 & 1.00 & 1.00 \\
synthetic\_n300\_d2\_k10\_s20 & 1.00 & 1.00 & 1.00 & 1.00 & 1.00 & 1.00 & 1.00 & 1.00 \\
synthetic\_n300\_d2\_k10\_s30 & 1.00 & 1.00 & 1.00 & 1.00 & 1.00 & 1.00 & 1.00 & 1.00 \\
synthetic\_n300\_d2\_k10\_s40 & 1.00 & 1.00 & 1.00 & 1.00 & 1.00 & 1.00 & 1.00 & 1.00 \\
synthetic\_n300\_d2\_k10\_s50 & 1.00 & 1.00 & 1.00 & 1.00 & 1.00 & 1.00 & 1.00 & 1.00 \\
synthetic\_n300\_d2\_k20\_s10 & 1.00 & 1.00 & 1.00 & 1.00 & 1.00 & 1.00 & 1.00 & 1.00 \\
synthetic\_n300\_d2\_k20\_s20 & 1.00 & 1.00 & 1.00 & 1.00 & 1.00 & 1.00 & 1.00 & 1.00 \\
synthetic\_n300\_d2\_k20\_s30 & 1.00 & 1.00 & 1.00 & 1.00 & 1.00 & 1.00 & 1.00 & 1.00 \\
synthetic\_n300\_d2\_k20\_s40 & 1.00 & 1.00 & 1.00 & 1.00 & 1.00 & 1.00 & 1.00 & 1.00 \\
synthetic\_n300\_d2\_k20\_s50 & 1.00 & 1.00 & 1.00 & 1.00 & 1.00 & 1.00 & 1.00 & 1.00 \\
synthetic\_n300\_d2\_k50\_s10 & 1.00 & 1.00 & 1.00 & 1.00 & 1.00 & 1.00 & 1.00 & 1.00 \\
synthetic\_n300\_d2\_k50\_s20 & 1.00 & 1.00 & 1.00 & 1.00 & 1.00 & 1.00 & 1.00 & 1.00 \\
synthetic\_n300\_d2\_k50\_s30 & 1.00 & 1.00 & 1.00 & 1.00 & 1.00 & 1.00 & 1.00 & 1.00 \\
synthetic\_n300\_d2\_k50\_s40 & 1.00 & 1.00 & 1.00 & 1.00 & 1.00 & 1.00 & 1.00 & 1.00 \\
synthetic\_n300\_d2\_k50\_s50 & 1.00 & 1.00 & 1.00 & 1.00 & 1.00 & 1.00 & 1.00 & 1.00 \\
\bottomrule
\end{tabular}
}
\end{table}
\begin{table}[H]
\caption{Clustering inertia on dataset collection 2 with 15.0\% constraints}
\label{tab:obj_COL2_15}
\centering
\resizebox{0.9\textwidth}{!}{%
\begin{tabular}{c|cccccccc}
\toprule
Constraints Level & \multicolumn{8}{c}{0.15} \\
Method & COPKM\_P & KSKM & KSKM\_E & KSKM\_repo & KSKM\_E\_repo & BLPKMCC & PCCC & COPKM \\
Dataset &  &  &  &  &  &  &  &  \\
\midrule
synthetic\_n300\_d2\_k10\_s10 & 4.94 & 4.94 & 4.94 & 4.94 & 4.94 & 11.78 & 4.94 & 4.94 \\
synthetic\_n300\_d2\_k10\_s20 & 26.52 & 18.59 & 18.59 & 18.59 & 18.59 & 24.82 & 18.59 & 26.25 \\
synthetic\_n300\_d2\_k10\_s30 & 43.30 & 38.74 & 38.75 & 38.74 & 38.75 & 47.34 & 38.81 & 47.11 \\
synthetic\_n300\_d2\_k10\_s40 & 70.70 & 61.07 & 60.85 & 61.36 & 60.90 & 65.11 & 61.41 & 62.79 \\
synthetic\_n300\_d2\_k10\_s50 & 92.41 & 82.89 & 82.63 & 84.17 & 83.17 & 86.92 & 83.62 & 95.56 \\
synthetic\_n300\_d2\_k20\_s10 & 8.98 & 6.16 & 6.16 & 6.07 & 6.02 & 11.43 & 5.98 & 8.67 \\
synthetic\_n300\_d2\_k20\_s20 & 22.34 & 19.15 & 19.00 & 19.56 & 19.31 & 22.37 & 19.43 & 22.93 \\
synthetic\_n300\_d2\_k20\_s30 & 36.67 & 32.20 & 32.18 & 33.02 & 32.43 & 36.95 & 32.81 & 36.57 \\
synthetic\_n300\_d2\_k20\_s40 & 51.52 & 46.14 & 46.14 & 47.68 & 46.93 & 50.63 & 47.03 & 53.05 \\
synthetic\_n300\_d2\_k20\_s50 & 63.22 & 56.31 & 55.23 & 58.55 & 57.27 & 60.66 & 57.47 & 63.27 \\
synthetic\_n300\_d2\_k50\_s10 & 4.02 & 3.29 & 3.32 & 3.50 & 3.50 & 5.56 & 3.36 & 4.02 \\
synthetic\_n300\_d2\_k50\_s20 & 9.42 & 8.34 & 8.36 & 8.65 & 8.55 & 10.41 & 8.73 & 9.43 \\
synthetic\_n300\_d2\_k50\_s30 & 13.30 & 11.59 & 11.63 & 12.18 & 12.04 & 13.93 & 12.19 & 13.26 \\
synthetic\_n300\_d2\_k50\_s40 & 16.23 & 14.39 & 14.30 & 15.08 & 15.14 & 17.08 & 14.84 & 15.98 \\
synthetic\_n300\_d2\_k50\_s50 & 18.97 & 17.37 & 17.24 & 18.49 & 18.40 & 20.30 & 17.96 & 19.24 \\
\bottomrule
\end{tabular}
}
\end{table}
\begin{table}[H]
\caption{CPU time on dataset collection 2 with 15.0\% constraints}
\label{tab:time_COL2_15}
\centering
\resizebox{0.9\textwidth}{!}{%
\begin{tabular}{c|cccccccc}
\toprule
Constraints Level & \multicolumn{8}{c}{0.15} \\
Method & COPKM\_P & KSKM & KSKM\_E & KSKM\_repo & KSKM\_E\_repo & BLPKMCC & PCCC & COPKM \\
Dataset &  &  &  &  &  &  &  &  \\
\midrule
synthetic\_n300\_d2\_k10\_s10 & 0.01 & 0.02 & 0.06 & 0.02 & 0.08 & 0.53 & 0.84 & 0.01 \\
synthetic\_n300\_d2\_k10\_s20 & 0.01 & 0.02 & 0.15 & 0.03 & 0.13 & 0.64 & 1.89 & 0.01 \\
synthetic\_n300\_d2\_k10\_s30 & 0.01 & 0.03 & 0.17 & 0.03 & 0.14 & 0.85 & 1.41 & 0.01 \\
synthetic\_n300\_d2\_k10\_s40 & 0.01 & 0.04 & 0.39 & 0.03 & 0.15 & 0.96 & 2.72 & 0.01 \\
synthetic\_n300\_d2\_k10\_s50 & 0.01 & 0.12 & 0.93 & 0.06 & 0.31 & 0.90 & 2.80 & 0.01 \\
synthetic\_n300\_d2\_k20\_s10 & 0.01 & 0.07 & 1.50 & 0.05 & 1.52 & 1.68 & 3.18 & 0.01 \\
synthetic\_n300\_d2\_k20\_s20 & 0.01 & 0.14 & 3.06 & 0.06 & 1.32 & 2.01 & 4.79 & 0.01 \\
synthetic\_n300\_d2\_k20\_s30 & 0.01 & 0.14 & 3.23 & 0.06 & 1.14 & 1.85 & 5.13 & 0.01 \\
synthetic\_n300\_d2\_k20\_s40 & 0.01 & 0.17 & 4.02 & 0.07 & 1.04 & 1.94 & 5.23 & 0.01 \\
synthetic\_n300\_d2\_k20\_s50 & 0.01 & 0.32 & 3.88 & 0.08 & 1.15 & 2.46 & 4.31 & 0.01 \\
synthetic\_n300\_d2\_k50\_s10 & 0.01 & 0.19 & 9.57 & 0.06 & 4.67 & 4.51 & 13.08 & 0.02 \\
synthetic\_n300\_d2\_k50\_s20 & 0.01 & 0.18 & 9.33 & 0.07 & 6.68 & 4.70 & 10.85 & 0.02 \\
synthetic\_n300\_d2\_k50\_s30 & 0.01 & 0.20 & 9.17 & 0.07 & 5.20 & 4.36 & 9.89 & 0.02 \\
synthetic\_n300\_d2\_k50\_s40 & 0.01 & 0.19 & 8.85 & 0.06 & 4.10 & 5.00 & 9.54 & 0.02 \\
synthetic\_n300\_d2\_k50\_s50 & 0.01 & 0.23 & 10.30 & 0.05 & 2.33 & 4.21 & 9.83 & 0.02 \\
\bottomrule
\end{tabular}
}
\end{table}
\begin{table}[H]
\caption{ARI on dataset collection 2 with 15.0\% constraints}
\label{tab:ARI_COL2_15}
\centering
\resizebox{0.9\textwidth}{!}{%
\begin{tabular}{c|cccccccc}
\toprule
Constraints Level & \multicolumn{8}{c}{0.15} \\
Method & COPKM\_P & KSKM & KSKM\_E & KSKM\_repo & KSKM\_E\_repo & BLPKMCC & PCCC & COPKM \\
Dataset &  &  &  &  &  &  &  &  \\
\midrule
synthetic\_n300\_d2\_k10\_s10 & 0.99 & 0.99 & 0.99 & 0.99 & 0.99 & 0.93 & 0.99 & 0.99 \\
synthetic\_n300\_d2\_k10\_s20 & 0.90 & 0.93 & 0.93 & 0.93 & 0.93 & 0.88 & 0.94 & 0.90 \\
synthetic\_n300\_d2\_k10\_s30 & 0.81 & 0.85 & 0.85 & 0.85 & 0.85 & 0.76 & 0.83 & 0.77 \\
synthetic\_n300\_d2\_k10\_s40 & 0.66 & 0.73 & 0.73 & 0.72 & 0.73 & 0.69 & 0.74 & 0.71 \\
synthetic\_n300\_d2\_k10\_s50 & 0.52 & 0.59 & 0.59 & 0.59 & 0.59 & 0.58 & 0.59 & 0.53 \\
synthetic\_n300\_d2\_k20\_s10 & 0.81 & 0.93 & 0.94 & 0.92 & 0.93 & 0.75 & 0.94 & 0.83 \\
synthetic\_n300\_d2\_k20\_s20 & 0.53 & 0.57 & 0.58 & 0.56 & 0.57 & 0.54 & 0.57 & 0.52 \\
synthetic\_n300\_d2\_k20\_s30 & 0.41 & 0.44 & 0.44 & 0.44 & 0.43 & 0.42 & 0.43 & 0.41 \\
synthetic\_n300\_d2\_k20\_s40 & 0.29 & 0.33 & 0.32 & 0.32 & 0.31 & 0.31 & 0.32 & 0.28 \\
synthetic\_n300\_d2\_k20\_s50 & 0.22 & 0.22 & 0.22 & 0.23 & 0.23 & 0.24 & 0.22 & 0.21 \\
synthetic\_n300\_d2\_k50\_s10 & 0.72 & 0.80 & 0.80 & 0.77 & 0.76 & 0.65 & 0.79 & 0.72 \\
synthetic\_n300\_d2\_k50\_s20 & 0.40 & 0.43 & 0.42 & 0.42 & 0.42 & 0.41 & 0.42 & 0.40 \\
synthetic\_n300\_d2\_k50\_s30 & 0.22 & 0.24 & 0.24 & 0.24 & 0.24 & 0.25 & 0.24 & 0.23 \\
synthetic\_n300\_d2\_k50\_s40 & 0.15 & 0.16 & 0.16 & 0.15 & 0.16 & 0.17 & 0.16 & 0.15 \\
synthetic\_n300\_d2\_k50\_s50 & 0.11 & 0.12 & 0.12 & 0.12 & 0.12 & 0.13 & 0.12 & 0.11 \\
\bottomrule
\end{tabular}
}
\end{table}
\begin{table}[H]
\caption{Algorithm success rate on dataset collection 2 with 15.0\% constraints}
\label{tab:rate_success_COL2_15}
\centering
\resizebox{0.9\textwidth}{!}{%
\begin{tabular}{c|cccccccc}
\toprule
Constraints Level & \multicolumn{8}{c}{0.15} \\
Method & COPKM\_P & KSKM & KSKM\_E & KSKM\_repo & KSKM\_E\_repo & BLPKMCC & PCCC & COPKM \\
Dataset &  &  &  &  &  &  &  &  \\
\midrule
synthetic\_n300\_d2\_k10\_s10 & 1.00 & 1.00 & 1.00 & 1.00 & 1.00 & 1.00 & 1.00 & 1.00 \\
synthetic\_n300\_d2\_k10\_s20 & 1.00 & 1.00 & 1.00 & 1.00 & 1.00 & 1.00 & 1.00 & 1.00 \\
synthetic\_n300\_d2\_k10\_s30 & 1.00 & 1.00 & 1.00 & 1.00 & 1.00 & 1.00 & 1.00 & 1.00 \\
synthetic\_n300\_d2\_k10\_s40 & 1.00 & 1.00 & 1.00 & 1.00 & 1.00 & 1.00 & 1.00 & 1.00 \\
synthetic\_n300\_d2\_k10\_s50 & 1.00 & 1.00 & 1.00 & 1.00 & 1.00 & 1.00 & 1.00 & 1.00 \\
synthetic\_n300\_d2\_k20\_s10 & 1.00 & 1.00 & 1.00 & 1.00 & 1.00 & 1.00 & 1.00 & 1.00 \\
synthetic\_n300\_d2\_k20\_s20 & 1.00 & 1.00 & 1.00 & 1.00 & 1.00 & 1.00 & 1.00 & 1.00 \\
synthetic\_n300\_d2\_k20\_s30 & 1.00 & 1.00 & 1.00 & 1.00 & 1.00 & 1.00 & 1.00 & 1.00 \\
synthetic\_n300\_d2\_k20\_s40 & 1.00 & 1.00 & 1.00 & 1.00 & 1.00 & 1.00 & 1.00 & 1.00 \\
synthetic\_n300\_d2\_k20\_s50 & 1.00 & 1.00 & 1.00 & 1.00 & 1.00 & 1.00 & 1.00 & 1.00 \\
synthetic\_n300\_d2\_k50\_s10 & 1.00 & 1.00 & 1.00 & 1.00 & 1.00 & 1.00 & 1.00 & 1.00 \\
synthetic\_n300\_d2\_k50\_s20 & 1.00 & 1.00 & 1.00 & 1.00 & 1.00 & 1.00 & 1.00 & 1.00 \\
synthetic\_n300\_d2\_k50\_s30 & 1.00 & 1.00 & 1.00 & 1.00 & 1.00 & 1.00 & 1.00 & 1.00 \\
synthetic\_n300\_d2\_k50\_s40 & 1.00 & 1.00 & 1.00 & 1.00 & 1.00 & 1.00 & 1.00 & 1.00 \\
synthetic\_n300\_d2\_k50\_s50 & 1.00 & 1.00 & 1.00 & 1.00 & 1.00 & 1.00 & 1.00 & 1.00 \\
\bottomrule
\end{tabular}
}
\end{table}
\begin{table}[H]
\caption{Clustering inertia on dataset collection 2 with 20.0\% constraints}
\label{tab:obj_COL2_20}
\centering
\resizebox{0.9\textwidth}{!}{%
\begin{tabular}{c|cccccccc}
\toprule
Constraints Level & \multicolumn{8}{c}{0.2} \\
Method & COPKM\_P & KSKM & KSKM\_E & KSKM\_repo & KSKM\_E\_repo & BLPKMCC & PCCC & COPKM \\
Dataset &  &  &  &  &  &  &  &  \\
\midrule
synthetic\_n300\_d2\_k10\_s10 & 4.96 & 4.96 & 4.96 & 4.96 & 4.96 & 4.96 & 4.96 & 4.96 \\
synthetic\_n300\_d2\_k10\_s20 & 19.03 & 19.03 & 19.03 & 19.03 & 19.03 & 19.03 & 19.34 & 19.03 \\
synthetic\_n300\_d2\_k10\_s30 & 40.00 & 39.99 & 39.99 & 39.99 & 40.00 & 39.99 & 40.13 & 39.99 \\
synthetic\_n300\_d2\_k10\_s40 & 77.47 & 66.07 & 66.07 & 66.14 & 66.14 & 66.09 & 66.15 & 67.06 \\
synthetic\_n300\_d2\_k10\_s50 & 105.76 & 93.24 & 93.24 & 93.70 & 93.25 & 93.28 & 93.57 & - \\
synthetic\_n300\_d2\_k20\_s10 & 10.25 & 6.10 & 6.50 & 6.10 & 6.10 & 14.42 & 6.11 & 10.27 \\
synthetic\_n300\_d2\_k20\_s20 & 26.32 & 21.38 & 21.05 & 22.34 & 21.45 & 25.78 & 22.12 & 27.29 \\
synthetic\_n300\_d2\_k20\_s30 & 43.20 & 37.80 & 37.31 & 38.61 & 38.23 & 41.00 & 38.22 & 43.99 \\
synthetic\_n300\_d2\_k20\_s40 & 60.25 & 52.74 & 52.63 & 54.01 & 52.93 & 57.09 & 52.95 & 62.91 \\
synthetic\_n300\_d2\_k20\_s50 & 73.11 & 65.71 & 64.47 & 68.82 & 66.02 & 71.72 & 66.59 & 75.05 \\
synthetic\_n300\_d2\_k50\_s10 & 4.12 & 3.63 & 3.66 & 3.72 & 3.69 & 5.65 & 3.74 & 4.19 \\
synthetic\_n300\_d2\_k50\_s20 & 10.20 & 9.10 & 9.17 & 9.63 & 9.46 & 11.18 & 9.80 & 10.17 \\
synthetic\_n300\_d2\_k50\_s30 & 15.11 & 13.62 & 13.67 & 14.50 & 14.30 & 15.59 & 14.06 & 15.17 \\
synthetic\_n300\_d2\_k50\_s40 & 19.72 & 17.64 & 17.81 & 18.58 & 18.54 & 20.53 & 18.44 & 19.95 \\
synthetic\_n300\_d2\_k50\_s50 & 24.76 & 22.96 & 22.85 & 24.22 & 23.71 & 26.10 & 23.53 & 24.86 \\
\bottomrule
\end{tabular}
}
\end{table}
\input{tables/COL2/COL2_0.2_time}
\input{tables/COL2/COL2_0.2_ARI}
\input{tables/COL2/COL2_0.2_rate_success}
\input{tables/COL2/COL2_0.25_obj}
\input{tables/COL2/COL2_0.25_time}
\input{tables/COL2/COL2_0.25_ARI}
\input{tables/COL2/COL2_0.25_rate_success}
\input{tables/COL2/COL2_0.3_obj}
\input{tables/COL2/COL2_0.3_time}
\input{tables/COL2/COL2_0.3_ARI}
\input{tables/COL2/COL2_0.3_rate_success}
\input{tables/COL2/COL2_0.35_obj}
\input{tables/COL2/COL2_0.35_time}
\input{tables/COL2/COL2_0.35_ARI}
\input{tables/COL2/COL2_0.35_rate_success}
\input{tables/COL2/COL2_0.4_obj}
\input{tables/COL2/COL2_0.4_time}
\input{tables/COL2/COL2_0.4_ARI}
\input{tables/COL2/COL2_0.4_rate_success}
\input{tables/COL2/COL2_0.45_obj}
\input{tables/COL2/COL2_0.45_time}
\input{tables/COL2/COL2_0.45_ARI}
\input{tables/COL2/COL2_0.45_rate_success}
\input{tables/COL2/COL2_0.5_obj}
\input{tables/COL2/COL2_0.5_time}
\input{tables/COL2/COL2_0.5_ARI}
\input{tables/COL2/COL2_0.5_rate_success}
\begin{table}[H]
\caption{Clustering inertia on dataset collection 3 with 5.0\% constraints}
\label{tab:obj_COL3_5}
\centering
\resizebox{0.9\textwidth}{!}{%
\begin{tabular}{c|cccccc}
\toprule
Constraints Level & \multicolumn{6}{c}{0.05} \\
Method & COPKM\_P & KSKM & KSKM\_E & BLPKMCC & PCCC & COPKM \\
Dataset &  &  &  &  &  &  \\
\midrule
synthetic\_n1000\_d2\_k10 & 96.58 & 64.48 & 64.48 & 69.81 & 76.71 & 104.15 \\
synthetic\_n1000\_d2\_k100 & 16.65 & 15.13 & 15.19 & 18.67 & 15.91 & 16.62 \\
synthetic\_n1000\_d2\_k2 & 911.71 & 911.71 & 911.71 & 911.71 & 912.38 & 911.71 \\
synthetic\_n1000\_d2\_k20 & 68.94 & 64.21 & 63.92 & 69.79 & 64.42 & 69.95 \\
synthetic\_n1000\_d2\_k5 & 70.99 & 69.54 & 68.58 & 66.17 & 66.18 & 66.86 \\
\bottomrule
\end{tabular}
}
\end{table}

\begin{table}[H]
\caption{CPU time on dataset collection 3 with 5.0\% constraints}
\label{tab:time_COL3_5}
\centering
\resizebox{0.9\textwidth}{!}{%
\begin{tabular}{c|cccccc}
\toprule
Constraints Level & \multicolumn{6}{c}{0.05} \\
Method & COPKM\_P & KSKM & KSKM\_E & BLPKMCC & PCCC & COPKM \\
Dataset &  &  &  &  &  &  \\
\midrule
synthetic\_n1000\_d2\_k10 & 0.05 & 0.20 & 1.51 & 2.08 & 2.05 & 0.07 \\
synthetic\_n1000\_d2\_k100 & 0.25 & 1.12 & 56.19 & 30.88 & 76.61 & 0.47 \\
synthetic\_n1000\_d2\_k2 & 0.02 & 0.02 & 0.05 & 0.13 & 0.15 & 0.02 \\
synthetic\_n1000\_d2\_k20 & 0.08 & 0.45 & 8.22 & 7.75 & 16.59 & 0.15 \\
synthetic\_n1000\_d2\_k5 & 0.04 & 0.13 & 0.22 & 0.81 & 1.95 & 0.04 \\
synthetic\_n1000\_d2\_k50 & 0.09 & 0.74 & 31.43 & 19.84 & 33.10 & 0.19 \\
synthetic\_n2000\_d2\_k10 & 0.10 & 0.40 & 1.57 & 10.20 & 8.32 & 0.12 \\
synthetic\_n2000\_d2\_k100 & 0.41 & 2.21 & 240.94 & 221.15 & 260.91 & 0.77 \\
synthetic\_n2000\_d2\_k2 & 0.02 & 0.03 & 0.04 & 0.37 & 0.14 & 0.02 \\
synthetic\_n2000\_d2\_k20 & 0.20 & 1.17 & 23.97 & 43.12 & 51.88 & 0.28 \\
synthetic\_n2000\_d2\_k5 & 0.06 & 0.23 & 0.38 & 1.80 & 2.77 & 0.05 \\
synthetic\_n2000\_d2\_k50 & 0.26 & 1.81 & 129.70 & 119.06 & 163.28 & 0.45 \\
synthetic\_n5000\_d2\_k10 & 0.15 & 0.14 & 2.01 & 25.18 & 47.63 & 0.12 \\
synthetic\_n5000\_d2\_k100 & 1.33 & 15.04 & 1970.20 & 2101.76 & 2728.27 & 2.17 \\
synthetic\_n5000\_d2\_k2 & 0.03 & 0.04 & 0.05 & 1.69 & 0.11 & 0.03 \\
synthetic\_n5000\_d2\_k20 & 0.53 & 2.67 & 163.93 & 388.28 & 394.30 & - \\
synthetic\_n5000\_d2\_k5 & 0.06 & 0.08 & 0.29 & 6.90 & 1.95 & 0.06 \\
synthetic\_n5000\_d2\_k50 & 0.98 & 7.79 & 793.07 & 1120.95 & 1212.75 & 1.26 \\
synthetic\_n500\_d2\_k10 & 0.02 & 0.17 & 0.54 & 0.68 & 2.08 & 0.04 \\
synthetic\_n500\_d2\_k100 & 0.04 & 0.59 & 13.15 & 7.40 & 15.21 & 0.09 \\
synthetic\_n500\_d2\_k2 & 0.01 & 0.02 & 0.03 & 0.07 & 0.18 & 0.01 \\
synthetic\_n500\_d2\_k20 & 0.03 & 0.37 & 3.20 & 2.18 & 4.99 & 0.05 \\
synthetic\_n500\_d2\_k5 & 0.02 & 0.12 & 0.19 & 0.28 & 0.82 & 0.02 \\
synthetic\_n500\_d2\_k50 & 0.03 & 0.44 & 8.22 & 3.94 & 10.03 & 0.06 \\
\bottomrule
\end{tabular}
}
\end{table}

\begin{table}[H]
\caption{ARI on dataset collection 3 with 5.0\% constraints}
\label{tab:ARI_COL3_5}
\centering
\resizebox{0.9\textwidth}{!}{%
\begin{tabular}{c|cccccc}
\toprule
Constraints Level & \multicolumn{6}{c}{0.05} \\
Method & COPKM\_P & KSKM & KSKM\_E & BLPKMCC & PCCC & COPKM \\
Dataset &  &  &  &  &  &  \\
\midrule
synthetic\_n1000\_d2\_k10 & 0.86 & 0.93 & 0.93 & 0.90 & 0.90 & 0.85 \\
synthetic\_n1000\_d2\_k100 & 0.24 & 0.25 & 0.25 & 0.25 & 0.24 & 0.24 \\
synthetic\_n1000\_d2\_k2 & 1.00 & 1.00 & 1.00 & 1.00 & 1.00 & 1.00 \\
synthetic\_n1000\_d2\_k20 & 0.58 & 0.64 & 0.65 & 0.59 & 0.64 & 0.58 \\
synthetic\_n1000\_d2\_k5 & 0.92 & 0.93 & 0.93 & 0.93 & 0.93 & 0.92 \\
synthetic\_n1000\_d2\_k50 & 0.41 & 0.43 & 0.43 & 0.41 & 0.42 & 0.41 \\
synthetic\_n2000\_d2\_k10 & 0.88 & 0.94 & 0.94 & 0.93 & 0.94 & 0.93 \\
synthetic\_n2000\_d2\_k100 & 0.25 & 0.25 & 0.25 & 0.26 & 0.25 & 0.25 \\
synthetic\_n2000\_d2\_k2 & 1.00 & 1.00 & 1.00 & 1.00 & 1.00 & 1.00 \\
synthetic\_n2000\_d2\_k20 & 0.60 & 0.68 & 0.68 & 0.63 & 0.68 & 0.61 \\
synthetic\_n2000\_d2\_k5 & 0.95 & 0.94 & 0.94 & 0.94 & 0.94 & 0.95 \\
synthetic\_n2000\_d2\_k50 & 0.43 & 0.44 & 0.44 & 0.44 & 0.44 & 0.43 \\
synthetic\_n5000\_d2\_k10 & 0.98 & 0.98 & 0.98 & 0.98 & 0.98 & 0.98 \\
synthetic\_n5000\_d2\_k100 & 0.25 & 0.26 & 0.26 & 0.26 & 0.26 & 0.25 \\
synthetic\_n5000\_d2\_k2 & 1.00 & 1.00 & 1.00 & 1.00 & 1.00 & 1.00 \\
synthetic\_n5000\_d2\_k20 & 0.68 & 0.79 & 0.81 & 0.74 & 0.81 & - \\
synthetic\_n5000\_d2\_k5 & 1.00 & 1.00 & 1.00 & 1.00 & 1.00 & 1.00 \\
synthetic\_n5000\_d2\_k50 & 0.46 & 0.50 & 0.50 & 0.48 & 0.50 & 0.46 \\
synthetic\_n500\_d2\_k10 & 0.86 & 0.90 & 0.90 & 0.79 & 0.90 & 0.85 \\
synthetic\_n500\_d2\_k100 & 0.22 & 0.21 & 0.22 & 0.22 & 0.21 & 0.22 \\
synthetic\_n500\_d2\_k2 & 1.00 & 1.00 & 1.00 & 0.99 & 1.00 & 0.99 \\
synthetic\_n500\_d2\_k20 & 0.55 & 0.60 & 0.60 & 0.54 & 0.60 & 0.56 \\
synthetic\_n500\_d2\_k5 & 0.86 & 0.91 & 0.91 & 0.91 & 0.91 & 0.83 \\
synthetic\_n500\_d2\_k50 & 0.38 & 0.39 & 0.40 & 0.38 & 0.40 & 0.38 \\
\bottomrule
\end{tabular}
}
\end{table}

\begin{table}[H]
\caption{Algorithm success rate on dataset collection 3 with 5.0\% constraints}
\label{tab:rate_success_COL3_5}
\centering
\resizebox{0.9\textwidth}{!}{%
\begin{tabular}{c|cccccc}
\toprule
Constraints Level & \multicolumn{6}{c}{0.05} \\
Method & COPKM\_P & KSKM & KSKM\_E & BLPKMCC & PCCC & COPKM \\
Dataset &  &  &  &  &  &  \\
\midrule
synthetic\_n1000\_d2\_k10 & 1.00 & 1.00 & 1.00 & 1.00 & 1.00 & 1.00 \\
synthetic\_n1000\_d2\_k100 & 1.00 & 1.00 & 1.00 & 1.00 & 1.00 & 1.00 \\
synthetic\_n1000\_d2\_k2 & 1.00 & 1.00 & 1.00 & 1.00 & 1.00 & 0.80 \\
synthetic\_n1000\_d2\_k20 & 1.00 & 1.00 & 1.00 & 1.00 & 1.00 & 1.00 \\
synthetic\_n1000\_d2\_k5 & 1.00 & 1.00 & 1.00 & 1.00 & 1.00 & 0.70 \\
synthetic\_n1000\_d2\_k50 & 1.00 & 1.00 & 1.00 & 1.00 & 1.00 & 1.00 \\
synthetic\_n2000\_d2\_k10 & 1.00 & 1.00 & 1.00 & 1.00 & 1.00 & 0.80 \\
synthetic\_n2000\_d2\_k100 & 1.00 & 1.00 & 1.00 & 1.00 & 1.00 & 1.00 \\
synthetic\_n2000\_d2\_k2 & 1.00 & 1.00 & 1.00 & 1.00 & 1.00 & 1.00 \\
synthetic\_n2000\_d2\_k20 & 1.00 & 1.00 & 1.00 & 1.00 & 1.00 & 1.00 \\
synthetic\_n2000\_d2\_k5 & 1.00 & 1.00 & 1.00 & 1.00 & 1.00 & 0.60 \\
synthetic\_n2000\_d2\_k50 & 1.00 & 1.00 & 1.00 & 1.00 & 1.00 & 1.00 \\
synthetic\_n5000\_d2\_k10 & 1.00 & 1.00 & 1.00 & 1.00 & 1.00 & 0.60 \\
synthetic\_n5000\_d2\_k100 & 1.00 & 1.00 & 1.00 & 1.00 & 1.00 & 1.00 \\
synthetic\_n5000\_d2\_k2 & 1.00 & 1.00 & 1.00 & 1.00 & 1.00 & 1.00 \\
synthetic\_n5000\_d2\_k20 & 1.00 & 1.00 & 1.00 & 1.00 & 1.00 & 0.00 \\
synthetic\_n5000\_d2\_k5 & 1.00 & 1.00 & 1.00 & 1.00 & 1.00 & 0.90 \\
synthetic\_n5000\_d2\_k50 & 1.00 & 1.00 & 1.00 & 1.00 & 1.00 & 1.00 \\
synthetic\_n500\_d2\_k10 & 1.00 & 1.00 & 1.00 & 1.00 & 1.00 & 1.00 \\
synthetic\_n500\_d2\_k100 & 1.00 & 1.00 & 1.00 & 1.00 & 1.00 & 1.00 \\
synthetic\_n500\_d2\_k2 & 1.00 & 1.00 & 1.00 & 1.00 & 1.00 & 0.80 \\
synthetic\_n500\_d2\_k20 & 1.00 & 1.00 & 1.00 & 1.00 & 1.00 & 1.00 \\
synthetic\_n500\_d2\_k5 & 1.00 & 1.00 & 1.00 & 1.00 & 1.00 & 1.00 \\
synthetic\_n500\_d2\_k50 & 1.00 & 1.00 & 1.00 & 1.00 & 1.00 & 1.00 \\
\bottomrule
\end{tabular}
}
\end{table}

\begin{table}[H]
\caption{Clustering inertia on dataset collection 3 with 10.0\% constraints}
\label{tab:obj_COL3_10}
\centering
\resizebox{0.9\textwidth}{!}{%
\begin{tabular}{c|cccccc}
\toprule
Constraints Level & \multicolumn{6}{c}{0.1} \\
Method & COPKM\_P & KSKM & KSKM\_E & BLPKMCC & PCCC & COPKM \\
Dataset &  &  &  &  &  &  \\
\midrule
synthetic\_n1000\_d2\_k10 & 66.18 & 97.61 & 66.18 & 66.18 & 66.19 & 66.18 \\
synthetic\_n1000\_d2\_k100 & 20.02 & 18.54 & 18.57 & 21.90 & 19.43 & 20.06 \\
synthetic\_n1000\_d2\_k2 & 911.71 & 911.71 & 911.71 & 911.71 & 911.71 & 911.71 \\
synthetic\_n1000\_d2\_k20 & 82.95 & 72.03 & 71.18 & 83.39 & 71.91 & 91.01 \\
synthetic\_n1000\_d2\_k5 & 69.11 & 69.11 & 69.11 & 69.11 & 69.11 & 69.11 \\
\bottomrule
\end{tabular}
}
\end{table}

\begin{table}[H]
\caption{CPU time on dataset collection 3 with 10.0\% constraints}
\label{tab:time_COL3_10}
\centering
\resizebox{0.9\textwidth}{!}{%
\begin{tabular}{c|cccccc}
\toprule
Constraints Level & \multicolumn{6}{c}{0.1} \\
Method & COPKM\_P & KSKM & KSKM\_E & BLPKMCC & PCCC & COPKM \\
Dataset &  &  &  &  &  &  \\
\midrule
synthetic\_n1000\_d2\_k10 & 0.03 & 0.19 & 0.60 & 4.39 & 7.30 & 0.04 \\
synthetic\_n1000\_d2\_k100 & 0.12 & 1.32 & 239.52 & 124.82 & 195.77 & 0.28 \\
synthetic\_n1000\_d2\_k2 & 0.01 & 0.01 & 0.01 & 0.29 & 0.02 & 0.01 \\
synthetic\_n1000\_d2\_k20 & 0.06 & 0.80 & 19.88 & 23.30 & 43.07 & 0.08 \\
synthetic\_n1000\_d2\_k5 & 0.02 & 0.04 & 0.08 & 1.01 & 0.49 & 0.01 \\
synthetic\_n1000\_d2\_k50 & 0.10 & 0.96 & 90.60 & 64.56 & 109.20 & 0.18 \\
synthetic\_n2000\_d2\_k10 & 0.04 & 0.06 & 0.43 & 10.37 & 5.20 & 0.04 \\
synthetic\_n2000\_d2\_k100 & 0.37 & 4.65 & 1176.05 & 610.46 & 1322.20 & 0.69 \\
synthetic\_n2000\_d2\_k2 & 0.01 & 0.02 & 0.02 & 0.81 & 0.04 & 0.01 \\
synthetic\_n2000\_d2\_k20 & 0.10 & 0.51 & 57.24 & 79.62 & 88.49 & - \\
synthetic\_n2000\_d2\_k5 & 0.02 & 0.03 & 0.07 & 2.88 & 0.30 & 0.02 \\
synthetic\_n2000\_d2\_k50 & 0.18 & 3.83 & 382.86 & 308.61 & 511.08 & 0.23 \\
synthetic\_n5000\_d2\_k10 & 0.07 & 0.06 & 0.23 & 36.82 & 1.99 & 0.07 \\
synthetic\_n5000\_d2\_k100 & 0.96 & 44.15 & 3606.06 & 3674.76 & 3753.36 & 1.31 \\
synthetic\_n5000\_d2\_k2 & 0.05 & 0.05 & 0.05 & 4.91 & 0.20 & 0.06 \\
synthetic\_n5000\_d2\_k20 & 0.11 & 0.27 & 7.48 & 128.34 & 33.82 & 0.14 \\
synthetic\_n5000\_d2\_k5 & 0.07 & 0.06 & 0.08 & 13.56 & 0.18 & 0.06 \\
synthetic\_n5000\_d2\_k50 & 0.39 & 6.38 & 3602.02 & 3585.40 & 3405.78 & - \\
synthetic\_n500\_d2\_k10 & 0.02 & 0.08 & 0.48 & 1.92 & 2.76 & 0.02 \\
synthetic\_n500\_d2\_k100 & 0.05 & 0.67 & 48.53 & 17.36 & 35.25 & 0.09 \\
synthetic\_n500\_d2\_k2 & 0.01 & 0.01 & 0.01 & 0.10 & 0.03 & 0.01 \\
synthetic\_n500\_d2\_k20 & 0.03 & 0.31 & 10.29 & 4.70 & 9.36 & 0.04 \\
synthetic\_n500\_d2\_k5 & 0.01 & 0.03 & 0.09 & 0.35 & 0.62 & 0.01 \\
synthetic\_n500\_d2\_k50 & 0.04 & 0.44 & 26.97 & 11.35 & 24.37 & 0.07 \\
\bottomrule
\end{tabular}
}
\end{table}

\begin{table}[H]
\caption{ARI on dataset collection 3 with 10.0\% constraints}
\label{tab:ARI_COL3_10}
\centering
\resizebox{0.9\textwidth}{!}{%
\begin{tabular}{c|cccccc}
\toprule
Constraints Level & \multicolumn{6}{c}{0.1} \\
Method & COPKM\_P & KSKM & KSKM\_E & BLPKMCC & PCCC & COPKM \\
Dataset &  &  &  &  &  &  \\
\midrule
synthetic\_n1000\_d2\_k10 & 0.98 & 0.95 & 0.98 & 0.98 & 0.98 & 0.98 \\
synthetic\_n1000\_d2\_k100 & 0.26 & 0.26 & 0.26 & 0.27 & 0.26 & 0.26 \\
synthetic\_n1000\_d2\_k2 & 1.00 & 1.00 & 1.00 & 1.00 & 1.00 & 1.00 \\
synthetic\_n1000\_d2\_k20 & 0.66 & 0.75 & 0.76 & 0.66 & 0.75 & 0.62 \\
synthetic\_n1000\_d2\_k5 & 0.99 & 0.99 & 0.99 & 0.99 & 0.99 & 0.99 \\
synthetic\_n1000\_d2\_k50 & 0.43 & 0.45 & 0.45 & 0.44 & 0.44 & 0.43 \\
synthetic\_n2000\_d2\_k10 & 0.99 & 0.99 & 0.99 & 0.99 & 0.99 & 0.99 \\
synthetic\_n2000\_d2\_k100 & 0.27 & 0.28 & 0.28 & 0.28 & 0.28 & 0.27 \\
synthetic\_n2000\_d2\_k2 & 1.00 & 1.00 & 1.00 & 1.00 & 1.00 & 1.00 \\
synthetic\_n2000\_d2\_k20 & 0.84 & 0.89 & 0.90 & 0.90 & 0.90 & - \\
synthetic\_n2000\_d2\_k5 & 1.00 & 1.00 & 1.00 & 1.00 & 1.00 & 1.00 \\
synthetic\_n2000\_d2\_k50 & 0.47 & 0.53 & 0.53 & 0.50 & 0.52 & 0.44 \\
synthetic\_n5000\_d2\_k10 & 1.00 & 1.00 & 1.00 & 1.00 & 1.00 & 1.00 \\
synthetic\_n5000\_d2\_k100 & 0.32 & 0.36 & 0.36 & 0.34 & 0.35 & 0.32 \\
synthetic\_n5000\_d2\_k2 & 1.00 & 1.00 & 1.00 & 1.00 & 1.00 & 1.00 \\
synthetic\_n5000\_d2\_k20 & 0.98 & 0.98 & 0.98 & 0.98 & 0.98 & 0.98 \\
synthetic\_n5000\_d2\_k5 & 1.00 & 1.00 & 1.00 & 1.00 & 1.00 & 1.00 \\
synthetic\_n5000\_d2\_k50 & 0.71 & 0.74 & 0.78 & 0.58 & 0.71 & - \\
synthetic\_n500\_d2\_k10 & 0.92 & 0.95 & 0.95 & 0.82 & 0.95 & 0.83 \\
synthetic\_n500\_d2\_k100 & 0.23 & 0.22 & 0.22 & 0.22 & 0.22 & 0.23 \\
synthetic\_n500\_d2\_k2 & 1.00 & 1.00 & 1.00 & 1.00 & 1.00 & 1.00 \\
synthetic\_n500\_d2\_k20 & 0.58 & 0.64 & 0.64 & 0.58 & 0.63 & 0.57 \\
synthetic\_n500\_d2\_k5 & 0.97 & 0.97 & 0.97 & 0.97 & 0.97 & 0.96 \\
synthetic\_n500\_d2\_k50 & 0.39 & 0.41 & 0.40 & 0.38 & 0.40 & 0.39 \\
\bottomrule
\end{tabular}
}
\end{table}

\begin{table}[H]
\caption{Algorithm success rate on dataset collection 3 with 10.0\% constraints}
\label{tab:rate_success_COL3_10}
\centering
\resizebox{0.9\textwidth}{!}{%
\begin{tabular}{c|cccccc}
\toprule
Constraints Level & \multicolumn{6}{c}{0.1} \\
Method & COPKM\_P & KSKM & KSKM\_E & BLPKMCC & PCCC & COPKM \\
Dataset &  &  &  &  &  &  \\
\midrule
synthetic\_n1000\_d2\_k10 & 1.00 & 1.00 & 1.00 & 1.00 & 1.00 & 0.20 \\
synthetic\_n1000\_d2\_k100 & 1.00 & 1.00 & 1.00 & 1.00 & 1.00 & 1.00 \\
synthetic\_n1000\_d2\_k2 & 1.00 & 1.00 & 1.00 & 1.00 & 1.00 & 1.00 \\
synthetic\_n1000\_d2\_k20 & 1.00 & 1.00 & 1.00 & 1.00 & 1.00 & 0.70 \\
synthetic\_n1000\_d2\_k5 & 1.00 & 1.00 & 1.00 & 1.00 & 1.00 & 1.00 \\
synthetic\_n1000\_d2\_k50 & 1.00 & 1.00 & 1.00 & 1.00 & 1.00 & 1.00 \\
synthetic\_n2000\_d2\_k10 & 1.00 & 1.00 & 1.00 & 1.00 & 1.00 & 1.00 \\
synthetic\_n2000\_d2\_k100 & 1.00 & 1.00 & 1.00 & 1.00 & 1.00 & 1.00 \\
synthetic\_n2000\_d2\_k2 & 1.00 & 1.00 & 1.00 & 1.00 & 1.00 & 1.00 \\
synthetic\_n2000\_d2\_k20 & 1.00 & 1.00 & 1.00 & 1.00 & 1.00 & 0.00 \\
synthetic\_n2000\_d2\_k5 & 1.00 & 1.00 & 1.00 & 1.00 & 1.00 & 1.00 \\
synthetic\_n2000\_d2\_k50 & 1.00 & 1.00 & 1.00 & 1.00 & 1.00 & 1.00 \\
synthetic\_n5000\_d2\_k10 & 1.00 & 1.00 & 1.00 & 1.00 & 1.00 & 1.00 \\
synthetic\_n5000\_d2\_k100 & 1.00 & 1.00 & 1.00 & 0.40 & 1.00 & 1.00 \\
synthetic\_n5000\_d2\_k2 & 1.00 & 1.00 & 1.00 & 1.00 & 1.00 & 1.00 \\
synthetic\_n5000\_d2\_k20 & 1.00 & 1.00 & 1.00 & 1.00 & 1.00 & 0.60 \\
synthetic\_n5000\_d2\_k5 & 1.00 & 1.00 & 1.00 & 1.00 & 1.00 & 1.00 \\
synthetic\_n5000\_d2\_k50 & 1.00 & 1.00 & 1.00 & 0.80 & 1.00 & 0.00 \\
synthetic\_n500\_d2\_k10 & 1.00 & 1.00 & 1.00 & 1.00 & 1.00 & 0.90 \\
synthetic\_n500\_d2\_k100 & 1.00 & 1.00 & 1.00 & 1.00 & 1.00 & 1.00 \\
synthetic\_n500\_d2\_k2 & 1.00 & 1.00 & 1.00 & 1.00 & 1.00 & 1.00 \\
synthetic\_n500\_d2\_k20 & 1.00 & 1.00 & 1.00 & 1.00 & 1.00 & 1.00 \\
synthetic\_n500\_d2\_k5 & 1.00 & 1.00 & 1.00 & 1.00 & 1.00 & 0.50 \\
synthetic\_n500\_d2\_k50 & 1.00 & 1.00 & 1.00 & 1.00 & 1.00 & 1.00 \\
\bottomrule
\end{tabular}
}
\end{table}

\begin{table}[H]
\caption{Clustering inertia on dataset collection 3 with 15.0\% constraints}
\label{tab:obj_COL3_15}
\centering
\resizebox{0.9\textwidth}{!}{%
\begin{tabular}{c|cccccc}
\toprule
Constraints Level & \multicolumn{6}{c}{0.15} \\
Method & COPKM\_P & KSKM & KSKM\_E & BLPKMCC & PCCC & COPKM \\
Dataset &  &  &  &  &  &  \\
\midrule
synthetic\_n1000\_d2\_k10 & 66.91 & 66.91 & 66.91 & 66.91 & 66.99 & 66.91 \\
synthetic\_n1000\_d2\_k100 & 24.02 & 21.46 & 21.34 & 25.31 & 22.28 & 24.21 \\
synthetic\_n1000\_d2\_k2 & 911.71 & 911.71 & 911.71 & 911.71 & 911.71 & 911.71 \\
synthetic\_n1000\_d2\_k20 & 93.93 & 97.63 & 77.18 & 77.18 & 77.26 & - \\
synthetic\_n1000\_d2\_k5 & 69.58 & 69.58 & 69.58 & 69.58 & 69.58 & 69.58 \\
\bottomrule
\end{tabular}
}
\end{table}

\begin{table}[H]
\caption{CPU time on dataset collection 3 with 15.0\% constraints}
\label{tab:time_COL3_15}
\centering
\resizebox{0.9\textwidth}{!}{%
\begin{tabular}{c|cccccc}
\toprule
Constraints Level & \multicolumn{6}{c}{0.15} \\
Method & COPKM\_P & KSKM & KSKM\_E & BLPKMCC & PCCC & COPKM \\
Dataset &  &  &  &  &  &  \\
\midrule
synthetic\_n1000\_d2\_k10 & 0.02 & 0.10 & 0.34 & 4.40 & 1.48 & 0.02 \\
synthetic\_n1000\_d2\_k100 & 0.10 & 1.72 & 594.32 & 205.16 & 521.36 & 0.24 \\
synthetic\_n1000\_d2\_k2 & 0.01 & 0.01 & 0.01 & 0.46 & 0.02 & 0.01 \\
synthetic\_n1000\_d2\_k20 & 0.04 & 0.30 & 27.09 & 32.45 & 34.15 & - \\
synthetic\_n1000\_d2\_k5 & 0.01 & 0.02 & 0.03 & 1.39 & 0.08 & 0.01 \\
synthetic\_n1000\_d2\_k50 & 0.08 & 1.29 & 257.92 & 132.12 & 281.48 & 0.15 \\
synthetic\_n2000\_d2\_k10 & 0.03 & 0.05 & 0.17 & 12.95 & 1.00 & 0.03 \\
synthetic\_n2000\_d2\_k100 & 0.30 & 7.78 & 3430.25 & 1461.52 & 3076.50 & 0.53 \\
synthetic\_n2000\_d2\_k2 & 0.02 & 0.02 & 0.02 & 1.71 & 0.07 & 0.02 \\
synthetic\_n2000\_d2\_k20 & 0.04 & 0.25 & 4.35 & 42.56 & 16.23 & 0.06 \\
synthetic\_n2000\_d2\_k5 & 0.02 & 0.03 & 0.03 & 4.29 & 0.07 & 0.02 \\
synthetic\_n2000\_d2\_k50 & 0.16 & 3.28 & 1184.14 & 765.53 & 1041.88 & - \\
synthetic\_n5000\_d2\_k10 & 0.08 & 0.07 & 0.09 & 56.68 & 0.38 & 0.09 \\
synthetic\_n5000\_d2\_k100 & 0.73 & 21.58 & 3613.08 & 3849.72 & 3760.80 & - \\
synthetic\_n5000\_d2\_k2 & 0.09 & 0.09 & 0.09 & 10.94 & 0.44 & 0.10 \\
synthetic\_n5000\_d2\_k20 & 0.12 & 0.19 & 0.71 & 177.54 & 5.89 & 0.12 \\
synthetic\_n5000\_d2\_k5 & 0.10 & 0.10 & 0.10 & 27.68 & 0.35 & 0.10 \\
synthetic\_n5000\_d2\_k50 & 0.22 & 1.68 & 230.16 & 904.22 & 274.53 & - \\
synthetic\_n500\_d2\_k10 & 0.01 & 0.05 & 0.27 & 2.02 & 2.48 & 0.01 \\
synthetic\_n500\_d2\_k100 & 0.05 & 0.50 & 84.54 & 40.78 & 89.26 & 0.10 \\
synthetic\_n500\_d2\_k2 & 0.00 & 0.01 & 0.01 & 0.14 & 0.03 & 0.00 \\
synthetic\_n500\_d2\_k20 & 0.03 & 0.23 & 12.11 & 8.59 & 20.19 & 0.04 \\
synthetic\_n500\_d2\_k5 & 0.01 & 0.03 & 0.04 & 0.46 & 0.18 & 0.01 \\
synthetic\_n500\_d2\_k50 & 0.04 & 0.49 & 47.03 & 21.36 & 50.55 & 0.07 \\
\bottomrule
\end{tabular}
}
\end{table}

\begin{table}[H]
\caption{ARI on dataset collection 3 with 15.0\% constraints}
\label{tab:ARI_COL3_15}
\centering
\resizebox{0.9\textwidth}{!}{%
\begin{tabular}{c|cccccc}
\toprule
Constraints Level & \multicolumn{6}{c}{0.15} \\
Method & COPKM\_P & KSKM & KSKM\_E & BLPKMCC & PCCC & COPKM \\
Dataset &  &  &  &  &  &  \\
\midrule
synthetic\_n1000\_d2\_k10 & 1.00 & 1.00 & 1.00 & 1.00 & 1.00 & 1.00 \\
synthetic\_n1000\_d2\_k100 & 0.28 & 0.29 & 0.29 & 0.30 & 0.29 & 0.28 \\
synthetic\_n1000\_d2\_k2 & 1.00 & 1.00 & 1.00 & 1.00 & 1.00 & 1.00 \\
synthetic\_n1000\_d2\_k20 & 0.86 & 0.85 & 0.89 & 0.89 & 0.89 & - \\
synthetic\_n1000\_d2\_k5 & 1.00 & 1.00 & 1.00 & 1.00 & 1.00 & 1.00 \\
synthetic\_n1000\_d2\_k50 & 0.48 & 0.54 & 0.56 & 0.48 & 0.54 & 0.48 \\
synthetic\_n2000\_d2\_k10 & 1.00 & 1.00 & 1.00 & 1.00 & 1.00 & 1.00 \\
synthetic\_n2000\_d2\_k100 & 0.33 & 0.35 & 0.35 & 0.34 & 0.34 & 0.32 \\
synthetic\_n2000\_d2\_k2 & 1.00 & 1.00 & 1.00 & 1.00 & 1.00 & 1.00 \\
synthetic\_n2000\_d2\_k20 & 0.97 & 0.97 & 0.97 & 0.97 & 0.98 & 0.97 \\
synthetic\_n2000\_d2\_k5 & 1.00 & 1.00 & 1.00 & 1.00 & 1.00 & 1.00 \\
synthetic\_n2000\_d2\_k50 & 0.66 & 0.75 & 0.77 & 0.72 & 0.73 & - \\
synthetic\_n5000\_d2\_k10 & 1.00 & 1.00 & 1.00 & 1.00 & 1.00 & 1.00 \\
synthetic\_n5000\_d2\_k100 & 0.65 & 0.66 & 0.65 & 0.53 & 0.58 & - \\
synthetic\_n5000\_d2\_k2 & 1.00 & 1.00 & 1.00 & 1.00 & 1.00 & 1.00 \\
synthetic\_n5000\_d2\_k20 & 1.00 & 1.00 & 1.00 & 1.00 & 1.00 & 1.00 \\
synthetic\_n5000\_d2\_k5 & 1.00 & 1.00 & 1.00 & 1.00 & 1.00 & 1.00 \\
synthetic\_n5000\_d2\_k50 & 0.95 & 0.95 & 0.95 & 0.95 & 0.95 & - \\
synthetic\_n500\_d2\_k10 & 0.98 & 0.98 & 0.98 & 0.98 & 0.98 & 0.98 \\
synthetic\_n500\_d2\_k100 & 0.24 & 0.24 & 0.25 & 0.24 & 0.24 & 0.24 \\
synthetic\_n500\_d2\_k2 & 1.00 & 1.00 & 1.00 & 1.00 & 1.00 & 1.00 \\
synthetic\_n500\_d2\_k20 & 0.67 & 0.76 & 0.78 & 0.66 & 0.77 & 0.64 \\
synthetic\_n500\_d2\_k5 & 0.99 & 0.99 & 0.99 & 0.99 & 0.99 & 0.99 \\
synthetic\_n500\_d2\_k50 & 0.42 & 0.44 & 0.45 & 0.42 & 0.44 & 0.42 \\
\bottomrule
\end{tabular}
}
\end{table}

\begin{table}[H]
\caption{Algorithm success rate on dataset collection 3 with 15.0\% constraints}
\label{tab:rate_success_COL3_15}
\centering
\resizebox{0.9\textwidth}{!}{%
\begin{tabular}{c|cccccc}
\toprule
Constraints Level & \multicolumn{6}{c}{0.15} \\
Method & COPKM\_P & KSKM & KSKM\_E & BLPKMCC & PCCC & COPKM \\
Dataset &  &  &  &  &  &  \\
\midrule
synthetic\_n1000\_d2\_k10 & 1.00 & 1.00 & 1.00 & 1.00 & 1.00 & 1.00 \\
synthetic\_n1000\_d2\_k100 & 1.00 & 1.00 & 1.00 & 1.00 & 1.00 & 1.00 \\
synthetic\_n1000\_d2\_k2 & 1.00 & 1.00 & 1.00 & 1.00 & 1.00 & 1.00 \\
synthetic\_n1000\_d2\_k20 & 1.00 & 1.00 & 1.00 & 1.00 & 1.00 & 0.00 \\
synthetic\_n1000\_d2\_k5 & 1.00 & 1.00 & 1.00 & 1.00 & 1.00 & 1.00 \\
synthetic\_n1000\_d2\_k50 & 1.00 & 1.00 & 1.00 & 1.00 & 1.00 & 1.00 \\
synthetic\_n2000\_d2\_k10 & 1.00 & 1.00 & 1.00 & 1.00 & 1.00 & 1.00 \\
synthetic\_n2000\_d2\_k100 & 1.00 & 1.00 & 1.00 & 1.00 & 1.00 & 1.00 \\
synthetic\_n2000\_d2\_k2 & 1.00 & 1.00 & 1.00 & 1.00 & 1.00 & 1.00 \\
synthetic\_n2000\_d2\_k20 & 1.00 & 1.00 & 1.00 & 1.00 & 1.00 & 0.80 \\
synthetic\_n2000\_d2\_k5 & 1.00 & 1.00 & 1.00 & 1.00 & 1.00 & 1.00 \\
synthetic\_n2000\_d2\_k50 & 1.00 & 1.00 & 1.00 & 1.00 & 1.00 & 0.00 \\
synthetic\_n5000\_d2\_k10 & 1.00 & 1.00 & 1.00 & 1.00 & 1.00 & 1.00 \\
synthetic\_n5000\_d2\_k100 & 1.00 & 1.00 & 1.00 & 0.80 & 0.50 & 0.00 \\
synthetic\_n5000\_d2\_k2 & 1.00 & 1.00 & 1.00 & 1.00 & 1.00 & 1.00 \\
synthetic\_n5000\_d2\_k20 & 1.00 & 1.00 & 1.00 & 1.00 & 1.00 & 1.00 \\
synthetic\_n5000\_d2\_k5 & 1.00 & 1.00 & 1.00 & 1.00 & 1.00 & 1.00 \\
synthetic\_n5000\_d2\_k50 & 1.00 & 1.00 & 1.00 & 1.00 & 1.00 & 0.00 \\
synthetic\_n500\_d2\_k10 & 1.00 & 1.00 & 1.00 & 1.00 & 1.00 & 0.20 \\
synthetic\_n500\_d2\_k100 & 1.00 & 1.00 & 1.00 & 1.00 & 1.00 & 1.00 \\
synthetic\_n500\_d2\_k2 & 1.00 & 1.00 & 1.00 & 1.00 & 1.00 & 1.00 \\
synthetic\_n500\_d2\_k20 & 1.00 & 1.00 & 1.00 & 1.00 & 1.00 & 1.00 \\
synthetic\_n500\_d2\_k5 & 1.00 & 1.00 & 1.00 & 1.00 & 1.00 & 1.00 \\
synthetic\_n500\_d2\_k50 & 1.00 & 1.00 & 1.00 & 1.00 & 1.00 & 1.00 \\
\bottomrule
\end{tabular}
}
\end{table}

\begin{table}[H]
\caption{Clustering inertia on dataset collection 3 with 20.0\% constraints}
\label{tab:obj_COL3_20}
\centering
\resizebox{0.9\textwidth}{!}{%
\begin{tabular}{c|cccccc}
\toprule
Constraints Level & \multicolumn{6}{c}{0.2} \\
Method & COPKM\_P & KSKM & KSKM\_E & BLPKMCC & PCCC & COPKM \\
Dataset &  &  &  &  &  &  \\
\midrule
synthetic\_n1000\_d2\_k10 & 66.68 & 66.68 & 66.68 & 66.68 & 66.68 & 66.68 \\
synthetic\_n1000\_d2\_k100 & 30.31 & 27.39 & 27.41 & 30.69 & 28.40 & 30.84 \\
synthetic\_n1000\_d2\_k2 & 911.71 & 911.71 & 911.71 & 911.71 & 911.71 & 911.71 \\
synthetic\_n1000\_d2\_k20 & 80.46 & 80.46 & 80.46 & 80.46 & 80.47 & 80.47 \\
synthetic\_n1000\_d2\_k5 & 69.58 & 69.58 & 69.58 & 69.58 & 69.58 & 69.58 \\
\bottomrule
\end{tabular}
}
\end{table}

\begin{table}[H]
\caption{CPU time on dataset collection 3 with 20.0\% constraints}
\label{tab:time_COL3_20}
\centering
\resizebox{0.9\textwidth}{!}{%
\begin{tabular}{c|cccccc}
\toprule
Constraints Level & \multicolumn{6}{c}{0.2} \\
Method & COPKM\_P & KSKM & KSKM\_E & BLPKMCC & PCCC & COPKM \\
Dataset &  &  &  &  &  &  \\
\midrule
synthetic\_n1000\_d2\_k10 & 0.02 & 0.05 & 0.11 & 5.65 & 0.60 & 0.01 \\
synthetic\_n1000\_d2\_k100 & 0.12 & 2.34 & 961.17 & 401.50 & 783.38 & 0.23 \\
synthetic\_n1000\_d2\_k2 & 0.01 & 0.01 & 0.01 & 0.79 & 0.03 & 0.01 \\
synthetic\_n1000\_d2\_k20 & 0.02 & 0.12 & 1.50 & 19.56 & 12.19 & 0.03 \\
synthetic\_n1000\_d2\_k5 & 0.01 & 0.02 & 0.02 & 1.91 & 0.04 & 0.01 \\
synthetic\_n1000\_d2\_k50 & 0.07 & 2.22 & 367.42 & 226.44 & 237.22 & 0.09 \\
synthetic\_n2000\_d2\_k10 & 0.04 & 0.04 & 0.06 & 15.20 & 0.14 & 0.03 \\
synthetic\_n2000\_d2\_k100 & 0.23 & 9.83 & 3602.33 & 2703.46 & 3459.61 & 0.29 \\
synthetic\_n2000\_d2\_k2 & 0.03 & 0.03 & 0.03 & 3.00 & 0.11 & 0.03 \\
synthetic\_n2000\_d2\_k20 & 0.04 & 0.13 & 0.61 & 49.03 & 5.71 & 0.05 \\
synthetic\_n2000\_d2\_k5 & 0.03 & 0.03 & 0.04 & 7.56 & 0.10 & 0.03 \\
synthetic\_n2000\_d2\_k50 & 0.10 & 0.60 & 225.42 & 439.99 & 248.53 & - \\
synthetic\_n5000\_d2\_k10 & 0.18 & 0.14 & 0.13 & 102.51 & 0.60 & 0.18 \\
synthetic\_n5000\_d2\_k100 & 0.39 & 2.91 & 3605.94 & 3835.05 & 2491.94 & - \\
synthetic\_n5000\_d2\_k2 & 0.15 & 0.13 & 0.14 & 19.61 & 0.77 & 0.16 \\
synthetic\_n5000\_d2\_k20 & 0.18 & 0.23 & 0.29 & 212.72 & 0.95 & 0.18 \\
synthetic\_n5000\_d2\_k5 & 0.17 & 0.16 & 0.16 & 49.88 & 0.63 & 0.17 \\
synthetic\_n5000\_d2\_k50 & 0.22 & 1.09 & 8.35 & 1047.91 & 75.83 & 0.24 \\
synthetic\_n500\_d2\_k10 & 0.01 & 0.04 & 0.15 & 2.10 & 1.39 & 0.01 \\
synthetic\_n500\_d2\_k100 & 0.05 & 1.00 & 164.41 & 60.44 & 135.65 & 0.10 \\
synthetic\_n500\_d2\_k2 & 0.00 & 0.01 & 0.01 & 0.21 & 0.01 & 0.00 \\
synthetic\_n500\_d2\_k20 & 0.02 & 0.14 & 3.30 & 11.56 & 15.78 & - \\
synthetic\_n500\_d2\_k5 & 0.01 & 0.02 & 0.03 & 0.67 & 0.08 & 0.00 \\
synthetic\_n500\_d2\_k50 & 0.04 & 0.78 & 86.30 & 42.13 & 82.82 & 0.06 \\
\bottomrule
\end{tabular}
}
\end{table}

\begin{table}[H]
\caption{ARI on dataset collection 3 with 20.0\% constraints}
\label{tab:ARI_COL3_20}
\centering
\resizebox{0.9\textwidth}{!}{%
\begin{tabular}{c|cccccc}
\toprule
Constraints Level & \multicolumn{6}{c}{0.2} \\
Method & COPKM\_P & KSKM & KSKM\_E & BLPKMCC & PCCC & COPKM \\
Dataset &  &  &  &  &  &  \\
\midrule
synthetic\_n1000\_d2\_k10 & 1.00 & 1.00 & 1.00 & 1.00 & 1.00 & 1.00 \\
synthetic\_n1000\_d2\_k100 & 0.32 & 0.33 & 0.33 & 0.34 & 0.33 & 0.32 \\
synthetic\_n1000\_d2\_k2 & 1.00 & 1.00 & 1.00 & 1.00 & 1.00 & 1.00 \\
synthetic\_n1000\_d2\_k20 & 0.96 & 0.96 & 0.96 & 0.96 & 0.96 & 0.96 \\
synthetic\_n1000\_d2\_k5 & 1.00 & 1.00 & 1.00 & 1.00 & 1.00 & 1.00 \\
synthetic\_n1000\_d2\_k50 & 0.65 & 0.73 & 0.73 & 0.65 & 0.69 & 0.58 \\
synthetic\_n2000\_d2\_k10 & 1.00 & 1.00 & 1.00 & 1.00 & 1.00 & 1.00 \\
synthetic\_n2000\_d2\_k100 & 0.46 & 0.53 & 0.52 & 0.51 & 0.51 & 0.44 \\
synthetic\_n2000\_d2\_k2 & 1.00 & 1.00 & 1.00 & 1.00 & 1.00 & 1.00 \\
synthetic\_n2000\_d2\_k20 & 1.00 & 1.00 & 1.00 & 1.00 & 1.00 & 1.00 \\
synthetic\_n2000\_d2\_k5 & 1.00 & 1.00 & 1.00 & 1.00 & 1.00 & 1.00 \\
synthetic\_n2000\_d2\_k50 & 0.89 & 0.89 & 0.89 & 0.89 & 0.89 & - \\
synthetic\_n5000\_d2\_k10 & 1.00 & 1.00 & 1.00 & 1.00 & 1.00 & 1.00 \\
synthetic\_n5000\_d2\_k100 & 0.88 & 0.88 & 0.88 & 0.88 & 0.88 & - \\
synthetic\_n5000\_d2\_k2 & 1.00 & 1.00 & 1.00 & 1.00 & 1.00 & 1.00 \\
synthetic\_n5000\_d2\_k20 & 1.00 & 1.00 & 1.00 & 1.00 & 1.00 & 1.00 \\
synthetic\_n5000\_d2\_k5 & 1.00 & 1.00 & 1.00 & 1.00 & 1.00 & 1.00 \\
synthetic\_n5000\_d2\_k50 & 0.99 & 0.99 & 0.99 & 0.99 & 0.99 & 0.99 \\
synthetic\_n500\_d2\_k10 & 0.98 & 0.98 & 0.98 & 0.98 & 0.99 & 0.98 \\
synthetic\_n500\_d2\_k100 & 0.26 & 0.26 & 0.27 & 0.27 & 0.26 & 0.26 \\
synthetic\_n500\_d2\_k2 & 1.00 & 1.00 & 1.00 & 1.00 & 1.00 & 1.00 \\
synthetic\_n500\_d2\_k20 & 0.83 & 0.84 & 0.85 & 0.84 & 0.85 & - \\
synthetic\_n500\_d2\_k5 & 1.00 & 1.00 & 1.00 & 1.00 & 1.00 & 1.00 \\
synthetic\_n500\_d2\_k50 & 0.45 & 0.51 & 0.51 & 0.47 & 0.49 & 0.45 \\
\bottomrule
\end{tabular}
}
\end{table}

\begin{table}[H]
\caption{Algorithm success rate on dataset collection 3 with 20.0\% constraints}
\label{tab:rate_success_COL3_20}
\centering
\resizebox{0.9\textwidth}{!}{%
\begin{tabular}{c|cccccc}
\toprule
Constraints Level & \multicolumn{6}{c}{0.2} \\
Method & COPKM\_P & KSKM & KSKM\_E & BLPKMCC & PCCC & COPKM \\
Dataset &  &  &  &  &  &  \\
\midrule
synthetic\_n1000\_d2\_k10 & 1.00 & 1.00 & 1.00 & 1.00 & 1.00 & 1.00 \\
synthetic\_n1000\_d2\_k100 & 1.00 & 1.00 & 1.00 & 1.00 & 1.00 & 1.00 \\
synthetic\_n1000\_d2\_k2 & 1.00 & 1.00 & 1.00 & 1.00 & 1.00 & 1.00 \\
synthetic\_n1000\_d2\_k20 & 1.00 & 1.00 & 1.00 & 1.00 & 1.00 & 0.60 \\
synthetic\_n1000\_d2\_k5 & 1.00 & 1.00 & 1.00 & 1.00 & 1.00 & 1.00 \\
synthetic\_n1000\_d2\_k50 & 1.00 & 1.00 & 1.00 & 1.00 & 1.00 & 1.00 \\
synthetic\_n2000\_d2\_k10 & 1.00 & 1.00 & 1.00 & 1.00 & 1.00 & 1.00 \\
synthetic\_n2000\_d2\_k100 & 1.00 & 1.00 & 1.00 & 1.00 & 1.00 & 1.00 \\
synthetic\_n2000\_d2\_k2 & 1.00 & 1.00 & 1.00 & 1.00 & 1.00 & 1.00 \\
synthetic\_n2000\_d2\_k20 & 1.00 & 1.00 & 1.00 & 1.00 & 1.00 & 1.00 \\
synthetic\_n2000\_d2\_k5 & 1.00 & 1.00 & 1.00 & 1.00 & 1.00 & 1.00 \\
synthetic\_n2000\_d2\_k50 & 1.00 & 1.00 & 1.00 & 1.00 & 1.00 & 0.00 \\
synthetic\_n5000\_d2\_k10 & 1.00 & 1.00 & 1.00 & 1.00 & 1.00 & 1.00 \\
synthetic\_n5000\_d2\_k100 & 1.00 & 1.00 & 1.00 & 0.90 & 1.00 & 0.00 \\
synthetic\_n5000\_d2\_k2 & 1.00 & 1.00 & 1.00 & 1.00 & 1.00 & 1.00 \\
synthetic\_n5000\_d2\_k20 & 1.00 & 1.00 & 1.00 & 1.00 & 1.00 & 1.00 \\
synthetic\_n5000\_d2\_k5 & 1.00 & 1.00 & 1.00 & 1.00 & 1.00 & 1.00 \\
synthetic\_n5000\_d2\_k50 & 1.00 & 1.00 & 1.00 & 1.00 & 1.00 & 1.00 \\
synthetic\_n500\_d2\_k10 & 1.00 & 1.00 & 1.00 & 1.00 & 1.00 & 1.00 \\
synthetic\_n500\_d2\_k100 & 1.00 & 1.00 & 1.00 & 1.00 & 1.00 & 1.00 \\
synthetic\_n500\_d2\_k2 & 1.00 & 1.00 & 1.00 & 1.00 & 1.00 & 1.00 \\
synthetic\_n500\_d2\_k20 & 1.00 & 1.00 & 1.00 & 1.00 & 1.00 & 0.00 \\
synthetic\_n500\_d2\_k5 & 1.00 & 1.00 & 1.00 & 1.00 & 1.00 & 1.00 \\
synthetic\_n500\_d2\_k50 & 1.00 & 1.00 & 1.00 & 1.00 & 1.00 & 1.00 \\
\bottomrule
\end{tabular}
}
\end{table}

\begin{table}[H]
\caption{Clustering inertia on dataset collection 4 with 0.5\% constraints}
\label{tab:obj_COL4_0}
\centering
\resizebox{0.9\textwidth}{!}{%
\begin{tabular}{c|cccccc}
\toprule
Constraints Level & \multicolumn{6}{c}{0.005} \\
Method & COPKM\_P & KSKM & KSKM\_E & BLPKMCC & PCCC & COPKM \\
Dataset &  &  &  &  &  &  \\
\midrule
banana & 6519.10 & 6482.97 & 6482.97 & 6482.98 & 6486.67 & - \\
cifar10 & 127530232.16 & 126770854.91 & 126492578.88 & 126519249.09 & 126612825.71 & 127579490.59 \\
cifar100 & 90808595.10 & 90713888.95 & 90701179.10 & 91217697.77 & 91012502.35 & 90817932.44 \\
letter & 124107.98 & 123150.01 & 123194.49 & 124061.70 & 123658.15 & 124320.11 \\
mnist & 43436554.98 & 43212761.15 & 43179496.08 & 43184194.94 & 43263288.23 & 43437504.01 \\
shuttle & 296049.18 & 293608.03 & 293144.76 & 305160.20 & 299481.92 & 296829.94 \\
\bottomrule
\end{tabular}
}
\end{table}

\begin{table}[H]
\caption{CPU time on dataset collection 4 with 0.5\% constraints}
\label{tab:time_COL4_0}
\centering
\resizebox{0.9\textwidth}{!}{%
\begin{tabular}{c|cccccc}
\toprule
Constraints Level & \multicolumn{6}{c}{0.005} \\
Method & COPKM\_P & KSKM & KSKM\_E & BLPKMCC & PCCC & COPKM \\
Dataset &  &  &  &  &  &  \\
\midrule
banana & 0.37 & 0.10 & 0.11 & 1.94 & 4.73 & - \\
cifar10 & 55.18 & 322.69 & 645.73 & 1881.88 & 1224.33 & 16.07 \\
cifar100 & 620.46 & 1359.85 & 3128.66 & 3619.73 & 3638.03 & 166.84 \\
letter & 2.73 & 30.17 & 116.65 & 571.06 & 749.37 & 6.77 \\
mnist & 109.58 & 136.71 & 434.18 & 1805.80 & 804.60 & 12.76 \\
shuttle & 4.96 & 2.29 & 35.71 & 198.64 & 236.72 & 2.22 \\
\bottomrule
\end{tabular}
}
\end{table}

\begin{table}[H]
\caption{ARI on dataset collection 4 with 0.5\% constraints}
\label{tab:ARI_COL4_0}
\centering
\resizebox{0.9\textwidth}{!}{%
\begin{tabular}{c|cccccc}
\toprule
Constraints Level & \multicolumn{6}{c}{0.005} \\
Method & COPKM\_P & KSKM & KSKM\_E & BLPKMCC & PCCC & COPKM \\
Dataset &  &  &  &  &  &  \\
\midrule
banana & 0.01 & 0.01 & 0.01 & 0.01 & 0.02 & - \\
cifar10 & 0.04 & 0.04 & 0.04 & 0.04 & 0.04 & 0.04 \\
cifar100 & 0.02 & 0.02 & 0.02 & 0.02 & 0.02 & 0.02 \\
letter & 0.15 & 0.15 & 0.15 & 0.15 & 0.15 & 0.15 \\
mnist & 0.34 & 0.38 & 0.38 & 0.38 & 0.37 & 0.34 \\
shuttle & 0.61 & 0.61 & 0.62 & 0.46 & 0.62 & 0.60 \\
\bottomrule
\end{tabular}
}
\end{table}

\begin{table}[H]
\caption{Algorithm success rate on dataset collection 4 with 0.5\% constraints}
\label{tab:rate_success_COL4_0}
\centering
\resizebox{0.9\textwidth}{!}{%
\begin{tabular}{c|cccccc}
\toprule
Constraints Level & \multicolumn{6}{c}{0.005} \\
Method & COPKM\_P & KSKM & KSKM\_E & BLPKMCC & PCCC & COPKM \\
Dataset &  &  &  &  &  &  \\
\midrule
banana & 1.00 & 1.00 & 1.00 & 1.00 & 1.00 & 0.00 \\
cifar10 & 1.00 & 1.00 & 1.00 & 1.00 & 1.00 & 1.00 \\
cifar100 & 1.00 & 1.00 & 1.00 & 1.00 & 1.00 & 1.00 \\
letter & 1.00 & 1.00 & 1.00 & 1.00 & 1.00 & 1.00 \\
mnist & 1.00 & 1.00 & 1.00 & 1.00 & 1.00 & 1.00 \\
shuttle & 1.00 & 1.00 & 1.00 & 1.00 & 1.00 & 1.00 \\
\bottomrule
\end{tabular}
}
\end{table}

\begin{table}[H]
\caption{Clustering inertia on dataset collection 4 with 1.0\% constraints}
\label{tab:obj_COL4_1}
\centering
\resizebox{0.9\textwidth}{!}{%
\begin{tabular}{c|cccccc}
\toprule
Constraints Level & \multicolumn{6}{c}{0.01} \\
Method & COPKM\_P & KSKM & KSKM\_E & BLPKMCC & PCCC & COPKM \\
Dataset &  &  &  &  &  &  \\
\midrule
banana & 7707.22 & 7566.29 & 7566.29 & 7566.29 & 7567.65 & - \\
cifar10 & 147220022.93 & 147887573.71 & 147457869.95 & 160655672.87 & 157005767.08 & - \\
cifar100 & 93006014.31 & 92748064.89 & 92752048.21 & 94537010.08 & 93640143.35 & 92934054.98 \\
letter & 130686.28 & 129305.07 & 128946.60 & 129625.72 & 129235.82 & 130623.89 \\
mnist & 44631657.08 & 46121516.68 & 44644860.85 & 48776943.77 & 48542487.99 & - \\
shuttle & 330841.59 & 328046.93 & 327845.18 & 338035.86 & 331753.88 & 332760.92 \\
\bottomrule
\end{tabular}
}
\end{table}

\begin{table}[H]
\caption{CPU time on dataset collection 4 with 1.0\% constraints}
\label{tab:time_COL4_1}
\centering
\resizebox{0.9\textwidth}{!}{%
\begin{tabular}{c|cccccc}
\toprule
Constraints Level & \multicolumn{6}{c}{0.01} \\
Method & COPKM\_P & KSKM & KSKM\_E & BLPKMCC & PCCC & COPKM \\
Dataset &  &  &  &  &  &  \\
\midrule
banana & 0.18 & 0.20 & 0.31 & 2.50 & 5.22 & - \\
cifar10 & 26.65 & 118.05 & 3602.63 & 3624.14 & 3633.96 & - \\
cifar100 & 754.80 & 1266.71 & 3705.35 & 3949.69 & 3851.99 & 154.89 \\
letter & 12.79 & 17.44 & 262.23 & 1002.48 & 695.34 & 5.47 \\
mnist & 60.77 & 60.67 & 1357.07 & 3627.71 & 3790.51 & - \\
shuttle & 2.95 & 0.68 & 22.89 & 278.99 & 78.96 & 0.99 \\
\bottomrule
\end{tabular}
}
\end{table}

\begin{table}[H]
\caption{ARI on dataset collection 4 with 1.0\% constraints}
\label{tab:ARI_COL4_1}
\centering
\resizebox{0.9\textwidth}{!}{%
\begin{tabular}{c|cccccc}
\toprule
Constraints Level & \multicolumn{6}{c}{0.01} \\
Method & COPKM\_P & KSKM & KSKM\_E & BLPKMCC & PCCC & COPKM \\
Dataset &  &  &  &  &  &  \\
\midrule
banana & 0.01 & 0.01 & 0.01 & 0.01 & 0.02 & - \\
cifar10 & 0.05 & 0.04 & 0.04 & 0.02 & 0.03 & - \\
cifar100 & 0.02 & 0.02 & 0.02 & 0.02 & 0.02 & 0.02 \\
letter & 0.15 & 0.16 & 0.16 & 0.16 & 0.16 & 0.15 \\
mnist & 0.63 & 0.33 & 0.65 & 0.06 & 0.09 & - \\
shuttle & 0.95 & 0.94 & 0.95 & 0.93 & 0.94 & 0.95 \\
\bottomrule
\end{tabular}
}
\end{table}

\begin{table}[H]
\caption{Algorithm success rate on dataset collection 4 with 1.0\% constraints}
\label{tab:rate_success_COL4_1}
\centering
\resizebox{0.9\textwidth}{!}{%
\begin{tabular}{c|cccccc}
\toprule
Constraints Level & \multicolumn{6}{c}{0.01} \\
Method & COPKM\_P & KSKM & KSKM\_E & BLPKMCC & PCCC & COPKM \\
Dataset &  &  &  &  &  &  \\
\midrule
banana & 1.00 & 1.00 & 1.00 & 1.00 & 1.00 & 0.00 \\
cifar10 & 1.00 & 1.00 & 1.00 & 1.00 & 1.00 & 0.00 \\
cifar100 & 1.00 & 1.00 & 1.00 & 0.50 & 0.50 & 1.00 \\
letter & 1.00 & 1.00 & 1.00 & 1.00 & 1.00 & 1.00 \\
mnist & 1.00 & 1.00 & 1.00 & 0.20 & 1.00 & 0.00 \\
shuttle & 1.00 & 1.00 & 1.00 & 1.00 & 1.00 & 1.00 \\
\bottomrule
\end{tabular}
}
\end{table}

\begin{table}[H]
\caption{Clustering inertia on dataset collection 4 with 5.0\% constraints}
\label{tab:obj_COL4_5}
\centering
\resizebox{0.9\textwidth}{!}{%
\begin{tabular}{c|cccccc}
\toprule
Constraints Level & \multicolumn{6}{c}{0.05} \\
Method & COPKM\_P & KSKM & KSKM\_E & BLPKMCC & PCCC & COPKM \\
Dataset &  &  &  &  &  &  \\
\midrule
banana & 10556.11 & 10556.11 & 10556.11 & 10556.11 & 10556.11 & 10556.11 \\
cifar10 & 172608323.88 & 172608323.88 & 172608323.88 & 172608323.88 & 172608323.88 & 172608323.88 \\
cifar100 & 147181909.52 & 147067151.28 & 147074949.64 & - & - & - \\
letter & 208187.97 & 208141.24 & 208124.29 & 208124.32 & 208232.38 & - \\
mnist & 44514740.31 & 44514740.31 & 44514740.31 & 44514740.31 & 44514740.31 & 44514740.31 \\
shuttle & 367447.69 & 366680.49 & 366680.49 & 366760.67 & 367244.67 & 367716.94 \\
\bottomrule
\end{tabular}
}
\end{table}

\begin{table}[H]
\caption{CPU time on dataset collection 4 with 5.0\% constraints}
\label{tab:time_COL4_5}
\centering
\resizebox{0.9\textwidth}{!}{%
\begin{tabular}{c|cccccc}
\toprule
Constraints Level & \multicolumn{6}{c}{0.05} \\
Method & COPKM\_P & KSKM & KSKM\_E & BLPKMCC & PCCC & COPKM \\
Dataset &  &  &  &  &  &  \\
\midrule
banana & 0.03 & 0.01 & 0.01 & 1.55 & 0.08 & 0.03 \\
cifar10 & 2.63 & 2.95 & 3.30 & 1135.66 & 8.61 & 2.68 \\
cifar100 & 28.41 & 329.19 & 3696.78 & - & - & - \\
letter & 1.67 & 2.64 & 944.50 & 3376.45 & 1154.58 & - \\
mnist & 2.02 & 1.36 & 1.40 & 1572.45 & 9.13 & 2.07 \\
shuttle & 1.33 & 0.84 & 0.92 & 2746.80 & 10.37 & 1.37 \\
\bottomrule
\end{tabular}
}
\end{table}

\begin{table}[H]
\caption{ARI on dataset collection 4 with 5.0\% constraints}
\label{tab:ARI_COL4_5}
\centering
\resizebox{0.9\textwidth}{!}{%
\begin{tabular}{c|cccccc}
\toprule
Constraints Level & \multicolumn{6}{c}{0.05} \\
Method & COPKM\_P & KSKM & KSKM\_E & BLPKMCC & PCCC & COPKM \\
Dataset &  &  &  &  &  &  \\
\midrule
banana & 1.00 & 1.00 & 1.00 & 1.00 & 1.00 & 1.00 \\
cifar10 & 1.00 & 1.00 & 1.00 & 1.00 & 1.00 & 1.00 \\
cifar100 & 0.51 & 0.50 & 0.50 & - & - & - \\
letter & 0.86 & 0.86 & 0.86 & 0.86 & 0.86 & - \\
mnist & 1.00 & 1.00 & 1.00 & 1.00 & 1.00 & 1.00 \\
shuttle & 1.00 & 1.00 & 1.00 & 1.00 & 1.00 & 1.00 \\
\bottomrule
\end{tabular}
}
\end{table}

\begin{table}[H]
\caption{Algorithm success rate on dataset collection 4 with 5.0\% constraints}
\label{tab:rate_success_COL4_5}
\centering
\resizebox{0.9\textwidth}{!}{%
\begin{tabular}{c|cccccc}
\toprule
Constraints Level & \multicolumn{6}{c}{0.05} \\
Method & COPKM\_P & KSKM & KSKM\_E & BLPKMCC & PCCC & COPKM \\
Dataset &  &  &  &  &  &  \\
\midrule
banana & 1.00 & 1.00 & 1.00 & 1.00 & 1.00 & 1.00 \\
cifar10 & 1.00 & 1.00 & 1.00 & 1.00 & 1.00 & 1.00 \\
cifar100 & 1.00 & 1.00 & 1.00 & 0.00 & 0.00 & 0.00 \\
letter & 1.00 & 1.00 & 1.00 & 1.00 & 1.00 & 0.00 \\
mnist & 1.00 & 1.00 & 1.00 & 1.00 & 1.00 & 1.00 \\
shuttle & 1.00 & 1.00 & 1.00 & 1.00 & 1.00 & 1.00 \\
\bottomrule
\end{tabular}
}
\end{table}

\subsubsection{Results: No Mutation}
\input{tables/COL1/COL1_0.05_obj_no_mutation}
\input{tables/COL1/COL1_0.05_time_no_mutation}
\input{tables/COL1/COL1_0.05_ARI_no_mutation}
\input{tables/COL1/COL1_0.05_rate_success_no_mutation}
\input{tables/COL1/COL1_0.1_obj_no_mutation}
\input{tables/COL1/COL1_0.1_time_no_mutation}
\input{tables/COL1/COL1_0.1_ARI_no_mutation}
\input{tables/COL1/COL1_0.1_rate_success_no_mutation}
\input{tables/COL1/COL1_0.15_obj_no_mutation}
\input{tables/COL1/COL1_0.15_time_no_mutation}
\input{tables/COL1/COL1_0.15_ARI_no_mutation}
\input{tables/COL1/COL1_0.15_rate_success_no_mutation}
\input{tables/COL1/COL1_0.2_obj_no_mutation}
\input{tables/COL1/COL1_0.2_time_no_mutation}
\input{tables/COL1/COL1_0.2_ARI_no_mutation}
\input{tables/COL1/COL1_0.2_rate_success_no_mutation}
\input{tables/COL2/COL2_0.05_obj_no_mutation}
\input{tables/COL2/COL2_0.05_time_no_mutation}
\input{tables/COL2/COL2_0.05_ARI_no_mutation}
\input{tables/COL2/COL2_0.05_rate_success_no_mutation}
\input{tables/COL2/COL2_0.1_obj_no_mutation}
\input{tables/COL2/COL2_0.1_time_no_mutation}
\input{tables/COL2/COL2_0.1_ARI_no_mutation}
\input{tables/COL2/COL2_0.1_rate_success_no_mutation}
\input{tables/COL2/COL2_0.15_obj_no_mutation}
\input{tables/COL2/COL2_0.15_time_no_mutation}
\input{tables/COL2/COL2_0.15_ARI_no_mutation}
\input{tables/COL2/COL2_0.15_rate_success_no_mutation}
\input{tables/COL2/COL2_0.2_obj_no_mutation}
\input{tables/COL2/COL2_0.2_time_no_mutation}
\input{tables/COL2/COL2_0.2_ARI_no_mutation}
\input{tables/COL2/COL2_0.2_rate_success_no_mutation}
\input{tables/COL2/COL2_0.25_obj_no_mutation}
\input{tables/COL2/COL2_0.25_time_no_mutation}
\input{tables/COL2/COL2_0.25_ARI_no_mutation}
\input{tables/COL2/COL2_0.25_rate_success_no_mutation}
\input{tables/COL2/COL2_0.3_obj_no_mutation}
\input{tables/COL2/COL2_0.3_time_no_mutation}
\input{tables/COL2/COL2_0.3_ARI_no_mutation}
\input{tables/COL2/COL2_0.3_rate_success_no_mutation}
\input{tables/COL2/COL2_0.35_obj_no_mutation}
\input{tables/COL2/COL2_0.35_time_no_mutation}
\input{tables/COL2/COL2_0.35_ARI_no_mutation}
\input{tables/COL2/COL2_0.35_rate_success_no_mutation}
\input{tables/COL2/COL2_0.4_obj_no_mutation}
\input{tables/COL2/COL2_0.4_time_no_mutation}
\input{tables/COL2/COL2_0.4_ARI_no_mutation}
\input{tables/COL2/COL2_0.4_rate_success_no_mutation}
\input{tables/COL2/COL2_0.45_obj_no_mutation}
\input{tables/COL2/COL2_0.45_time_no_mutation}
\input{tables/COL2/COL2_0.45_ARI_no_mutation}
\input{tables/COL2/COL2_0.45_rate_success_no_mutation}
\input{tables/COL2/COL2_0.5_obj_no_mutation}
\input{tables/COL2/COL2_0.5_time_no_mutation}
\input{tables/COL2/COL2_0.5_ARI_no_mutation}
\input{tables/COL2/COL2_0.5_rate_success_no_mutation}
\begin{table}[H]
\caption{Clustering inertia on dataset collection 3 with 5.0\% constraints (without centroid mutation)}
\label{tab:obj_COL3_5_no_mutation}
\centering
\resizebox{0.66\textwidth}{!}{%
\begin{tabular}{c|ccc}
\toprule
Constraints Level & \multicolumn{3}{c}{0.05} \\
Method & KSKM & KSKM\_E & PCCC \\
Dataset &  &  &  \\
\midrule
synthetic\_n1000\_d2\_k10 & 92.69 & 87.75 & 85.40 \\
synthetic\_n1000\_d2\_k100 & 16.58 & 16.44 & 16.43 \\
synthetic\_n1000\_d2\_k2 & 911.71 & 911.71 & 912.38 \\
synthetic\_n1000\_d2\_k20 & 68.48 & 67.89 & 66.66 \\
synthetic\_n1000\_d2\_k5 & 69.54 & 69.06 & 66.18 \\
synthetic\_n1000\_d2\_k50 & 33.12 & 32.69 & 31.92 \\
synthetic\_n2000\_d2\_k10 & 150.03 & 144.59 & 144.32 \\
synthetic\_n2000\_d2\_k100 & 41.06 & 40.66 & 40.50 \\
synthetic\_n2000\_d2\_k2 & 1770.35 & 1770.35 & 1770.35 \\
synthetic\_n2000\_d2\_k20 & 147.65 & 143.08 & 140.76 \\
synthetic\_n2000\_d2\_k5 & 136.15 & 136.15 & 136.15 \\
synthetic\_n2000\_d2\_k50 & 71.88 & 70.86 & 70.80 \\
synthetic\_n5000\_d2\_k10 & 334.67 & 333.78 & 332.23 \\
synthetic\_n5000\_d2\_k100 & 120.02 & 118.01 & 117.50 \\
synthetic\_n5000\_d2\_k2 & 4493.84 & 4493.84 & 4493.84 \\
synthetic\_n5000\_d2\_k20 & 503.48 & 404.80 & 430.70 \\
synthetic\_n5000\_d2\_k5 & 352.58 & 352.58 & 352.58 \\
synthetic\_n5000\_d2\_k50 & 209.26 & 203.97 & 203.41 \\
synthetic\_n500\_d2\_k10 & 38.86 & 38.75 & 37.11 \\
synthetic\_n500\_d2\_k100 & 7.35 & 7.34 & 7.32 \\
synthetic\_n500\_d2\_k2 & 449.71 & 449.71 & 449.71 \\
synthetic\_n500\_d2\_k20 & 32.00 & 31.81 & 31.99 \\
synthetic\_n500\_d2\_k5 & 40.05 & 40.05 & 40.64 \\
synthetic\_n500\_d2\_k50 & 14.12 & 14.06 & 13.83 \\
\bottomrule
\end{tabular}
}
\end{table}

\begin{table}[H]
\caption{CPU time on dataset collection 3 with 5.0\% constraints (without centroid mutation)}
\label{tab:time_COL3_5_no_mutation}
\centering
\resizebox{0.66\textwidth}{!}{%
\begin{tabular}{c|ccc}
\toprule
Constraints Level & \multicolumn{3}{c}{0.05} \\
Method & KSKM & KSKM\_E & PCCC \\
Dataset &  &  &  \\
\midrule
synthetic\_n1000\_d2\_k10 & 0.01 & 0.16 & 0.53 \\
synthetic\_n1000\_d2\_k100 & 0.02 & 2.12 & 10.64 \\
synthetic\_n1000\_d2\_k2 & 0.00 & 0.01 & 0.02 \\
synthetic\_n1000\_d2\_k20 & 0.01 & 0.34 & 2.81 \\
synthetic\_n1000\_d2\_k5 & 0.01 & 0.04 & 0.29 \\
synthetic\_n1000\_d2\_k50 & 0.02 & 1.04 & 7.40 \\
synthetic\_n2000\_d2\_k10 & 0.02 & 0.36 & 2.34 \\
synthetic\_n2000\_d2\_k100 & 0.07 & 7.42 & 76.81 \\
synthetic\_n2000\_d2\_k2 & 0.00 & 0.01 & 0.03 \\
synthetic\_n2000\_d2\_k20 & 0.03 & 1.64 & 14.73 \\
synthetic\_n2000\_d2\_k5 & 0.01 & 0.09 & 0.75 \\
synthetic\_n2000\_d2\_k50 & 0.05 & 5.07 & 42.97 \\
synthetic\_n5000\_d2\_k10 & 0.02 & 0.92 & 8.74 \\
synthetic\_n5000\_d2\_k100 & 0.34 & 48.25 & 667.32 \\
synthetic\_n5000\_d2\_k2 & 0.01 & 0.01 & 0.03 \\
synthetic\_n5000\_d2\_k20 & 0.10 & 19.16 & 150.32 \\
synthetic\_n5000\_d2\_k5 & 0.01 & 0.09 & 0.38 \\
synthetic\_n5000\_d2\_k50 & 0.23 & 29.99 & 277.19 \\
synthetic\_n500\_d2\_k10 & 0.00 & 0.04 & 0.22 \\
synthetic\_n500\_d2\_k100 & 0.01 & 0.38 & 2.06 \\
synthetic\_n500\_d2\_k2 & 0.00 & 0.00 & 0.03 \\
synthetic\_n500\_d2\_k20 & 0.01 & 0.10 & 0.95 \\
synthetic\_n500\_d2\_k5 & 0.00 & 0.02 & 0.13 \\
synthetic\_n500\_d2\_k50 & 0.01 & 0.20 & 1.32 \\
\bottomrule
\end{tabular}
}
\end{table}

\begin{table}[H]
\caption{ARI on dataset collection 3 with 5.0\% constraints (without centroid mutation)}
\label{tab:ARI_COL3_5_no_mutation}
\centering
\resizebox{0.66\textwidth}{!}{%
\begin{tabular}{c|ccc}
\toprule
Constraints Level & \multicolumn{3}{c}{0.05} \\
Method & KSKM & KSKM\_E & PCCC \\
Dataset &  &  &  \\
\midrule
synthetic\_n1000\_d2\_k10 & 0.86 & 0.88 & 0.88 \\
synthetic\_n1000\_d2\_k100 & 0.24 & 0.24 & 0.24 \\
synthetic\_n1000\_d2\_k2 & 1.00 & 1.00 & 1.00 \\
synthetic\_n1000\_d2\_k20 & 0.59 & 0.59 & 0.60 \\
synthetic\_n1000\_d2\_k5 & 0.93 & 0.93 & 0.93 \\
synthetic\_n1000\_d2\_k50 & 0.41 & 0.41 & 0.42 \\
synthetic\_n2000\_d2\_k10 & 0.92 & 0.93 & 0.93 \\
synthetic\_n2000\_d2\_k100 & 0.25 & 0.25 & 0.25 \\
synthetic\_n2000\_d2\_k2 & 1.00 & 1.00 & 1.00 \\
synthetic\_n2000\_d2\_k20 & 0.61 & 0.63 & 0.65 \\
synthetic\_n2000\_d2\_k5 & 0.94 & 0.94 & 0.94 \\
synthetic\_n2000\_d2\_k50 & 0.43 & 0.43 & 0.44 \\
synthetic\_n5000\_d2\_k10 & 0.98 & 0.98 & 0.98 \\
synthetic\_n5000\_d2\_k100 & 0.26 & 0.26 & 0.26 \\
synthetic\_n5000\_d2\_k2 & 1.00 & 1.00 & 1.00 \\
synthetic\_n5000\_d2\_k20 & 0.67 & 0.76 & 0.72 \\
synthetic\_n5000\_d2\_k5 & 1.00 & 1.00 & 1.00 \\
synthetic\_n5000\_d2\_k50 & 0.47 & 0.48 & 0.48 \\
synthetic\_n500\_d2\_k10 & 0.86 & 0.86 & 0.87 \\
synthetic\_n500\_d2\_k100 & 0.22 & 0.22 & 0.22 \\
synthetic\_n500\_d2\_k2 & 1.00 & 1.00 & 0.99 \\
synthetic\_n500\_d2\_k20 & 0.56 & 0.57 & 0.57 \\
synthetic\_n500\_d2\_k5 & 0.89 & 0.89 & 0.89 \\
synthetic\_n500\_d2\_k50 & 0.38 & 0.38 & 0.39 \\
\bottomrule
\end{tabular}
}
\end{table}

\begin{table}[H]
\caption{Algorithm success rate on dataset collection 3 with 5.0\% constraints (without centroid mutation)}
\label{tab:rate_success_COL3_5_no_mutation}
\centering
\resizebox{0.66\textwidth}{!}{%
\begin{tabular}{c|ccc}
\toprule
Constraints Level & \multicolumn{3}{c}{0.05} \\
Method & KSKM & KSKM\_E & PCCC \\
Dataset &  &  &  \\
\midrule
synthetic\_n1000\_d2\_k10 & 1.00 & 1.00 & 1.00 \\
synthetic\_n1000\_d2\_k100 & 1.00 & 1.00 & 1.00 \\
synthetic\_n1000\_d2\_k2 & 1.00 & 1.00 & 1.00 \\
synthetic\_n1000\_d2\_k20 & 1.00 & 1.00 & 1.00 \\
synthetic\_n1000\_d2\_k5 & 1.00 & 1.00 & 1.00 \\
synthetic\_n1000\_d2\_k50 & 1.00 & 1.00 & 1.00 \\
synthetic\_n2000\_d2\_k10 & 1.00 & 1.00 & 1.00 \\
synthetic\_n2000\_d2\_k100 & 1.00 & 1.00 & 1.00 \\
synthetic\_n2000\_d2\_k2 & 1.00 & 1.00 & 1.00 \\
synthetic\_n2000\_d2\_k20 & 1.00 & 1.00 & 1.00 \\
synthetic\_n2000\_d2\_k5 & 1.00 & 1.00 & 1.00 \\
synthetic\_n2000\_d2\_k50 & 1.00 & 1.00 & 1.00 \\
synthetic\_n5000\_d2\_k10 & 1.00 & 1.00 & 1.00 \\
synthetic\_n5000\_d2\_k100 & 1.00 & 1.00 & 1.00 \\
synthetic\_n5000\_d2\_k2 & 1.00 & 1.00 & 1.00 \\
synthetic\_n5000\_d2\_k20 & 1.00 & 1.00 & 1.00 \\
synthetic\_n5000\_d2\_k5 & 1.00 & 1.00 & 1.00 \\
synthetic\_n5000\_d2\_k50 & 1.00 & 1.00 & 1.00 \\
synthetic\_n500\_d2\_k10 & 1.00 & 1.00 & 1.00 \\
synthetic\_n500\_d2\_k100 & 1.00 & 1.00 & 1.00 \\
synthetic\_n500\_d2\_k2 & 1.00 & 1.00 & 1.00 \\
synthetic\_n500\_d2\_k20 & 1.00 & 1.00 & 1.00 \\
synthetic\_n500\_d2\_k5 & 1.00 & 1.00 & 1.00 \\
synthetic\_n500\_d2\_k50 & 1.00 & 1.00 & 1.00 \\
\bottomrule
\end{tabular}
}
\end{table}

\begin{table}[H]
\caption{Clustering inertia on dataset collection 3 with 10.0\% constraints (without centroid mutation)}
\label{tab:obj_COL3_10_no_mutation}
\centering
\resizebox{0.66\textwidth}{!}{%
\begin{tabular}{c|ccc}
\toprule
Constraints Level & \multicolumn{3}{c}{0.1} \\
Method & KSKM & KSKM\_E & PCCC \\
Dataset &  &  &  \\
\midrule
synthetic\_n1000\_d2\_k10 & 106.50 & 66.18 & 66.19 \\
synthetic\_n1000\_d2\_k100 & 19.79 & 19.72 & 19.55 \\
synthetic\_n1000\_d2\_k2 & 911.71 & 911.71 & 911.71 \\
synthetic\_n1000\_d2\_k20 & 85.91 & 78.50 & 76.81 \\
synthetic\_n1000\_d2\_k5 & 69.11 & 69.11 & 69.11 \\
synthetic\_n1000\_d2\_k50 & 37.28 & 36.41 & 36.26 \\
synthetic\_n2000\_d2\_k10 & 130.68 & 130.68 & 130.72 \\
synthetic\_n2000\_d2\_k100 & 51.59 & 50.34 & 50.17 \\
synthetic\_n2000\_d2\_k2 & 1770.35 & 1770.35 & 1770.35 \\
synthetic\_n2000\_d2\_k20 & 176.09 & 169.87 & 154.26 \\
synthetic\_n2000\_d2\_k5 & 140.66 & 140.66 & 140.67 \\
synthetic\_n2000\_d2\_k50 & 93.29 & 88.43 & 87.94 \\
synthetic\_n5000\_d2\_k10 & 336.12 & 336.12 & 336.12 \\
synthetic\_n5000\_d2\_k100 & 203.34 & 187.79 & 186.98 \\
synthetic\_n5000\_d2\_k2 & 4493.84 & 4493.84 & 4493.84 \\
synthetic\_n5000\_d2\_k20 & 410.96 & 410.96 & 410.99 \\
synthetic\_n5000\_d2\_k5 & 353.15 & 353.15 & 353.15 \\
synthetic\_n5000\_d2\_k50 & 438.88 & 381.39 & 330.17 \\
synthetic\_n500\_d2\_k10 & 42.54 & 41.93 & 46.43 \\
synthetic\_n500\_d2\_k100 & 7.59 & 7.56 & 7.56 \\
synthetic\_n500\_d2\_k2 & 449.70 & 449.70 & 449.70 \\
synthetic\_n500\_d2\_k20 & 36.68 & 35.96 & 34.17 \\
synthetic\_n500\_d2\_k5 & 33.97 & 33.97 & 33.99 \\
synthetic\_n500\_d2\_k50 & 15.97 & 15.61 & 15.54 \\
\bottomrule
\end{tabular}
}
\end{table}

\begin{table}[H]
\caption{CPU time on dataset collection 3 with 10.0\% constraints (without centroid mutation)}
\label{tab:time_COL3_10_no_mutation}
\centering
\resizebox{0.66\textwidth}{!}{%
\begin{tabular}{c|ccc}
\toprule
Constraints Level & \multicolumn{3}{c}{0.1} \\
Method & KSKM & KSKM\_E & PCCC \\
Dataset &  &  &  \\
\midrule
synthetic\_n1000\_d2\_k10 & 0.01 & 0.21 & 1.53 \\
synthetic\_n1000\_d2\_k100 & 0.03 & 7.46 & 45.50 \\
synthetic\_n1000\_d2\_k2 & 0.00 & 0.00 & 0.01 \\
synthetic\_n1000\_d2\_k20 & 0.01 & 1.23 & 7.58 \\
synthetic\_n1000\_d2\_k5 & 0.00 & 0.02 & 0.12 \\
synthetic\_n1000\_d2\_k50 & 0.03 & 3.67 & 25.62 \\
synthetic\_n2000\_d2\_k10 & 0.01 & 0.16 & 0.89 \\
synthetic\_n2000\_d2\_k100 & 0.09 & 36.69 & 259.15 \\
synthetic\_n2000\_d2\_k2 & 0.00 & 0.00 & 0.02 \\
synthetic\_n2000\_d2\_k20 & 0.03 & 7.13 & 24.88 \\
synthetic\_n2000\_d2\_k5 & 0.00 & 0.02 & 0.07 \\
synthetic\_n2000\_d2\_k50 & 0.06 & 17.08 & 128.03 \\
synthetic\_n5000\_d2\_k10 & 0.02 & 0.09 & 0.48 \\
synthetic\_n5000\_d2\_k100 & 0.46 & 760.15 & 3232.93 \\
synthetic\_n5000\_d2\_k2 & 0.01 & 0.01 & 0.11 \\
synthetic\_n5000\_d2\_k20 & 0.03 & 1.85 & 6.91 \\
synthetic\_n5000\_d2\_k5 & 0.02 & 0.02 & 0.11 \\
synthetic\_n5000\_d2\_k50 & 0.22 & 994.35 & 2239.26 \\
synthetic\_n500\_d2\_k10 & 0.00 & 0.10 & 0.46 \\
synthetic\_n500\_d2\_k100 & 0.01 & 1.50 & 5.76 \\
synthetic\_n500\_d2\_k2 & 0.00 & 0.00 & 0.01 \\
synthetic\_n500\_d2\_k20 & 0.01 & 0.32 & 1.78 \\
synthetic\_n500\_d2\_k5 & 0.00 & 0.02 & 0.11 \\
synthetic\_n500\_d2\_k50 & 0.01 & 0.81 & 3.93 \\
\bottomrule
\end{tabular}
}
\end{table}

\begin{table}[H]
\caption{ARI on dataset collection 3 with 10.0\% constraints (without centroid mutation)}
\label{tab:ARI_COL3_10_no_mutation}
\centering
\resizebox{0.66\textwidth}{!}{%
\begin{tabular}{c|ccc}
\toprule
Constraints Level & \multicolumn{3}{c}{0.1} \\
Method & KSKM & KSKM\_E & PCCC \\
Dataset &  &  &  \\
\midrule
synthetic\_n1000\_d2\_k10 & 0.94 & 0.98 & 0.98 \\
synthetic\_n1000\_d2\_k100 & 0.26 & 0.26 & 0.26 \\
synthetic\_n1000\_d2\_k2 & 1.00 & 1.00 & 1.00 \\
synthetic\_n1000\_d2\_k20 & 0.65 & 0.68 & 0.69 \\
synthetic\_n1000\_d2\_k5 & 0.99 & 0.99 & 0.99 \\
synthetic\_n1000\_d2\_k50 & 0.43 & 0.43 & 0.44 \\
synthetic\_n2000\_d2\_k10 & 0.99 & 0.99 & 0.99 \\
synthetic\_n2000\_d2\_k100 & 0.27 & 0.27 & 0.28 \\
synthetic\_n2000\_d2\_k2 & 1.00 & 1.00 & 1.00 \\
synthetic\_n2000\_d2\_k20 & 0.88 & 0.89 & 0.90 \\
synthetic\_n2000\_d2\_k5 & 1.00 & 1.00 & 1.00 \\
synthetic\_n2000\_d2\_k50 & 0.47 & 0.50 & 0.51 \\
synthetic\_n5000\_d2\_k10 & 1.00 & 1.00 & 1.00 \\
synthetic\_n5000\_d2\_k100 & 0.32 & 0.35 & 0.35 \\
synthetic\_n5000\_d2\_k2 & 1.00 & 1.00 & 1.00 \\
synthetic\_n5000\_d2\_k20 & 0.98 & 0.98 & 0.98 \\
synthetic\_n5000\_d2\_k5 & 1.00 & 1.00 & 1.00 \\
synthetic\_n5000\_d2\_k50 & 0.67 & 0.70 & 0.71 \\
synthetic\_n500\_d2\_k10 & 0.92 & 0.91 & 0.90 \\
synthetic\_n500\_d2\_k100 & 0.22 & 0.22 & 0.22 \\
synthetic\_n500\_d2\_k2 & 1.00 & 1.00 & 1.00 \\
synthetic\_n500\_d2\_k20 & 0.58 & 0.59 & 0.62 \\
synthetic\_n500\_d2\_k5 & 0.97 & 0.97 & 0.97 \\
synthetic\_n500\_d2\_k50 & 0.39 & 0.39 & 0.40 \\
\bottomrule
\end{tabular}
}
\end{table}

\begin{table}[H]
\caption{Algorithm success rate on dataset collection 3 with 10.0\% constraints (without centroid mutation)}
\label{tab:rate_success_COL3_10_no_mutation}
\centering
\resizebox{0.66\textwidth}{!}{%
\begin{tabular}{c|ccc}
\toprule
Constraints Level & \multicolumn{3}{c}{0.1} \\
Method & KSKM & KSKM\_E & PCCC \\
Dataset &  &  &  \\
\midrule
synthetic\_n1000\_d2\_k10 & 1.00 & 1.00 & 1.00 \\
synthetic\_n1000\_d2\_k100 & 1.00 & 1.00 & 1.00 \\
synthetic\_n1000\_d2\_k2 & 1.00 & 1.00 & 1.00 \\
synthetic\_n1000\_d2\_k20 & 1.00 & 1.00 & 1.00 \\
synthetic\_n1000\_d2\_k5 & 1.00 & 1.00 & 1.00 \\
synthetic\_n1000\_d2\_k50 & 1.00 & 1.00 & 1.00 \\
synthetic\_n2000\_d2\_k10 & 1.00 & 1.00 & 1.00 \\
synthetic\_n2000\_d2\_k100 & 1.00 & 1.00 & 1.00 \\
synthetic\_n2000\_d2\_k2 & 1.00 & 1.00 & 1.00 \\
synthetic\_n2000\_d2\_k20 & 1.00 & 1.00 & 1.00 \\
synthetic\_n2000\_d2\_k5 & 1.00 & 1.00 & 1.00 \\
synthetic\_n2000\_d2\_k50 & 1.00 & 1.00 & 1.00 \\
synthetic\_n5000\_d2\_k10 & 1.00 & 1.00 & 1.00 \\
synthetic\_n5000\_d2\_k100 & 1.00 & 1.00 & 1.00 \\
synthetic\_n5000\_d2\_k2 & 1.00 & 1.00 & 1.00 \\
synthetic\_n5000\_d2\_k20 & 1.00 & 1.00 & 1.00 \\
synthetic\_n5000\_d2\_k5 & 1.00 & 1.00 & 1.00 \\
synthetic\_n5000\_d2\_k50 & 1.00 & 1.00 & 1.00 \\
synthetic\_n500\_d2\_k10 & 1.00 & 1.00 & 1.00 \\
synthetic\_n500\_d2\_k100 & 1.00 & 1.00 & 1.00 \\
synthetic\_n500\_d2\_k2 & 1.00 & 1.00 & 1.00 \\
synthetic\_n500\_d2\_k20 & 1.00 & 1.00 & 1.00 \\
synthetic\_n500\_d2\_k5 & 1.00 & 1.00 & 1.00 \\
synthetic\_n500\_d2\_k50 & 1.00 & 1.00 & 1.00 \\
\bottomrule
\end{tabular}
}
\end{table}

\begin{table}[H]
\caption{Clustering inertia on dataset collection 3 with 15.0\% constraints (without centroid mutation)}
\label{tab:obj_COL3_15_no_mutation}
\centering
\resizebox{0.66\textwidth}{!}{%
\begin{tabular}{c|ccc}
\toprule
Constraints Level & \multicolumn{3}{c}{0.15} \\
Method & KSKM & KSKM\_E & PCCC \\
Dataset &  &  &  \\
\midrule
synthetic\_n1000\_d2\_k10 & 66.91 & 66.91 & 66.99 \\
synthetic\_n1000\_d2\_k100 & 23.55 & 22.93 & 22.79 \\
synthetic\_n1000\_d2\_k2 & 911.71 & 911.71 & 911.71 \\
synthetic\_n1000\_d2\_k20 & 115.41 & 77.20 & 77.25 \\
synthetic\_n1000\_d2\_k5 & 69.58 & 69.58 & 69.58 \\
synthetic\_n1000\_d2\_k50 & 46.43 & 43.67 & 43.15 \\
synthetic\_n2000\_d2\_k10 & 131.52 & 131.52 & 131.54 \\
synthetic\_n2000\_d2\_k100 & 68.73 & 65.94 & 66.29 \\
synthetic\_n2000\_d2\_k2 & 1770.35 & 1770.35 & 1770.35 \\
synthetic\_n2000\_d2\_k20 & 160.77 & 160.77 & 160.78 \\
synthetic\_n2000\_d2\_k5 & 140.71 & 140.71 & 140.71 \\
synthetic\_n2000\_d2\_k50 & 132.31 & 117.31 & 112.06 \\
synthetic\_n5000\_d2\_k10 & 336.35 & 336.35 & 336.35 \\
synthetic\_n5000\_d2\_k100 & 347.52 & 291.70 & 313.21 \\
synthetic\_n5000\_d2\_k2 & 4493.84 & 4493.84 & 4493.84 \\
synthetic\_n5000\_d2\_k20 & 414.46 & 414.46 & 414.46 \\
synthetic\_n5000\_d2\_k5 & 353.15 & 353.15 & 353.15 \\
synthetic\_n5000\_d2\_k50 & 285.08 & 285.04 & 285.23 \\
synthetic\_n500\_d2\_k10 & 33.52 & 33.52 & 33.55 \\
synthetic\_n500\_d2\_k100 & 8.65 & 8.59 & 8.56 \\
synthetic\_n500\_d2\_k2 & 449.70 & 449.70 & 449.70 \\
synthetic\_n500\_d2\_k20 & 44.90 & 40.82 & 40.65 \\
synthetic\_n500\_d2\_k5 & 34.14 & 34.14 & 34.14 \\
synthetic\_n500\_d2\_k50 & 17.84 & 17.49 & 17.25 \\
\bottomrule
\end{tabular}
}
\end{table}

\begin{table}[H]
\caption{CPU time on dataset collection 3 with 15.0\% constraints (without centroid mutation)}
\label{tab:time_COL3_15_no_mutation}
\centering
\resizebox{0.66\textwidth}{!}{%
\begin{tabular}{c|ccc}
\toprule
Constraints Level & \multicolumn{3}{c}{0.15} \\
Method & KSKM & KSKM\_E & PCCC \\
Dataset &  &  &  \\
\midrule
synthetic\_n1000\_d2\_k10 & 0.00 & 0.07 & 0.36 \\
synthetic\_n1000\_d2\_k100 & 0.03 & 20.07 & 104.72 \\
synthetic\_n1000\_d2\_k2 & 0.00 & 0.00 & 0.01 \\
synthetic\_n1000\_d2\_k20 & 0.02 & 3.18 & 7.21 \\
synthetic\_n1000\_d2\_k5 & 0.00 & 0.01 & 0.03 \\
synthetic\_n1000\_d2\_k50 & 0.03 & 10.22 & 58.47 \\
synthetic\_n2000\_d2\_k10 & 0.01 & 0.05 & 0.25 \\
synthetic\_n2000\_d2\_k100 & 0.12 & 136.62 & 4147.90 \\
synthetic\_n2000\_d2\_k2 & 0.00 & 0.00 & 0.04 \\
synthetic\_n2000\_d2\_k20 & 0.01 & 0.83 & 2.98 \\
synthetic\_n2000\_d2\_k5 & 0.01 & 0.01 & 0.04 \\
synthetic\_n2000\_d2\_k50 & 0.07 & 170.18 & 267.80 \\
synthetic\_n5000\_d2\_k10 & 0.03 & 0.04 & 0.23 \\
synthetic\_n5000\_d2\_k100 & 0.52 & 2862.11 & 3682.24 \\
synthetic\_n5000\_d2\_k2 & 0.03 & 0.03 & 0.25 \\
synthetic\_n5000\_d2\_k20 & 0.04 & 0.25 & 1.60 \\
synthetic\_n5000\_d2\_k5 & 0.03 & 0.03 & 0.22 \\
synthetic\_n5000\_d2\_k50 & 0.10 & 75.60 & 53.16 \\
synthetic\_n500\_d2\_k10 & 0.00 & 0.08 & 0.47 \\
synthetic\_n500\_d2\_k100 & 0.01 & 2.45 & 14.78 \\
synthetic\_n500\_d2\_k2 & 0.00 & 0.00 & 0.00 \\
synthetic\_n500\_d2\_k20 & 0.01 & 0.80 & 3.34 \\
synthetic\_n500\_d2\_k5 & 0.00 & 0.01 & 0.05 \\
synthetic\_n500\_d2\_k50 & 0.01 & 1.42 & 7.66 \\
\bottomrule
\end{tabular}
}
\end{table}

\begin{table}[H]
\caption{ARI on dataset collection 3 with 15.0\% constraints (without centroid mutation)}
\label{tab:ARI_COL3_15_no_mutation}
\centering
\resizebox{0.66\textwidth}{!}{%
%
}
\end{table}

\begin{table}[H]
\caption{Algorithm success rate on dataset collection 3 with 15.0\% constraints (without centroid mutation)}
\label{tab:rate_success_COL3_15_no_mutation}
\centering
\resizebox{0.66\textwidth}{!}{%
%
}
\end{table}

\begin{table}[H]
\caption{Clustering inertia on dataset collection 3 with 20.0\% constraints (without centroid mutation)}
\label{tab:obj_COL3_20_no_mutation}
\centering
\resizebox{0.66\textwidth}{!}{%
%
}
\end{table}

\subsubsection{Results: No Must-Link Constraints}
\begin{table}[H]
\caption{Clustering inertia on dataset collection 1 with 5.0\% constraints (without musk-link constraints)}
\label{tab:obj_COL1_5_no_ml}
\centering
\resizebox{0.9\textwidth}{!}{%
%
}
\end{table}

\begin{table}[H]
\caption{CPU time on dataset collection 1 with 5.0\% constraints (without musk-link constraints)}
\label{tab:time_COL1_5_no_ml}
\centering
\resizebox{0.9\textwidth}{!}{%
%
}
\end{table}

\begin{table}[H]
\caption{ARI on dataset collection 1 with 5.0\% constraints (without musk-link constraints)}
\label{tab:ARI_COL1_5_no_ml}
\centering
\resizebox{0.9\textwidth}{!}{%
%
}
\end{table}

\begin{table}[H]
\caption{Algorithm success rate on dataset collection 1 with 5.0\% constraints (without musk-link constraints)}
\label{tab:rate_success_COL1_5_no_ml}
\centering
\resizebox{0.9\textwidth}{!}{%
%
}
\end{table}

\begin{table}[H]
\caption{Clustering inertia on dataset collection 1 with 10.0\% constraints (without musk-link constraints)}
\label{tab:obj_COL1_10_no_ml}
\centering
\resizebox{0.9\textwidth}{!}{%
%
}
\end{table}

\begin{table}[H]
\caption{CPU time on dataset collection 1 with 10.0\% constraints (without musk-link constraints)}
\label{tab:time_COL1_10_no_ml}
\centering
\resizebox{0.9\textwidth}{!}{%
%
}
\end{table}

\begin{table}[H]
\caption{ARI on dataset collection 1 with 10.0\% constraints (without musk-link constraints)}
\label{tab:ARI_COL1_10_no_ml}
\centering
\resizebox{0.9\textwidth}{!}{%
%
}
\end{table}

\begin{table}[H]
\caption{Algorithm success rate on dataset collection 1 with 10.0\% constraints (without musk-link constraints)}
\label{tab:rate_success_COL1_10_no_ml}
\centering
\resizebox{0.9\textwidth}{!}{%
%
}
\end{table}

\begin{table}[H]
\caption{Clustering inertia on dataset collection 1 with 15.0\% constraints (without musk-link constraints)}
\label{tab:obj_COL1_15_no_ml}
\centering
\resizebox{0.9\textwidth}{!}{%
%
}
\end{table}

\begin{table}[H]
\caption{CPU time on dataset collection 1 with 15.0\% constraints (without musk-link constraints)}
\label{tab:time_COL1_15_no_ml}
\centering
\resizebox{0.9\textwidth}{!}{%
%
}
\end{table}

\begin{table}[H]
\caption{ARI on dataset collection 1 with 15.0\% constraints (without musk-link constraints)}
\label{tab:ARI_COL1_15_no_ml}
\centering
\resizebox{0.9\textwidth}{!}{%
%
}
\end{table}

\begin{table}[H]
\caption{Algorithm success rate on dataset collection 1 with 15.0\% constraints (without musk-link constraints)}
\label{tab:rate_success_COL1_15_no_ml}
\centering
\resizebox{0.9\textwidth}{!}{%
%
}
\end{table}

\begin{table}[H]
\caption{Clustering inertia on dataset collection 1 with 20.0\% constraints (without musk-link constraints)}
\label{tab:obj_COL1_20_no_ml}
\centering
\resizebox{0.9\textwidth}{!}{%
%
}
\end{table}

\subsubsection{Comparison with S\_MDEClust}

\input{tables/COL1/COL1_0.05_obj_S_MDEClust}
\input{tables/COL1/COL1_0.05_time_S_MDEClust}
\input{tables/COL1/COL1_0.05_ARI_S_MDEClust}
\input{tables/COL1/COL1_0.05_rate_success_S_MDEClust}
\input{tables/COL1/COL1_0.1_obj_S_MDEClust}
\input{tables/COL1/COL1_0.1_time_S_MDEClust}
\input{tables/COL1/COL1_0.1_ARI_S_MDEClust}
\input{tables/COL1/COL1_0.1_rate_success_S_MDEClust}
\input{tables/COL1/COL1_0.15_obj_S_MDEClust}
\input{tables/COL1/COL1_0.15_time_S_MDEClust}
\input{tables/COL1/COL1_0.15_ARI_S_MDEClust}
\input{tables/COL1/COL1_0.15_rate_success_S_MDEClust}
\input{tables/COL1/COL1_0.2_obj_S_MDEClust}
\input{tables/COL1/COL1_0.2_time_S_MDEClust}
\input{tables/COL1/COL1_0.2_ARI_S_MDEClust}
\input{tables/COL1/COL1_0.2_rate_success_S_MDEClust}
\input{tables/COL2/COL2_0.05_obj_S_MDEClust}
\input{tables/COL2/COL2_0.05_time_S_MDEClust}
\input{tables/COL2/COL2_0.05_ARI_S_MDEClust}
\input{tables/COL2/COL2_0.05_rate_success_S_MDEClust}
\input{tables/COL2/COL2_0.1_obj_S_MDEClust}
\input{tables/COL2/COL2_0.1_time_S_MDEClust}
\input{tables/COL2/COL2_0.1_ARI_S_MDEClust}
\input{tables/COL2/COL2_0.1_rate_success_S_MDEClust}
\input{tables/COL2/COL2_0.15_obj_S_MDEClust}
\input{tables/COL2/COL2_0.15_time_S_MDEClust}
\input{tables/COL2/COL2_0.15_ARI_S_MDEClust}
\input{tables/COL2/COL2_0.15_rate_success_S_MDEClust}
\input{tables/COL2/COL2_0.2_obj_S_MDEClust}
\input{tables/COL2/COL2_0.2_time_S_MDEClust}
\input{tables/COL2/COL2_0.2_ARI_S_MDEClust}
\input{tables/COL2/COL2_0.2_rate_success_S_MDEClust}
\input{tables/COL2/COL2_0.25_obj_S_MDEClust}
\input{tables/COL2/COL2_0.25_time_S_MDEClust}
\input{tables/COL2/COL2_0.25_ARI_S_MDEClust}
\input{tables/COL2/COL2_0.25_rate_success_S_MDEClust}
\input{tables/COL2/COL2_0.3_obj_S_MDEClust}
\input{tables/COL2/COL2_0.3_time_S_MDEClust}
\input{tables/COL2/COL2_0.3_ARI_S_MDEClust}
\input{tables/COL2/COL2_0.3_rate_success_S_MDEClust}
\input{tables/COL2/COL2_0.35_obj_S_MDEClust}
\input{tables/COL2/COL2_0.35_time_S_MDEClust}
\input{tables/COL2/COL2_0.35_ARI_S_MDEClust}
\input{tables/COL2/COL2_0.35_rate_success_S_MDEClust}
\input{tables/COL2/COL2_0.4_obj_S_MDEClust}
\input{tables/COL2/COL2_0.4_time_S_MDEClust}
\input{tables/COL2/COL2_0.4_ARI_S_MDEClust}
\input{tables/COL2/COL2_0.4_rate_success_S_MDEClust}
\input{tables/COL2/COL2_0.45_obj_S_MDEClust}
\input{tables/COL2/COL2_0.45_time_S_MDEClust}
\input{tables/COL2/COL2_0.45_ARI_S_MDEClust}
\input{tables/COL2/COL2_0.45_rate_success_S_MDEClust}
\input{tables/COL2/COL2_0.5_obj_S_MDEClust}
\input{tables/COL2/COL2_0.5_time_S_MDEClust}
\input{tables/COL2/COL2_0.5_ARI_S_MDEClust}
\input{tables/COL2/COL2_0.5_rate_success_S_MDEClust}